\documentclass[a4paper]{book}
\usepackage[T1]{fontenc}
\usepackage[utf8]{inputenc}
\usepackage{fullpage}
\usepackage{amsmath,amssymb}
\usepackage{amsthm}
\usepackage{graphicx,graphics}
\usepackage{hyperref}
\usepackage[capitalise]{cleveref}
\usepackage{enumitem}
\usepackage{natbib}

\usepackage{titlesec}
\usepackage{float}
\usepackage{subfig}
\usepackage{array}
\usepackage{booktabs}
\usepackage{caption}
\usepackage{algorithm}
\usepackage{algpseudocode}
\usepackage[titletoc]{appendix}
\usepackage[nottoc]{tocbibind}
\hypersetup{
  colorlinks=true,
  citecolor=magenta,
  urlcolor=blue
}
\newtheorem{theorem}{Theorem}[chapter]

\newtheorem{lemma}[theorem]{Lemma}

\newtheorem{assumption}{Assumption}
\crefname{assumption}{Assumption}{Assumptions}

\newcommand{\R}[0]{\mathbb{R}}
\newcommand{\N}[0]{\mathbb{N}}
\newcommand{\Exp}[1]{E\!\left[#1\right]}

\newcommand{\RS}{{$\Xi$ }}
\newcommand{\dotp}[1]{\langle #1 \rangle}

\DeclareMathOperator{\rank}{rank}
\DeclareMathOperator{\trace}{trace}
\DeclareMathOperator{\diag}{diag}

\DeclareMathOperator{\conv}{conv}

\title{Influence of Recommender Systems on Users: A Dynamical Systems Analysis}
\author{Prabhat Lankireddy\thanks{Center for Machine Intelligence and Data science, Indian Institute of Technology Bombay}
  \and Jayakrishnan Nair\thanks{Department of Electrical Engineering, Indian Institute of Technology Bombay}
  \and D Manjunath\footnotemark[2]}
\date{2025}

\begin{document}

\frontmatter
%=====================================================================
% Create the title page
\maketitle

%=====================================================================

% Abstract
\section*{Abstract}
We analyze the unintended effects that recommender systems have on the
preferences of users that they are learning. We consider a contextual
multi-armed bandit recommendation algorithm that learns optimal
product recommendations based on user and product attributes. It is
well known that the sequence of recommendations affects user
preferences. However, typical learning algorithms treat the user
attributes as static and disregard the impact of their recommendations
on user preferences. Our interest is to analyze the effect of this
mismatch between the model assumption of a static environment and the
reality of an evolving environment affected by the recommendations. To
perform this analysis, we introduce a model for the coupled evolution of
a linear bandit recommendation system and its users, whose preferences
are drawn towards the recommendations made by the algorithm. We
describe a method, that is grounded in stochastic approximation
theory, to come up with a dynamical system model that asymptotically
approximates the mean behavior of the stochastic model. The resulting
dynamical system captures the coupled evolution of the population
preferences and the learning algorithm. Analyzing this dynamical
system gives insight into the long-term properties of user preferences
and the learning algorithm. Under certain conditions, we show that the
RS is able to learn the population preferences in spite of the model
mismatch. We discuss and characterize the relation between various
parameters of the model and the long term preferences of users in this
work. A key observation is that the exploration-exploitation tradeoff
used by the recommendation algorithm significantly affects the long
term preferences of users. Algorithms that exploit more can polarize
user preferences, leading to the well-known filter bubble phenomenon.

% Table of contents
\tableofcontents

% Spacing Rules

% Equation Spacing
% \abovedisplayskip=8mm
% \abovedisplayshortskip=8mm
% \belowdisplayskip=8mm
% \belowdisplayshortskip=8mm

% Chpater, section, paragraph spacing
\setlength{\parskip}{1mm}
\titlespacing{\chapter}{0cm}{10mm}{10mm}
\titleformat{\chapter}[display]
  {\normalfont\huge\bfseries\centering}
  {\chaptertitlename\ \thechapter}{20pt}{\Huge}
  
  \titlespacing*{\section}
  {0pt}{4mm}{4mm}
  \titlespacing*{\subsection}
  {0pt}{2mm}{2mm}
\pagebreak
%\newpage  
%\cleardoublepage\pagenumbering{arabic}
%\pagenumbering{arabic}

%=====================================================================
\mainmatter
% \newpage
% \pagebreak

%===========================
% Chapter 1
\chapter{Introduction}

\section{Background and Motivation}
\label{sec:background}

Recommender systems (RS) use algorithms to suggest relevant items to their users from a large pool of items. These items can be, for example, music, images, movies in media streaming sites, products in e-commerce platforms, or articles in news websites. According to \citet{resnick1997recommender}, RSs assist and augment the natural social process of people recommending items to each other. Today, RSs are everywhere and they do more than just assist the process of recommendations. The recommendation algorithms used today are powerful and were demonstrated to significantly enhance user engagement on platforms such as YouTube~\citep{covington2016deep} and Amazon~\citep{smith2017twoamazon}. Recommendation algorithms play an important role in determining what items reach the user population. 

The omnipresence of recommendation algorithms has the following consequence: they affect its users in ways possibly unplanned. A benign example of this is a person who discovers her new favorite album through suggestions from a music recommender system. In this example, the consumption pattern of the user is influenced by the recommender. A more serious example is how a social media platform like Facebook is perhaps incentivised to recommend anger-inciting content such as hate speech to its users \citep{munn2020angry}. This can happen when the content recommendation algorithm is trained to maximize engagement, and users tend to engage (like, comment and share) more with hateful content. A second consequence is that recommender systems tend to recommend smaller sets of content over time, that the users agree with, in order to maximize engagement---a phenomenon known as a \textit{filter bubble} \citep{nguyen2014exploring}. These point to an important research gap: the need to better understand the interaction between a recommendation algorithm and its users.

The need for such work is well advocated by the research community. \cite{franklin2022recognising} propose a multi-disciplinary approach to understand how user preferences are affected by modern machine learning systems. They classify preference changes into different types and discuss which kind of changes can be permissible. To better understand the consequences of using recommendation systems, \citet{dean2024accounting} argue for the development of \textit{formal models of interaction} between recommendation systems and their users. Such models can further be used to develop responsible recommender systems. Our work develops one such formal model of interaction with the aim of understanding unintended consequences of recommender systems on their users.

\section{Research objectives}

People are influenced by the content they consume and, naturally, they are influenced by RSs that suggest content to them.
However, many RSs are not designed to take into account the influence of recommendations on their users. Such algorithms assume that user preferences are static, i.e., preferences do not change with time. One well-known example of this is the matrix factorization algorithm~\citep{koren2009matrix} in which the rating matrix, which encodes all user-item preferences, is assumed to be time-invariant.
Another example is the contextual bandit recommendation algorithm, in which the reward/rating for a given recommendation is modeled as a noisy linear function of a static vector that encapsulates user preferences.
We argue that the assumption of static user preferences is a strong assumption for these algorithms and it often does not hold. Such an assumption introduces a mismatch between the environment model of RS and the true environment.
For such algorithms, the effect of recommendations on their users are \textit{unintended consequences}, or side effects.
The objective of our work is to model the interaction between such recommender systems (algorithms that assume static user preferences) and its users whose preferences change with time.
%In particular, we consider a contextual bandit recommender system as the algorithm, and users whose preferences tend to be drawn towards recommendations shown to them.
Specifically, we consider the contextual bandit recommendation algorithm that learns from interaction with its users while assuming that they have time-invariant preferences, while the user preferences are evolving in response to the recommendations. That is, unbeknownst to the algorithm, it is influencing the environment that it is learning.

In this work, we motivate the identification of possibly unintended consequences due to RSs by presenting a theoretical method to analyze the interaction between an RS and its users. Such an analysis is enabled by models of human influence that have been developed within the fields of opinion dynamics and cognitive science. Using tools from stochastic approximation theory, the proposed method constructs a deterministic model, a dynamical system, that captures the ``mean'' behavior of the stochastic model of interaction. Further analysis of the resulting deterministic model with the help of dynamical systems theory gives insight into the unintended consequences of interest.

\section{Literature Review}
In the following, we present an brief overview of literature on understanding how recommender systems affect their users. First, we discuss existing work that models the interaction between recommender systems and its users. Our work contributes to this body of literature.

To understand the effect of recommender systems on their users, it is necessary to discuss and understand the behavior of recommender systems and user preference dynamics in isolation.
Hence, in the section after the next, we present various settings and assumptions around which recommendation algorithms are designed. Following that, we discuss the different ways in which user preferences are modeled in the literature.

\subsection{Interactions between RS and users}
To understand the effects of AI systems on their users, \cite{dean2024accounting} argue for the development of formal models that describe the interaction between AI systems and their users, which is the focus of our work. In this section, we discuss other existing literature that share a similar goal.

One of the earliest works in this domain is the work by \cite{jiang2019degenerate}, in which the feedback loop between a recommender system and a user is model and phenomena like echo chambers and filter bubbles are shown to be emergent.
The work by \cite{rossi2021closed} discusses the effect of news recommendations on the opinion of users. They consider two types of content with opposing positions on a certain topic, 1 and -1, and model the users with scalar opinions in the interval $(-1,1)$ that indicates the position with which the user agrees more. They show that decreasing randomness in recommendations tends to increase the effectiveness of the algorithm while also polarizing users' opinions more.
\cite{kalimeris2021preference} consider a model in which the recommender uses an algorithm similar to matrix factorization to recommend items to users whose preferences tend to gradually shift towards the recommended items. They analyze the model and discuss its consequences---user engagement increases at the cost of reduced diversity of preferences and the emergence of echo chambers.
\cite{dean2022preference} analyze a preference dynamics model of biased assimilation, in which the users agree more with recommendations that align with their current belief while rejecting recommendations that do not. They show that the problem of regret minimization is trivial in such a setting after knowing the type of content that the user likes. They propose a recommendation scheme that minimizes the effect of recommendations on the user's initial opinion.
\cite{brown2022diversified} consider an adversarial bandit setting in which a recommender, which maximizes the long term reward that is regularized to include some level of exploration, interacts with a user with an unknown preference model. In such a setting, they characterize the set of preference models that can be learned by the recommender which can then be used to minimize regret.
\cite{kleinberg2024challenge} consider a model consisting of a media-recommending platform that aims to maximize user utility but observes only the user's engagement on the platform. They argue that a user's observed behavior is not indicative of the user's underlying preferences, by showing that users derive minimal utility from the platform if the recommender algorithm uses the observed engagement as a proxy for user utility.

In the spirit of the proposal made by \cite{dean2024accounting} towards the development of formal interaction models, we propose a model of interaction between RS and users. We focus our analysis on a contextual bandit recommendation system; such algorithms are well-established and known to work well in applications such as news and music recommendations. Further, we discuss a method of analysis rooted in dynamical systems theory in order to understand the long term properties of such an interaction. This analysis method is derived from the ODE method of stochastic approximation, which lets us analyze a stochastic system using a deterministic system that captures the asymptotic behavior of the former. Moreover, this method is powerful because it can be used on models other than the one discussed in this thesis. We also fill an important gap that has not been discussed so far in the literature: the impact of recommendations on user preferences in the presence of multiple users.

%The idea is to theoretically understand the dynamics of learning user preferences. All algorithm development belongs to this section.
\subsection{Modeling the learning algorithm}
\paragraph{Algorithms that assume static preferences}
Most latent factor models, i.e., models that assume that user behavior is characterized by unknown parameters that can be learned, assume that user preferences are static. For example, consider matrix factorization \citep{koren2009matrix} which is a classic latent factor model and one of the most popular recommendation algorithms used today.
Matrix factorization assumes the existence of a \textit{rating matrix}, which contains the ratings each user associates to each item available to the recommender system. The algorithm is given a sparsely-filled rating matrix, and the goal of the algorithm is to find these missing values using the observed values. To assume that a global rating matrix exists is the same as assuming that user preferences are \textit{static}, i.e., preferences of a user towards any given item does not change with time. The same is true for many classical recommendation algorithms based on collaborative filtering; we refer the reader to the monograph by \citet{ekstrand2011collaborative} for more information about such recommendation algorithms.

With the advent of deep learning, many recommendation algorithms that use deep learning have emerged (refer to the survey by \citet{zhang2019deep} for more on this topic). However, while these algorithms have been used to improve recommendations by learning nonlinear relationships between users and items, most settings still assume that the user preferences remain invariant with time. As an example, consider neural collaborative filtering algorithms which are deep learning counterparts of classical collaborative filtering algorithms like matrix factorization. Neural collaborative filtering algorithms also assume the existence of a fixed unknown rating matrix, and hence affects users similar to how matrix factorization algorithms do the same. That said, certain deep learning algorithms account for dynamic user preferences; we discuss these exceptions in a subsequent paragraph dedicated to algorithms that learn user preference dynamics from data.

Another class of algorithms that are used for recommendations are \textit{multi-armed bandit algorithms}, which are online learning algorithms that learn from feedback after every recommendation. There are many variants of multi-armed bandit algorithms, and in this work, we assume the use of such algorithm. Specifically, we focus on \textit{contextual linear bandits}, which can make personalized recommendations based on a context that captures information about a given user. Analysis of these algorithms is attractive for multiple reasons. Firstly, these algorithms assume static user preferences, and represent other such algorithms well. Moreover, bandit-based recommendation algorithms are well-known and used in the industry---examples include news recommendation~\citep{li2010contextual} and music recommendation~\citep{mcinerney2018explore}. These algorithms also naturally incorporate side information available about users and items, through the use of \textit{context vectors}.

\paragraph{Algorithms that explicitly model user dynamics} In such work, the authors design recommendation algorithms while assuming that the user preferences change according to a prescribed model. For instance, \citet{meshram2015restless} study a setting in which user preferences are modeled as a restless multi-armed bandit. Particularly, the user interest towards an item is modeled as a Markov decision process in which the transition probabilities dictate the user preference dynamics. \citet{kleinberg2018recharging} consider a stochastic bandit problem in which the mean reward of an arm increases with the time that the arm has been last chosen. \citet{shah2018bandit} discuss the setting in which future user arrivals depend on the experiences of similar users in the past. \citet{zhou2021incentivized} consider a similar setting in which the algorithm does not choose the arm, but assists the user in choosing the arm by providing incentives. Such work is very relevant today since user preferences are known to be dynamic, and such algorithms are expected to perform better than classical recommendation algorithms. That said, unlike the work discussed above, the focus of our work is not algorithmic development in such settings. We focus on analyzing what happens to the user preferences as they consume recommendations made by algorithms.

\paragraph{Algorithms that learn user preference dynamics from data}
In such work, the user preference dynamics are implicitly/explicitly learned from data. Certain variants of the matrix factorization algorithm take temporal dynamics of user preferences into account. For example, \citet{koren2009collaborativetemporal} assumes a linear model for preference evolution and predicts a time-dependent user-item rating, where the linear model is learned from the dataset through cross validation.
Certain deep-learning based recommender systems also take user preference dynamics into account. Consider the problem of next-item prediction, in which the algorithm must recommend an item to the user given a history of items that the user consumed. Some of these algorithms learn short-term preferences or session-based preferences, in which the preference dynamics within a single user session is taken into account to predict the next item \citep{li2017neural}. There are other works that consider the users' long-term preferences in addition to session-based preferences \citep{li2018learning}. Such deep learning recommenders implicitly learn the user preference dynamics from data.
Moreover, the broad area of reinforcement learning-based recommender systems considers user preferences as dynamic by treating them as time-varying states. Such algorithms pose the problem of recommendations as a Markov decision process in which the algorithm is the agent and users are the environment that provide rewards. The goal of the agent learn a recommendation policy that maximizes the expected long-term reward. We refer the reader to a survey by \citet{afsar2022reinforcement} for more information about reinforcement learning-based recommenders.
While all these algorithms are shown to improve the effectiveness of recommendations, the effect of recommendations on the user preferences remains unclear.

\subsection{Modeling the influence on the users}

There are broadly two approaches to modeling user preference dynamics: using pre-existing models from literature, or learning them from data. There is a large amount of work on modeling how the opinions of people change as they interact with other people; the field of work is called \textit{opinion dynamics}. Such work can be extended to model how user preferences change due to recommendations. For example, the idea of \emph{biased assimilation}, in which users update their beliefs to those that closely align with their current beliefs, is discussed in \cite{dandekar2013biased}. In this work, the proposed model of influence is extended to understand the impact of recommendations made by an algorithm on its users. We refer the reader to the excellent tutorial by \citet{proskurnikov2017tutorial} for an overview of well-known models in opinion dynamics and the theory underlying such models. One can also leverage models from psychology literature to understand user preference dynamics. \cite{curmei2022towards} do exactly that by presenting three user preference models grounded in psychology literature.

In the second approach, the work by \cite{carroll2022estimating} proposes a method to learn user preferences from data. They argue for the development of recommender systems that estimate the effect of their recommendations on its users and use these estimates to make recommendations that avoid undesirable shifts in users' preferences. Another work that learns user preferences from data is by \citet{wang2023causal}, in which a causal graph representing the factors influencing user preferences is learned from data. This work primarily focuses on preference shifts caused due to external factors and not necessarily on the shifts introduced by the recommendations themselves.

\section{Outline}
	The subject matter of the thesis is presented as described in the following.
\begin{itemize}
%\item	\cref{ch:lit-review} discusses the existing work on the topic, while also comparing the current work with other similar works in order to highlight the differences.
\item	\cref{ch:recommendation-model} descibes the learning algorithm and introduces the method of asymptotic analysis that we use throughout the text. The analysis method is applied to the algorithm to show that the algorithm asymptotically reaches the optimal policy.
\item	\cref{ch:one-user-model} introduces a model of interaction between a recommendation algorithm and a user whose preferences change with time. The analysis method introduced in the previous chapter is used to understand the asymptotic properties of the given model of interaction. The effect of the recommendation algorithm on the asymptotic properties is discussed.
\item \cref{ch:multi-user-model}  extends the model of interaction introduced in the previous chapter to include multiple users. 
\item	\cref{ch:conclusion} summarizes the results and inferences obtained from the analysis of the presented models. The scope for future and continuation of this research work are also reported.  
\end{itemize}

\section{Other related work}
Here, we compare our theme of work with other popular themes and explain the differences.

\paragraph{Empirical work}
There is plenty of work that attempts to understand the effect of recommender systems on its users by designing experiments, gathering real world data from users and testing hypotheses. For instance, \cite{adomavicius2013recommender} discuss how ratings of recommended items on e-commerce website impact preferences of users at the time of consumption. \cite{porcaro2024assessing} studied the impact of music recommendations on the listening patterns of users over time. While this body of work has goals similar to ours, the research methodology used is significantly different---our work takes a theoretical approach to the problem rather than an empirical approach. It is expected that both kinds of work complement each other and improve the overall understanding we have about the human-algorithm interaction. For example, \citet{nguyen2014exploring} empirically observed that users receiving movie recommendations from a RS received a narrowing set of recommendations over time, and the users who consumed such recommendations had a positive experience. This observation is reflected in the results of our work: algorithms that aggressively recommend certain set of items makes users prefer a narrowing set of items over time.

\paragraph{Bias and causality}
The problem of bias in recommender systems arises due to imbalance present in the data used to train recommendation algorithms. This often happens when the algorithm assumes that the training data is independently and identically distributed. For example, consider an algorithm recommending two items to a user. As the algorithm learns that the user prefers item 1 over item 2 (without loss of generality), it starts recommending item 1 more often, which eventually causes the recommendation history to contain more data on recommending item 1 as compared to that of item 2. Training on such biased data causes the algorithm to recommend item 1---this effectively causes a positive feedback loop in the training process. \citet{mansoury2020feedback} proposed a simulation-based method to characterize this feedback loop and understand the phenomenon of popularity bias i.e. the phenomenon in which the algorithm aggressively recommends a small set of available items and ignores most other items. \cite{chaney2018algorithmic} discuss how such feedback loops increases homogenity in the user consumption patterns without improving utility. Causal recommender systems address how to correct the bias process during training. For more reading on the topic, we refer the reader to the surveys by \cite{chen2023bias} and \cite{gao2024causal} on bias and causality in recommender systems respectively.

Our work on shifts in user preferences is different from the work on bias in recommendation algorithms. The subject of the former are the users, while the subject of the latter is the recommender. Moreover, the term ``feedback loops'' is used in both of these works in different contexts. When discussing bias, the focus is on feedback loops that are caused by imbalance in the training data. This is orthogonal to feedback loops caused due to the shifts in user preferences; one can observe bias in recommendation systems even under the assumption of static preferences.

%\paragraph{Fairness}

% \paragraph{Trust}
% The concept of trust in recommendation systems discusses the users' belief in RS to make relevant recommendations. While trust can be dynamic, it is orthogonal to the idea of dynamic user preferences. Typically trust is studied for entire RS's rather than individual recommendations. \cite{brown2020stackelberg} studies trust from a game theoretic perspective.

\paragraph{Performative prediction} The literature on performative prediction \citep{hardt2022performative} studies the setting in which predictions made by a machine learning algorithm affect the distribution of the data used to train such an algorithm. This causes algorithms that are subsequently trained on the affected data distribution to perform differently from the initial algorithm, thereby creating a feedback loop between the data and the algorithm. Recommender systems fit well in this setting because the recommendations, which are the algorithms predictions, affect user preferences i.e. the data distribution. This is an effective paradigm to understand recommender systems that use supervised learning algorithms. Our work differs from performative prediction literature because we consider an online learning framework in which data streaming and learning happens simultaneously. Moreover, existing work in this literature does not focus on user preferences, which is the primary focus of this work.

\paragraph{Alignment between Recommmendations and User Utility}
There is work that questions whether the way in which recommender systems work aligns with the goals of its users. For example, a user browsing a video-streaming platform like YouTube might want to quickly find the video she is looking for, but the algorithm recommends potentially irrelevant items with the goal of maximizing engagement. \citet{kleinberg2024challenge} consider a model of interaction between a recommendation algorithm and a user to show that a user's content consumption behavior might not be representative of user preferences. Hence, training recommendation algorithms with the content consumption patterns of users might not improve user utility. To improve the alignment between the objectives of the algorithms and its users, \citet{agarwal2024system} proposes to replace user engagement time with the probability of a user returning to the platform as a measure of utility. The justification is that the latter captures user utility while the former captures users' impulsive responses. The difference between such work and our work arises from the difference between utility and preferences. Utility is derived from preferences, and the work on improving utility focuses on the question of understanding user preferences well. Our work, on the other hand, focuses on what happens to preferences over time as the user is exposed to the recommendation algorithm.

% \paragraph{Context-aware recommender systems}
% \cite{adomavicius2010context}

\paragraph{Choice architecture and persuasion}
The process of user arriving at the best choice can be viewed as a collaborative effort involving the user and the recommendation algorithm. \cite{jameson2015human} discusses this viewpoint in detail, and argues that the designer of a recommender system must understand the process of how humans make decisions. Recommendations influence user preferences, and this work discusses how to effectively influence preferences in order to help users achieve their goals---this is called choice architecture. A key difference between this literature and our work is that the focus of the former is on making effective recommendations in the short-term, while the latter focuses on the long-term effect of recommendations on user preferences.

%===========================
% Chapter 2
\chapter{Recommendation Algorithm Analysis}
\label{ch:recommendation-model}

The objective of this chapter is to first introduce the the class of
recommendation algorithms of interest and then to set up method of
analysis that will be used subsequently.

We pose the recommendation problem as a contextual linear bandit
problem~\citep{lattimore2020bandit}. To solve this problem, we
present and analyze an algorithm that uses a softmax-based policy for
recommendation along with a follow the regularized leader (FTRL)
scheme~\citep{shalev2012online} for improving the policy online. This model of recommendation also has the essential properties of interest: (i) it
assumes stationary user preferences, (ii) it uses side information about
users and items that is known to the recommender, (iii) the algorithm
has performance guarantees (sub-linear regret/convergence to the
optimal policy), and (iv) it is convenient from an analysis point of view
due to the small number of hyperparameters, and because the
recommendation policy is continuous and differentiable with respect to
its parameters. We argue that the qualitative results that arise from the
analysis of this model carry over to more complex models with
similar properties.

After discussing the algorithm, we introduce an analysis method used to understand the asymptotic properties of the algorithm. The analysis method is known as the \emph{Ordinary Differential Equation (ODE) method} of stochastic approximation, and the idea underlying the method is to obtain an ODE with asymptotic properties identical to those of the algorithm. This method is powerful because it lets us analyze a deterministic system to understand the asymptotic properties of a more complicated stochastic system of interest. We discuss the ODE method in detail and apply this ODE method to analyze models of recommender systems and users in the current chapter and subsequent chapters as well.
%TODO: Write about how this chapter and model discussed helps in the subsequent chapters. Also do that in the conclusion.

The recommendation algorithm is developed in
\cref{sec:level-1-model-description}. Following that, we discuss the
ODE method of stochastic approximation, which is our analysis method
of choice, in \cref{sec:ode-method}. Finally, we analyze the long-term
behavior of the model using the ODE method and thereby establish
guarantees for the algorithm in \cref{sec:level-1-asymptotic-ode}.

% In this chapter, a simple model of interaction between a user and a recommendation system (RS) is presented and discussed. Before discussing the model, a preliminary discussion on the algorithm used by RS is presented along with the assumptions and guarantees that come with it. Subsequently, the model of a user is discussed, following which the model of interaction is presented. A special case of the model, in which RS recommends only two items, is discussed to aid the reader.

\section{Problem Setting and Algorithm Description}
\label{sec:level-1-model-description}
%TODO: Consider dedicating a symbol to RS. This symbol refers to the 'current' recommender system and not a general RS.
Consider a recommender system \RS giving recommendations to a user at discrete times
$t=1,2,3,\ldots$ from a pool of $K \in \mathbb{N}$ items. Each item
indexed by $k$ is associated with a $d$-dimensional attribute $w_k.$ A
possible interpretation is that each component of an item attribute
represents a specific property of the corresponding item; items with a
similar properties have attributes that are closer to each other
(according to some distance metric defined over $\R^d$). These
attributes are used by \RS to determine the recommendation. Let $W =
\begin{bmatrix}
  w_1 & w_2 & \cdots & w_K
\end{bmatrix}
$ be the $d \times K$ matrix containing all item attributes available
to \RS. We are interested in the setting where $K \geq d$; if the
attribute size $d$ were larger than $K,$ then one might be better off
choosing $W=I_K$ (the identity matrix of dimension $K$), which uses
fewer parameters for each item despite not making use of the
similarities between items. Moreover, it is assumed that $W$ is full
rank i.e., $\rank(W)=d.$ If $\rank(W)<d,$ then there are redundant
components in the item attributes which do not add additional
information because they are linear combinations of other available
components.

After making a recommendation to the user, \RS receives a reward from
the user, which is used to improve subsequent recommendations. \RS
makes the following assumption about the structure of the reward,
i.e.,
\begin{equation}
  \label{eq:reward-learner-model}
  \text{Reward} = \left< w, \psi \right> + \epsilon.
\end{equation}
Here $w\in\R^d$ is the attribute of the item recommended,
$\psi \in \R^d$ is the unknown time-invariant \emph{user preference vector,}
$\left< \cdot,\cdot \right>$ is the standard inner product on $\R^d,$
and $\epsilon$ is real valued noise with zero mean and finite variance
$\sigma^2.$
% with variance proxy $\sigma^2.$
$\epsilon$ is assumed to be independent of both $w$ and $\psi.$ The
user preference vector $\psi$ has the following intuitive
interpretation: each component of $\psi$ is the affinity of the user
towards the corresponding component of the item attribute. This leads
to the interpretation that $\dotp{w,\psi}$ is the overall affinity of
the user with preference $\psi$ towards the item with attribute $w.$
The reward that \RS observes is a noisy representation of that overall
affinity. For example, in the setting of media recommendations, the
action of the user clicking over content recommended by \RS can be
treated as the reward. In this case, the reward is modeled as a
Bernoulli random variable, and the probability of clicking (to be
interpreted as the expected reward) on an item with attribute $w$ is
represented by $\dotp{w,\psi}.$

The goal of \RS is to maximize the sum of all future rewards. To
that end, \RS must ensure a tradeoff between \emph{exploration} (making recommendations uniformly at random) and \emph{exploitation} (using past
information to make appropriate recommendations). There are two parts
to \RS---a recommendation policy used to recommend items to the user,
and a learning algorithm that uses past data to improve the
recommendation policy. Both of these are discussed in the following.

\begin{algorithm}
  \caption{Recommendation policy and learning algorithm used by \RS}
  \begin{algorithmic}[1]
    \Require Number of items $K,$ attribute size $d,$ item attributes $w_k \in \R^{d}$ for all $k \in [K],$ regularization parameter $\lambda,$ recommendation parameter $a$
    % \Ensure Output description (e.g., sorted array $A$)
    \State $t \gets 1$
    \State $\theta_1 \gets 0_d$
    % \State $S_0 \gets \lambda \sum_{k=1}^K w_k w_k^{\top}$
    % \For{t=1,2,3,\ldots}
    \Loop
      \State Recommend $A_t \in [K]$ with $P(A_t=k) = \frac{\exp(a w_k^{\top} \theta_t)}{\sum_{j=1}^K \exp(a w_j^{\top} \theta_t)}$
      \State Collect reward $R_t$
      \State $\theta_{t+1} \gets  \arg \min_{\theta} \lambda \|\theta\|_{W W^{\top}}^2 + \sum_{s=1}^t \left( R_s - w_{A_s}^{\top} \theta \right)^2$
      \State $t \gets t+1$
    \EndLoop
    % \State \Return result
  \end{algorithmic}
\end{algorithm}

\paragraph{Recommendation Policy:} Let $A_t$ denote the item that \RS
recommends to the user at time $t.$ A parameterized stochastic policy
is used to recommend items to the user, and the parameter of the
policy at time $t$ is $\theta_t.$ We refer to $\theta_t$ as the
\emph{algorithm state}. In particular, at any given time $t,$ item $k$
is recommended with probability $\pi_k(\theta_t) := P(A_t=k),$ where
\begin{equation}
  \label{eq:level-1-rec-policy}
  \pi_k(\theta_t) = \frac{\exp(a w_k^{\top} \theta_t)}{\sum_{j=1}^K \exp(a w_j^{\top} \theta_t)}.
\end{equation}

The parameter $a \in (0,\infty)$ is called the exploration-exploitation
tradeoff parameter; it determines a balance between exploration and
exploitation done by the policy. As the value of $a$ increases, the
probability assigned to the ``best'' recommendations of the learned
policy increases, and as $a$ decreases, the distribution of
recommendations gets closer to a uniform distribution.
%TODO: Relate this to Boltzmann exploration if needed
% This parameter is useful for analysis because one can understand how
% the properties of the algorithm change as the exploration-exploitation
% tradeoff changes.
The quantity $w_k^{\top}\theta_t$ is the estimate of the expected
reward corresponding to item $k$ at time $t.$ From the definition of
reward in Eq.~\eqref{eq:reward-learner-model}, one can observe that
the estimate of the expected reward is close to the expected reward
when $\theta_t$ is close to $\psi.$

Softmax-based policies are popular in multi-armed bandit
literature~\cite[Section~2.7]{sutton1998reinforcement}, with the
desirable property of having a continuous and differentiable policy.
Such a property is desirable from an analysis perspective because
first-order/second-order methods from calculus can be used to
understand the properties of the algorithm better. It is also useful
for a practitioner because one can optimize softmax-based policies
using gradient descent and off-the-shelf automatic differentiation
packages. Another desirable property is that each value of $a$ defines
a recommendation policy. Because $a \in (0,\infty),$ the definition in
\cref{eq:level-1-rec-policy} introduces a class of policies that
correspond to different values of $a.$ Analysis of such a class of
policies is desirable because the properties of such policies are
similar to various other recommendation policies that determine their
own degree of exploration-exploitation tradeoff. We argue that the
results of our analysis would carry over to other similar policies as
well. For example, the softmax-based recommendation policy can be
interpreted as a smooth-distribution variant of the $\epsilon$-greedy
recommendation policy, another well-known algorithm for the
multi-armed bandit problem.

\paragraph{Learning Algorithm:} In response to the recommendation
$A_t,$ \RS receives a reward $R_t=w_{A_t}^{\top} \psi + \epsilon_t$
from the user. Each element in the noise sequence
$(\epsilon_t)_{t \in
  \mathbb{N}}$ % Regarding the notation for sequences: Use {} instead of () if that is preferrable
is independent of other elements in the sequence and is distributed
identically to $\epsilon,$ which is defined in
Eq.~\eqref{eq:reward-learner-model}. After receiving $R_t,$ \RS updates
the algorithm state from $\theta_t$ to $\theta_{t+1}$ such that the
new algorithm state $\theta_{t+1}$ minimizes a regularized least
squares loss between the rewards and the estimated rewards until time
$t,$ i.e.,
\begin{equation}
  \label{eq:level-1-theta-defn}
  \theta_{t+1} =  \arg \min_{\theta} \lambda \|\theta\|_{WW^{\top}}^2 + \sum_{s=1}^t \left( R_s - w_{A_s}^{\top} \theta \right)^2.
\end{equation}
Here $\|x\|_{WW^{\top}} = x^{\top} W W^{\top} x$ is the Mahalanobis
norm with respect to the positive definite matrix $WW^{\top}.$ This
update is equivalent to the Follow the Regularized Leader (FTRL)
scheme~\citep{shalev2012online} that is well known in the online
learning literature, with a least squares loss and a Mahalanobis norm
regularizer. The FTRL update is known to have desirable convergence
properties even in an adversarial setting, which makes it a popular
choice for online learning algorithms. An alternative interpretation
to the update in Eq.~\eqref{eq:level-1-theta-defn} is that \RS solves a
regularized least squares regression problem in an online setting---a
new data point is observed at every time $t,$ and $\theta_{t+1}$ is
the solution to the regularized least squares regression problem
framed with all the data available at time $t.$\footnote{The FTRL scheme is equivalent to online ridge regression in this chapter because uses preferences are assumed to be time-invariant. We consider time-varying user preferences in the subsequent chapters in which case the online ridge regression equivalence does not hold. For this reason, we prefer using the term FTRL instead of online linear regression throughout the thesis.}

Since the least squares loss on the right hand side of
Eq.~\eqref{eq:level-1-theta-defn} is strictly convex, there exists a
unique minimizer for the loss. Moreover, one can obtain a closed form
expression for $\theta_{t+1}$ by differentiating the loss and equating
it to zero, and that expression is given by
\begin{equation}
  \label{eq:level-1-learning-algo}
  \theta_{t+1} = \left( \sum_{s=1}^t w_{A_s} w_{A_s}^{\top} + \lambda WW^{\top} \right)^{-1} \left( \sum_{s=1}^{t}w_{A_s} R_s \right).
\end{equation}

% TODO: I can add a note here about the choice of the algorithm. This softmax algorithm does not excessively exploit like the UCB type algorithms do. however, one can change 'a' to control the level of exploration-exploitation tradeoff. Literature suggests that its potentially more useful to employ recommendation algorithms that explore more (to mitigate popularity bias, for example). Refer to work on incentivising exploration. Another reason for interest in such an algorithm is that the distribution of the algorithm is smooth, which makes it amenable for analysis (which will be discussed in the following section). Refer to softmax policy gradient/gradient bandit algorithms (Sutton and Barto, Section 2.8) , and also follow the regularized leader algorithm (online learning and OCO review).

\section{ODE Method of Stochastic Approximation}
\label{sec:ode-method}

Throughout this thesis, we are interested in understanding how the
states of the system of interest evolve with time. In this chapter, we
focus on understanding the long term properties of the time-varying
algorithm state in the presence of users whose preferences do not
change with time (we discuss time-varying user preferences in the
subsequent chapters). For this purpose, our analysis method of choice
is to use the ODE method of stochastic
approximation~\citep{borkar2009stochastic,prashanth2025gradient}.
There are multiple advantages to using this method of analysis. First,
it helps in understanding the asymptotic behavior of a stochastic
process (such as the one defined by \cref{eq:level-1-theta-defn,eq:level-1-learning-algo}) by analyzing a simpler deterministic dynamical system.
Second, this approach can be used to understand the asymptotic
properties of coupled systems; this is discussed at length in
subsequent chapters. Third, we can leverage well-developed tools from
dynamical systems theory in our analysis. These properties make this
method extremely suitable for asymptotic analysis provided that the
stochastic process of interest satisfies a specific kind of recursion.
We discuss the ODE method in the following.

Consider a sequence of $d$-dimensional vectors $(x_n)$ where the
discrete time $n \in \mathbb{Z}_+$ is a non-negative integer. Our goal
is to understand the properties of such a sequence given that each
element in the sequence satisfies the recursive relation
\begin{equation}
  \label{eq:ode-method-sa-recursion}
  x_{n+1} = x_n + a_n (h(x_n) + M_{n+1} + \beta_n).
\end{equation}
Here $x_0$ is assumed known, $h:\R^d \to \R^d$ is a function on $\R^d,$
$(a_n)$ is a sequence of scalars, and $(M_{n+1})$ and $(\beta_n)$ are
sequences on $\R^d$. We refer to the recursions of this type as \textit{stochastic approximation (SA) recursions.} Further, we make the following assumptions on the objects in the SA recursion.

\begin{assumption}[Square integrable step sizes]
  \label{A1}
  The ``stepsizes'' $(a_n)$ is a sequence of positive scalars
  satisfying $\sum_n a_n = \infty$ and $\sum_n a_n^2 < \infty.$
\end{assumption}

\begin{assumption}[Locally Lipschitz map]
  \label{A2}
  $h:\R^d \to \R^d$ is a locally Lipschitz map, i.e., for all
  $x_0 \in \R^d,$ there exist constants $\delta_0>0$ and $L>0$ such
  that
    \begin{equation*}
      \| x - x_0 \| < \delta_0 \implies \| h(x) - h(x_0) \| \leq L \|x - x_0\|.
    \end{equation*}
\end{assumption}

\begin{assumption}[Square integrable Martingale noise]
  \label{A3}
  $(M_n)$ is a martingale difference sequence with respect to the
  increasing family of $\sigma$-fields
  $\mathcal{F}_n := \sigma(x_m, M_m, m \leq n) = \sigma(x_0, M_1,
  \ldots M_n), n \geq 0$ i.e. for all $n \geq 0$
  \begin{equation*}
    E[M_{n+1}|\mathcal{F}_{n}] = 0 \quad a.s.
  \end{equation*}
  Moreover, $(M_n)$ are square-integrable with
  \begin{equation*}
    E[\| M_{n+1} \|^2 | \mathcal{F}_n] \leq K (1 + \| x_n \|^2) \quad a.s.
  \end{equation*}
  for all $n \geq 0$ for some constant $K > 0.$
\end{assumption}

\begin{assumption}[Bounded iterates]
  \label{A4}
  The iterates $(x_n)$ remain bounded a.s., i.e.,
  $\sup_n \|x_n\| < \infty$ a.s. for all $n \geq 0.$
\end{assumption}

\begin{assumption}[Asymptotically diminishing bias]
  \label{A5}
  The sequence $(\beta_n)$ converges to 0 almost surely, i.e.,
  $\lim_{n \to \infty} \beta_n = 0$ with probability 1.
\end{assumption}
These assumptions are reasonable for stochastic processes that follow recursions. Assumption 1 guarantees that the sequence of updates $(x_{n+1}-x_n)$ become smaller in magnitude over time, but not so small that the sum of norms of all updates is bounded. Assumption 2 guarantees that the update function is smooth, and assumption 3 bounds the second moment of the martingale noise, ensuring that it is attenuated over time. Assumption 4 ensures that the iterates $x_n$ do not diverge at any time, and assumption 5 removes the effect of bias terms on the asymptotic properties of $(x_n).$

Our interest here is in the long term behavior of the sequence
$(x_n).$ Specifically, we seek to obtain the limit set $L(x_n, n \geq 0)$ of the sequence $(x_n);$ this set contains the limit
points of all the convergent subsequences of $(x_n).$ Toward this, we
see that \cref{eq:ode-method-sa-recursion} is a stochastic approximation equation and the ODE method
of analyzing stochastic approximations prescribes a way to obtain the
limit set of $(x_n)$ \citep{borkar2009stochastic}. In this method, given \crefrange{A1}{A5},
the sequence $(x_n)$ is related to the trajectory $x(t)$ of the
continuous time ODE
\begin{equation}
  \label{eq:ode-method-asymptotic-ode}
  \dot{x}(t) = h(x(t))
  \end{equation}
with initial value $x(0) = x_0.$ Here $t \in \R$ determines the
continuous time variable of the ODE and is related to the discrete time
$n$ through the relation
\begin{equation*}
  t(n) = \sum_{m=1}^n a_m, \quad t(0) = 0
\end{equation*}
for all $n \in \N.$ We call \cref{eq:ode-method-asymptotic-ode} the \emph{asymptotic ODE} of $(x_n)$ because the mean squared error between
$x(t)$ and the continuous time sequence generated through the linear
interpolation of points in the sequence $(x_n)$ goes to 0 as
$t \to \infty.$ Qualitatively, one can interpret that the trajectory
$x(t)$ (with a given initial value $x(0)=x_0$) captures the mean
behavior of the ensemble of sequences $(x_n).$ The following result
\citep[Theorem~2.1]{borkar2009stochastic} formally establishes the
connection between $x_n$ and $x(t).$
\begin{lemma}[ODE method]
  Let $(x_n)$ satisfy the recurrence relation given by
  \cref{eq:ode-method-sa-recursion} along with
  \crefrange{A1}{A5}. Then, almost surely, the sequence $(x_n)$
  converges to its limit set, and that limit set is a (possibly sample
  path-dependent) connected internally chain transitive invariant set
  of the ODE given by \cref{eq:ode-method-asymptotic-ode}.
  \label{lemma:ode-method}
\end{lemma}

Using this result, one can obtain the limit set of $(x_n)$ by
analyzing the ODE in \cref{eq:ode-method-asymptotic-ode}. The set of
interest, according to the result, has 3 properties
this.
\begin{itemize}
\item \emph{Invariant}: A set $A$ is invariant if
  $x(0) \in A \implies x(t) \in A$ for all $t \in \R.$ That is, if the
  trajectory $x(t)$ is known to lie in an invariant set $A$ at some
  time $t,$ then the trajectory is contained in that set at all times.
\item \emph{Connected}: A set $A$ is connected if it cannot be
  expressed as the union of two or more disjoint non-empty subsets.
  Given a connected set $A,$ one can come up with a path (a sequence
  of points that are arbitrarily close to the neighboring points in
  the sequence) that begins and ends with any two points in $A,$ such
  that all points in the path belong to $A.$
\item \emph{Internally chain transitive}: A compact and invariant set
  $A$ is internally chain transitive if for any $x,y \in A$ and any
  $\epsilon>0,T>0,$ there exist $n+1$ points $x_0, x_1, \ldots,
  x_{n-1}, x_n=y$ in $A$ such that $\|x_0-x\|<\epsilon$ and the
  trajectory of \cref{eq:ode-method-asymptotic-ode} initiated at $x_i$
  lies in the $\epsilon$-neighborhood of $x_{i+1}$ for $i \in
  \{0,\ldots,n\}$ after a time $\geq T.$
\end{itemize}

%Given these assumptions, the ODE method of stochastic approximation
%gives us a continuous time trajectory $x(t),$ where $t \in \R_+,$
%that approximates the sequence $(x_n)$ in an expected sense. In
%particular, the mean squared error between $x(t)$ and the linear
%interpolation of $(x_n)$ goes to 0 as $t \to \infty.$

Connected internally chain transitive invariant sets often take
intuitive forms. Most common examples of such sets are equilibrium
sets and limit cycles. Equilibrium sets contain points for which the
time derivative given by the ODE in
\cref{eq:ode-method-asymptotic-ode} is 0. Such sets are invariant
because any trajectory starting from a point in the set stays at the
same point for all time, and they are observed to be chain transitive
by considering sequences of identical points in the equilibrium set.
Such equilibrium sets can be singleton sets, i.e., they are isolated
equilibria. Limit cycles are cyclic trajectories that repeatedly cover
the same path during some fixed time period. The following result
makes the connection between the limit sets of $(x_n)$ and isolated equilibrium points of the ODE, provided they exist.
\begin{lemma}
  \cite[Corollary~2.2]{borkar2009stochastic} If the only internally
  chain transitive sets for \cref{eq:ode-method-asymptotic-ode} are
  isolated equilibrium points, then $(x_n)$ converges a.s. to a
  possible sample path-dependent equilibrium point.
  \label{lemma:ode-method-isolated-equilibria}
\end{lemma}
This result is used throughout this thesis because we encounter ODEs for which there exist one or more isolated equilibria.
%Motivated by this result, we focus our attention on identifying such isolated equilibrium points for the asymptotic ODE.
In particular, we
are interested in \emph{asymptotically stable equilibria}---any
trajectory starting in a neighborhood around the equilibrium point
converges to the point as $t \to \infty.$ Such equilibria are useful because the sequence of interest, $(x_n),$ happens to converge to such equilibria (thanks to Lemma~\ref{lemma:ode-method-isolated-equilibria}), provided the sequence starts in the neighborhood of the equillibrium point. If the neighborhood is
arbitrarily small, then the equilibrium is said to be \emph{locally
  asymptotically stable}, and if it spans the entire space of points
that $x(t)$ can take, then the equilibrium is said to be
\emph{globally asymptotically stable}.

\section{Obtaining asymptotic ODE of the algorithm state}
\label{sec:level-1-asymptotic-ode}
% ODE TO ANALYZE evolution of THETA_T
%TOWARD THIS WE FIRST CAST 3.4 in the form of 3.5 and apply the lemmas
In this section, we use the ODE method of stochastic approximation to
analyze the long term behavior of the algorithm state $\theta_t.$
In particular, we show that the $\theta_t$ converges to the user
preference vector $\psi$ almost surely, by showing that $\psi$ is the
globally asymptotically stable equilibrium corresponding to the
asymptotic ODE of the recommendation algorithm. The first step in this process is to show that the dynamics of $\theta_t$ follows a recursive equation. Following that, the recursive equation in \cref{eq:learner-recursive-relation} is rewritten so that it is similar in form to the stochastic approximation equation given by \cref{eq:ode-method-sa-recursion}. Finally, after the stochastic approximation equation is obtained, the asymptotic ODE is described and analyzed.
\subsection{Obtaining recursions for the states of interest}
\label{ssec:level-1-obtaining-recursions}
We show that $\theta_{t+1}$ follows a recursive relation, i.e., it can be expressed as a function of $\theta_t.$
\begin{align*}
  \theta_t &= \left(\sum_{s=1}^{t-1} w_{A_s} w_{A_s}^{\top} + \lambda W W^{\top}\right)^{-1} \left( \sum_{s=1}^{t-1} w_{A_s} R_s \right) \\
  \left(\sum_{s=1}^{t-1} w_{A_s} w_{A_s}^{\top} + \lambda W W^{\top}\right) \theta_t &=  \left( \sum_{s=1}^{t-1} w_{A_s} R_s \right) \\
  \left(\sum_{s=1}^t w_{A_s} w_{A_s}^{\top}  + \lambda W W^{\top}\right) \theta_t - w_{A_t} w_{A_t}^{\top} \theta_t  &=  \left( \sum_{s=1}^t w_{A_s} R_s - w_{A_t} R_t \right) \\
  \left(\sum_{s=1}^t w_{A_s} w_{A_s}^{\top}  + \lambda W W^{\top}\right) \theta_t &=  \left( \sum_{s=1}^t w_{A_s} R_s \right) + \left( w_{A_t} w_{A_t}^{\top} \theta_t - w_{A_t} R_t \right) \\
  \theta_t &=  \theta_{t+1} +   \left(\sum_{s=1}^t w_{A_s} w_{A_s}^{\top}  + \lambda W W^{\top}\right)^{-1} \left( w_{A_t} w_{A_t}^{\top} \theta_t - w_{A_t} R_t \right) \\
  \theta_{t+1} - \theta_t &=   \left(\sum_{s=1}^t w_{A_s} w_{A_s}^{\top}  + \lambda W W^{\top}\right)^{-1} \left( w_{A_t} R_t - w_{A_t} w_{A_t}^{\top} \theta_t \right)
\end{align*}

In the preceding, the second-to-last step uses the definition of $\theta_{t+1}$ to obtain the first term on the RHS. To obtain a recursive equation similar to \cref{eq:ode-method-sa-recursion}, we must express the RHS of the above equation as a function of $\theta_t$ (similar to $h(x_t)$ in \cref{eq:ode-method-sa-recursion}). Most of the RHS can be expressed in terms of $\theta_t$ because $A_t$ depends on $\theta_t,$ and $R_t$ depends on $A_t,$ which in turn depends on $\theta_t.$ The only term we cannot express in terms of $\theta_t$ is $\sum_{s=1}^t w_{A_s} w_{A_s}^{\top};$ this is because $A_s$ depends on $\theta_s$ for $s<t.$
To get around this problem, we introduce an auxiliary algorithm state to eliminate  $\sum_{s=1}^t w_{A_s} w_{A_s}^{\top}$ from the recursive equation.
%recursive equation we obtained can be expressed as a function of $\theta_t$
Observe that
\begin{equation*}
  \sum_{s=1}^t w_{A_s} w_{A_s}^{\top} = \sum_{k=1}^K (\sum_{s=1}^t [A_s=k]) w_k w_k^{\top} = W
  \begin{bmatrix}
    \sum_{s=1}^t[A_s=1] & 0 & 0 \\
    0 & \ddots & 0 \\
    0 & 0 & \sum_{s=1}^t[A_s=K]\\
  \end{bmatrix}
  W^{\top},
\end{equation*}
where $[\cdot]$ is the Iverson bracket
\footnote{
  Note that the Iverson bracket is the generalization of the Kronecker delta function, i.e., given an expression $y,$ we have $[y]=1$ if $y$ is true, else $[y]=0.$
  % \begin{equation*}
  %   [y] =
  %   \begin{cases}
  %     1 & \text{if $y$ is true} \\
  %     0 & \text{else.} \\
  %   \end{cases}
  % \end{equation*}
}.
We define a vector $Q_t \in \R^K$ to be the auxiliary algorithm state, such that its $k$-th element is
\begin{equation}
  \label{eq:learner-q-defn}
  Q_{t}^k = \frac{1}{t} \left( \lambda + \sum_{s=1}^{t-1} [A_s = k] \right).
\end{equation}
The vector $Q_t$ captures the recommendation history until time $t-1.$ Using this definition, we can replace the term $\sum_{s=1}^t w_{A_s} w_{A_s}^{\top}$ with $Q_{t+1}$ because
\begin{align*}
  \sum_{s=1}^t w_{A_s} w_{A_s}^{\top}  + \lambda W W^{\top} &= \sum_{k=1}^K (\lambda + \sum_{s=1}^t [A_s = k]) w_k w_k^{\top} \\
  &= \sum_{k=1}^K (t+1) Q_{t+1}^k w_k w_k^{\top} = (t+1) W \diag(Q_{t+1}) W^{\top}.
\end{align*}
This gives us the following recursive relation for $\theta_t$
\begin{equation}
  \label{eq:learner-recursive-relation}
  \theta_{t+1} = \theta_t + \frac{1}{t+1} \left(\sum_{k=1}^K Q_{t+1}^k w_k w_k^{\top} \right)^{-1} \left( w_{A_t} R_t - w_{A_t} w_{A_t}^{\top} \theta_t \right).
\end{equation}
Because $Q_{t+1} = \frac{t}{t+1} Q_t + \frac{1}{t+1}[A_s=t],$ it is easy to see that the RHS of \cref{eq:learner-recursive-relation} can be expressed as a function of $\theta_t$ and $Q_t.$
Since $Q_t$ is used in the expression for $\theta_t,$ this variable is also treated like a algorithm state. We can now use stochastic approximation theory provided that we treat $(\theta_t,Q_t)$ as the sequence of interest. To proceed, we must obtain a recursive relation for $Q_t$ as well. From the definition of $Q_t$ in \cref{eq:learner-q-defn}, we have
\begin{align}
   t Q_t^k &= \lambda + \sum_{s=1}^{t-1} [A_s = k] \nonumber\\
  t Q_t^k + [A_{t} = k] &= \lambda + \sum_{s=1}^t [A_s = k]  \nonumber\\
  \frac{t}{t+1} Q_t^k + \frac{1}{t+1}[A_{t} = k] &= \frac{1}{t+1} \left( \lambda + \sum_{s=1}^t [A_s = k] \right) = Q_{t+1}^k \nonumber\\
  Q_{t+1}^k &= Q_t^k + \frac{1}{t+1} \left( [A_{t} = k] - Q_t^k \right)   \label{eq:learner-q-recursive-relation}
\end{align}
In the preceding, the second-to-last step uses the definition of $Q_{t+1}.$ Considering $(\theta_t,Q_t)$ as the full algorithm state, the \cref{eq:learner-recursive-relation,eq:learner-q-recursive-relation} together form the desired recursive relation

\subsection{Rewriting the obtained recursions as SA recursions}
\label{ssec:level-1-stochastic-approximation-recursions}
We now rewrite the \cref{eq:learner-recursive-relation,eq:learner-q-recursive-relation} such that they take the form of the stochastic approximation equation given in \cref{eq:ode-method-sa-recursion}. First, we find the function that appears in the asymptotic ODE (for the SA recursion given by \cref{eq:ode-method-sa-recursion}, this corresponds to the function $h$). We obtain such a function for $\theta_t,$ denoted by $h_\theta,$ by analyzing the term $(t+1)\Exp{\theta_{t+1}-\theta_t | \theta_t, Q_t}.$ To understand why, observe that \cref{eq:learner-recursive-relation} is rewritten as
\begin{equation*}
    (t+1) (\theta_{t+1}-\theta_t) = \frac{\theta_{t+1} - \theta_t}{\frac{1}{t+1}} = \left(\sum_{k=1}^K Q_{t+1}^k w_k w_k^{\top} \right)^{-1} \left( w_{A_t} R_t - w_{A_t} w_{A_t}^{\top} \theta_t \right).
\end{equation*}
The term $(t+1) (\theta_{t+1} - \theta_t)$ captures the deviation of $\theta_t.$  We can decompose this term as the sum of the expected deviation $(t+1) \Exp{\theta_{t+1}-\theta_t | \theta_t, Q_t},$ given the current value of $\theta_t$ and $Q_t,$ and noise  $(t+1) (\theta_{t+1} - \theta_t) - (t+1) \Exp{\theta_{t+1}-\theta_t | \theta_t, Q_t}.$ We use this intuition as motivation to analyze  $(t+1) \Exp{\theta_{t+1}-\theta_t | \theta_t, Q_t},$ and later in this section, we establish the relation between this term and the asymptotic ODE. Our goal now is to express  $(t+1) \Exp{\theta_{t+1}-\theta_t | \theta_t, Q_t}$ as a function of $\theta_t$ and $Q_t.$  From the recursive relation of $\theta_t$ given by \cref{eq:learner-recursive-relation}, we get
%Later in this section, we will show that $(t+1) \Exp{\theta_{t+1}-\theta_t | \theta_t, Q_t}$ corresponds to the function $h(\theta_t,Q_t)$ and the noise term corresponds to $M_{t+1}.$
%We calculate the mean ODE for the recursive relations given by Equations \eqref{eq:learner-recursive-relation} and \eqref{eq:learner-q-recursive-relation}.
%
\begin{align}
  &\frac{\Exp{\theta_{t+1} - \theta_t|\theta_t, Q_t}}{\frac{1}{t+1}}\nonumber\\
  &= \Exp{ \left(\sum_{k=1}^K Q_{t+1}^k w_k w_k^{\top} \right)^{-1} \left( w_{A_t} R_t - w_{A_t} w_{A_t}^{\top} \theta_t \right) \Bigg| \theta_t, Q_t} \nonumber\\
  &= \Exp{ \left(\sum_{k=1}^K \left( Q_t^k  +\frac{[A_t=k] - Q_t^k}{t+1} \right) w_k w_k^{\top} \right)^{-1} \left( w_{A_t} R_t - w_{A_t} w_{A_t}^{\top} \theta_t \right) \Bigg| \theta_t, Q_t} \nonumber\\
  &= \sum_{j=1}^K P(A_t=j) \left(\sum_{k=1}^K \left( \frac{t Q_t^k}{t+1}  +\frac{\delta_{jk}}{t+1} \right) w_k w_k^{\top} \right)^{-1} \left( w_j (w_j^{\top} \psi) - w_jw_j^{\top} \theta_t \right) \nonumber\\
  &= \sum_{j=1}^K P(A_t=j) \left(  \frac{t}{t+1} W \diag(Q_t) W^{\top}  +\frac{1}{t+1} w_j w_j^{\top} \right)^{-1} \left( w_j (w_j^{\top} \psi) - w_jw_j^{\top} \theta_t \right) \label{auxeq:1}
\end{align}
To simplify the inverse term on the RHS further, we use the following lemma from the work of \cite{miller1981inverse}.
\begin{lemma}
  If $A$ and $A+B$ are invertible and $B$ has rank 1, then let $g=\text{trace}(B A^{-1}).$ Then $g \neq -1$ and
  \begin{equation*}
    (A+B)^{-1} = A^{-1} - \frac{1}{1+g} A^{-1} B A^{-1}
  \end{equation*}
  \label{lemma:matrix-sum-inverse}
\end{lemma}
Choosing $A=\frac{t}{t+1}W \diag(Q_t) W^{\top}$ and $B=\frac{w_jw_j^{\top}}{t+1},$ we can see that $A$ is invertible and $B$ is rank 1. Moreover, $A+B = \frac{1}{t+1} W (\diag(Q_t + e_j)) W^{\top}$ is also invertible. As all the necessary conditions are satisfied, Lemma~\ref{lemma:matrix-sum-inverse} can be used. To apply the lemma, we must calculate $g=\text{trace}(B A^{-1}),$ i.e.,
\begin{align*}
  g &=  \trace \left( \frac{w_jw_j^{\top}}{t+1} \left( \frac{t}{t+1} W \diag(Q_t) W^{\top} \right)^{-1} \right) \\
    &= \frac{1}{t} \trace \left( w_j w_j^{\top} \left( W \diag(Q_t) W^{\top} \right)^{-1} \right) \\
    &= \frac{1}{t} \trace \left( w_j^{\top} \left( W \diag(Q_t) W^{\top} \right)^{-1} w_j \right) \\
    &= \frac{1}{t} \left( w_j^{\top} \left( W \diag(Q_t) W^{\top} \right)^{-1} w_j \right)
\end{align*}
From this, we know that $g \geq 0$ because $ \left( W \diag(Q_t) W^{\top} \right)^{-1}$ is positive definite. We now apply Lemma~\ref{lemma:matrix-sum-inverse} to get
\begin{align*}
  (A+B)^{-1} &= A^{-1} - \frac{1}{1+g} A^{-1} B A^{-1} \\
             &= A^{-1} ( I - \frac{1}{1+g} B A^{-1} ) \\
             &= \frac{t+1}{t} \left( W \diag(Q_t) W^{\top} \right)^{-1} \left( I - \frac{(t+1)}{(1+g)t(t+1)} w_j w_j^{\top} \left( W \diag(Q_t) W^{\top} \right)^{-1} \right)\\
             &= \left( 1 + \frac{1}{t} \right) (W \diag(Q_t) W^{\top})^{-1} \left( I - \frac{1}{t (1+g)} w_j w_j^{\top} (W \diag(Q_t) W^{\top} )^{-1} \right) 
\end{align*}
This gives us, for all $j \in [K],$
\begin{equation}
  \label{eq:inverse-term-breakdown}
  \left(  \frac{t}{t+1} W \diag(Q_t) W^{\top}  +\frac{1}{t+1} w_j w_j^{\top} \right)^{-1} = (W \diag(Q_t) W^{\top})^{-1} + \frac{1}{t} C_t^j.
\end{equation}
Here $C_t^j := (W \diag(Q_t) W^{\top})^{-1} \left( I - \left( 1 + \frac{1}{t} \right) \frac{w_jw_j^{\top}}{1+g} (W \diag(Q_t) W^{\top})^{-1}\right).$ Using this, we resume analyzing our expression of interest from \cref{auxeq:1}
\begin{align}
  &\frac{\Exp{\theta_{t+1} - \theta_t|\theta_t, Q_t}}{\frac{1}{t+1}}\nonumber\\
  &= \sum_{j=1}^K P(A_t=j) \left(  \frac{t}{t+1} W \diag(Q_t) W^{\top}  +\frac{1}{t+1} w_j w_j^{\top} \right)^{-1} \left( w_j (w_j^{\top} \psi) - w_jw_j^{\top} \theta_t \right) \nonumber\\
  &= \sum_{j=1}^K P(A_t=j) \left( (W \diag(Q_t) W^{\top})^{-1} + \frac{C_t^j}{t} \right) \left( w_j (w_j^{\top} \psi) - w_jw_j^{\top} \theta_t \right) \nonumber\\
  &= \sum_{j=1}^K P(A_t=j) (W \diag(Q_t) W^{\top})^{-1} \left( w_j (w_j^{\top} \psi) - w_jw_j^{\top} \theta_t \right)\nonumber\\
  &\quad + \sum_{j=1}^K P(A_t=j) \frac{C_t^j}{t}  \left( w_j (w_j^{\top} \psi) - w_jw_j^{\top} \theta_t \right) \nonumber\\
  &= (W \diag(Q_t) W^{\top})^{-1} W \diag{\pi(\theta_t)} W^{\top} (\psi - \theta_t) + \frac{1}{t} W \diag(\pi(\theta_t) \odot C_t) W^{\top} (\psi - \theta_t) \label{eq:expectation-breakdown}
\end{align}
where $C_t :=
\begin{bmatrix}
  C_t^1 & C_t^2 & \cdots & C_t^K
\end{bmatrix}^{\top}
$ and  $\pi(\theta) :=
\begin{bmatrix}
  \pi_1(\theta) & \pi_2(\theta) & \cdots & \pi_K(\theta)
\end{bmatrix}^{\top}
$ are $K$-dimensional vectors. $\pi(\theta_t)$ contains the probabilities of recommending all available items at time $t.$
It can be shown that the second term converges to 0 a.s. as $t \to \infty,$ and hence it does not contribute to the asymptotic ODE. Only the first term is related to the mean ODE.
We define $h_{\theta}(\theta,Q)$ as
% \begin{align*}
  %   h_{\theta}(\theta,Q) := \sum_{j=1}^K \pi_j(\theta) (W \diag(Q) W^{\top})^{-1} \left( w_j (w_j^{\top} \psi) - w_jw_j^{\top} \theta \right)
  % \end{align*}
  %   The above expression can be rewritten using matrix notation
\begin{equation}
  \label{eq:learner-mean-ode}
  h_{\theta}(\theta,Q) := (W \diag(Q) W^{\top})^{-1} W \diag(\pi(\theta)) W^{\top} (\psi - \theta)
\end{equation}
where $\pi(\theta) =
\begin{bmatrix}
  \pi_1(\theta) & \pi_2(\theta) & \cdots & \pi_K(\theta)
\end{bmatrix}^{\top}
$ is a $K$-dimensional vector containing the probabilities of recommending the available items at time $t.$ We use $h_{\theta}$ to express the asymptotic ODE of $\theta_t.$

We now switch attention to the asymptotic ODE of $Q_t.$
\begin{align*}
  \frac{E[Q^k_{t+1} - Q^k_t|\theta_t, Q_t]}{\frac{1}{t+1}} &= E[[A_{t}=k] - Q_t^k | \theta_t, Q_t] \\
                                                       &= E[[A_{t}=k] | \theta_t, Q_t] - Q_t^k \\
                                                       &= P(A_{t} = k) - Q_t^k 
\end{align*}
Since the above is applicable for all $k \in [K],$ it is straightforward to extend the relation to the entire vector $Q_t.$ We define the function $h_q$ to characterize the mean ODE of $Q_t.$
\begin{equation}
  \label{eq:learner-q-mean-ode}
  h_q(\theta, Q) := \pi(\theta) - Q
\end{equation}
Now that we have obtained the functions $h_{\theta}$ and $h_q,$ we are ready to write the stochastic approximation equations for $\theta_t$ and $Q_t.$ 

The following shows that Eqs.~\eqref{eq:learner-mean-ode} and \eqref{eq:learner-q-mean-ode} do indeed represent the asymptotic ODEs of $\theta_t$ and $Q_t$ respectively.   Define $\mathcal{F}_t := \sigma(\theta_s,Q_s,s \leq t)$ as the $\sigma$-field that contains all possible events upto and until time $t$. Using Equations~\eqref{eq:inverse-term-breakdown}, \eqref{eq:expectation-breakdown} and \eqref{eq:learner-mean-ode}, we get
  \begin{align*}
    \frac{\theta_{t+1}-\theta_t}{\frac{1}{t+1}} &= \left( \frac{\theta_{t+1}-\theta_t}{\frac{1}{t+1}} - \Exp{\frac{\theta_{t+1}-\theta_t}{\frac{1}{t+1}} \Bigg| \mathcal{F}_t} \right) +  \Exp{\frac{\theta_{t+1}-\theta_t}{\frac{1}{t+1}} \Bigg| \mathcal{F}_t} \\
                                                &= M_{t+1} + h_{\theta}(\theta_t,Q_t) + \frac{1}{t} W \diag(\pi(\theta_t \odot C_t)) W^{\top} (\psi - \theta_t) \\
                                                &= M_{t+1} + h_{\theta}(\theta_t,Q_t) + \gamma_{t+1}
  \end{align*}
  where $M_{t+1} := (t+1) (\theta_{t+1} - \theta_t - \Exp{\theta_{t+1} - \theta_t | \mathcal{F}_t})$ and $\gamma_t := \frac{1}{t} W \diag(\pi(\theta_t \odot C_t)) W^{\top} (\psi - \theta_t).$ This lets us rewrite Eq.~\eqref{eq:learner-recursive-relation} to obtain the stochastic approximation equation for $\theta_t$ as
  \begin{equation}
    \theta_{t+1} = \theta_t + \frac{1}{t+1} \left( h_{\theta}(\theta_t,Q_t) + M_{t+1} + \gamma_{t+1} \right).
    \label{eq:learner-stochastic-approximation-eqn}
  \end{equation}
  Using a similar procedure for $Q_t$ gives us the stochastic approximation equation for $Q_t,$ i.e.,
  \begin{equation}
    Q_{t+1} = Q_t + \frac{1}{t+1} (h_q(\theta_t,q_t) + M'_{t+1})
    \label{eq:learner-q-stochastic-approximation-eqn}
  \end{equation}
  where $M'_{t+1} := (t+1) (Q_{t+1}-Q_t - \Exp{Q_{t+1}-Q_t| \mathcal{F}_t}).$ 
\subsection{Obtaining the Asymptotic ODE from SA recursions}
Using \cref{eq:learner-stochastic-approximation-eqn,eq:learner-q-stochastic-approximation-eqn}, we apply the ODE method of stochastic approximation (Lemma~\ref{lemma:ode-method}) to obtain the asymptotic ODE of the algorithm states.
\begin{theorem}
  The sequence of algorithm states $(\theta_t, Q_t)$ converges to a connected internally chain recurrent set of the ODE
  \begin{equation}
    \begin{split}
      \dot{\theta}(\tau) &= h_{\theta}(\theta,q) \\
      \dot{q}(\tau) &= h_{q}(\theta,q)
    \end{split}
    \label{eq:level-1-ode}
  \end{equation}
\label{thm:level-1-ode}
\end{theorem}
\begin{proof}
  %The proof technique is to treat $(\theta_t,Q_t)$ as a single algorithm state and apply the ODE method on that.
  The idea of the proof is to use the ODE method discussed in \cref{sec:ode-method}. Concatenating both \cref{eq:learner-stochastic-approximation-eqn,eq:learner-q-stochastic-approximation-eqn} gives us the following recursive relation:
 \begin{equation*}
    \begin{bmatrix}
      \theta_{t+1} \\ Q_{t+1}
    \end{bmatrix} =
    \begin{bmatrix}
      \theta_t \\ Q_t
    \end{bmatrix} + \frac{1}{t+1} \left(
      \begin{bmatrix}
        h_{\theta}(\theta_t,Q_t) \\ h_q(\theta_t,Q_t)
      \end{bmatrix} +
      \begin{bmatrix}
        M_{t+1} \\ M'_{t+1}
      \end{bmatrix} +
      \begin{bmatrix}
        \gamma_{t+1} \\ 0
      \end{bmatrix}
    \right)
  \end{equation*}
  The rest of the proof shows that the above recursive relation satisfies the assumptions needed for the ODE method.\\

  \textbf{A1.} The stepsize sequence $(\frac{1}{t})$ satisfy $\sum_{t=1}^{\infty} \frac{1}{t} = \infty$ and $\sum_{t=1}^{\infty} \frac{1}{t^2} = \frac{\pi^2}{6} < \infty.$\\

  \textbf{A2.} The function $h =
  \begin{bmatrix}
    h_{\theta} \\ h_q
  \end{bmatrix}
  $ is locally Lipschitz if $h$ is a continuously differentiable function i.e. all partial derivatives of $h$ exist and are continuous. %TODO: show this in the appendix
  In the following, we show that each component of $h$ is continuously differentiable.
  Recall that
  \begin{align*}
    h_{\theta}(\theta,q) &= (W \diag(q) W^{\top})^{-1} W \diag(\pi(\theta)) W^{\top} (\psi - \theta)\\
    \implies \nabla_{\theta}h_{\theta}(\theta,q) &= (W \diag(q) W^{\top})^{-1} W \left( -\diag(\pi(\theta)) W^{\top} + \diag(W^{\top} (\psi - \theta)) \nabla_{\theta} \pi(\theta) \right)
  \end{align*}
  Since the derivative of softmax policy $\pi(\theta)$ with respect to $\theta$ is continuous, %TODO: Show this if needed
  $h_{\theta}$ is continuously differentiable with respect to $\theta.$ Now, we analyze the gradient with respect to $q.$
  \begin{equation*}
    \frac{\partial}{\partial q_k}  h_\theta(\theta,q) = - (W \diag(q) W^{\top})^{-1} w_k w_k^{\top} (W \diag(q) W^{\top})^{-1} \diag(\pi(\theta)) W^{\top} (\psi - \theta)^{\top}
  \end{equation*} %TODO: Check this
  Since this is also a continuous function, we see that $h_{\theta}$ is a continously differentiable function. Now we analyze $h_q.$ Recall that
  \begin{equation*}
    h_q(\theta, Q) = \pi(\theta) - Q
  \end{equation*}
  From this, we get
  \begin{equation*}
    \nabla_{\theta} h_q = \nabla_{\theta} \pi(\theta); \quad \nabla_q h_q = -I
  \end{equation*}
  Since softmax function in \cref{eq:level-1-rec-policy} is a continuously differentiable function with respect to $\theta,$ we also observe that $h_{q}$ is continuously differentiable. Hence, $h$ is locally Lipschitz.\\
  
  \textbf{A3.} Recall that $M_{t+1} = (t+1) (\theta_{t+1} - \theta_t - \Exp{\theta_{t+1} - \theta_t | \mathcal{F}_t}).$ It is straightforward to check that $M_{t+1}$ is a martingale difference sequence; see below.
  \begin{align*}
    \Exp{M_{t+1} | \mathcal{F}_t} &= \Exp{(t+1) (\theta_{t+1} - \theta_t - \Exp{\theta_{t+1} - \theta_t | \mathcal{F}_t}) | \mathcal{F}_t} \\
    &= (t+1) \left( \Exp{ \theta_{t+1} - \theta_t | \mathcal{F}_t} - \Exp{\theta_{t+1} - \theta_t | \mathcal{F}_t} \right) = 0.
  \end{align*}
  $M'_{t+1}$ is shown to be a martingale difference sequence using a similar argument. Now we check square integrability of $M_{t+1}.$
  \begin{align*}
    &\Exp{\|M_{t+1}\|^2 | \mathcal{F}_t} \\
    &= \Exp{ \left\|
      \begin{matrix}
        \left( W \diag{Q_{t+1}} W^{\top} \right)^{-1} \left( w_{A_t} R_t - w_{A_t} w_{A_t}^{\top} \theta_t \right) \\
        - \Exp{ \left( W \diag{Q_{t+1}} W^{\top} \right)^{-1} \left( w_{A_t} R_t - w_{A_t} w_{A_t}^{\top} \theta_t \right) \Big| \mathcal{F}_t }
      \end{matrix}
      \right\|^2 \Bigg| \mathcal{F}_t} \\
    &= \Exp{\left\| \left( W \diag{Q_{t+1}} W^{\top} \right)^{-1} \left( w_{A_t} R_t - w_{A_t} w_{A_t}^{\top} \theta_t \right) \right\|^2 \Bigg| \mathcal{F}_t} \\
      & \quad- \left\| \Exp{ \left( W \diag{Q_{t+1}} W^{\top} \right)^{-1} \left( w_{A_t} R_t - w_{A_t} w_{A_t}^{\top} \theta_t \right) \Bigg| \mathcal{F}_t} \right\|^2 \\
    &\leq  \Exp{\left\| \left( W \diag{Q_{t+1}} W^{\top} \right)^{-1} \left( w_{A_t} R_t - w_{A_t} w_{A_t}^{\top} \theta_t \right) \right\|^2 \Bigg| \mathcal{F}_t} \\
    &= \sum_{k=1}^K \pi_k(\theta_t) \Exp{\left\| \left( W \diag \left(\frac{t Q_t}{t+1} + \frac{e_k}{t+1} \right) W^{\top} \right)^{-1} \left( w_{A_t} R_t - w_{A_t} w_{A_t}^{\top} \theta_t \right) \right\|^2 \Bigg| \mathcal{F}_t, A_t=k}
  \end{align*}
  where $e_k \in \R^K$ is 1 at the $k$-th element and 0 everywhere else. Observe that the above sum is a convex combination of $K$ terms, and is upper bounded by the largest of those terms. \\ Let $k^{*} := \arg \min_{k \in [K]} \Exp{\left\| \left( W \diag{Q_{t+1}} W^{\top} \right)^{-1} \left( w_{A_t} R_t - w_{A_t} w_{A_t}^{\top} \theta_t \right) \right\|^2 \Bigg| \mathcal{F}_t, A_t=k}$ and let $Q^{*}_{t+1} := \frac{t Q_t}{t+1} + \frac{e_k}{t+1}.$ Using this, we obtain another upper bound:
  \begin{align*}
    &\sum_{k=1}^K \pi_k(\theta_t) \Exp{\left\| \left( W \diag \left(\frac{t Q_t}{t+1} + \frac{e_k}{t+1} \right) W^{\top} \right)^{-1} \left( w_{A_t} R_t - w_{A_t} w_{A_t}^{\top} \theta_t \right) \right\|^2 \Bigg| \mathcal{F}_t, A_t=k} \\
    &\leq \Exp{\left\| \left( W \diag(Q^{*}_{t+1}) W^{\top} \right)^{-1} \left( w_{k} R_t - w_{k} w_{k}^{\top} \theta_t \right) \right\|^2 \Bigg| \mathcal{F}_t, A_{t+1}=k}
    %&\leq \Exp{\left\| \left( W \diag(Q^{*}_{t+1}) W^{\top} \right)^{-1} \left( w_{k} (w_k^{\top}\psi + \epsilon_t) - w_{k} w_{k}^{\top} \theta_t \right) \right\|^2 \Bigg| \mathcal{F}_t, A_t=k}
  \end{align*}
  Recall that $R_t = w_{A_t}^{\top} \psi + \epsilon_t.$ We use this to further break down the upper bound.
  \begin{align*}
    &\Exp{\left\| \left( W \diag(Q^{*}_{t+1}) W^{\top} \right)^{-1} \left( w_{k} R_t - w_{k} w_{k}^{\top} \theta_t \right) \right\|^2 \Bigg| \mathcal{F}_t, A_{t+1}=k}\\
    &=\Exp{\left\| \left( W \diag(Q^{*}_{t+1}) W^{\top} \right)^{-1} \left( w_{k} (w_k^{\top} \psi + \epsilon_t) - w_{k} w_{k}^{\top} \theta_t \right) \right\|^2 \Bigg| \mathcal{F}_t, A_{t+1}=k} \\
    &=\Exp{\left\| \left( W \diag(Q^{*}_{t+1}) W^{\top} \right)^{-1} \left( w_{k} w_k^{\top} (\psi - \theta_t) \right) \right\|^2 \Bigg| \mathcal{F}_t, A_{t+1}=k} \\
    &\quad + \Exp{\left\| \left( W \diag(Q^{*}_{t+1}) W^{\top} \right)^{-1} w_{k} \epsilon_t \right\|^2 \Bigg| \mathcal{F}_t, A_{t+1}=k} \\
    &\quad + \Exp{ \dotp{ \left( W \diag(Q^{*}_{t+1}) W^{\top} \right)^{-1} \left( w_{k} w_k^{\top} (\psi - \theta_t) \right), \left( W \diag(Q^{*}_{t+1}) W^{\top} \right)^{-1} w_{k} \epsilon_t} \Bigg| \mathcal{F}_t, A_{t+1}=k}
  \end{align*}
  Observe that the only random variable in the above expression is $\epsilon_t.$ The third term in the above expression is 0 because $E[\epsilon_t] = 0,$ and the second term is bounded because $E[\epsilon_t^2] < \infty.$ We simplify the first two terms in the above expression further:
  \begin{align*}
    &\Exp{\left\| \left( W \diag(Q^{*}_{t+1}) W^{\top} \right)^{-1} \left( w_{k} R_t - w_{k} w_{k}^{\top} \theta_t \right) \right\|^2 \Bigg| \mathcal{F}_t, A_{t+1}=k}\\
    &= \left\| \left( W \diag(Q^{*}_{t+1}) W^{\top} \right)^{-1} \left( w_{k} w_k^{\top} (\psi - \theta_t) \right) \right\|^2 + \left\| \left( W \diag(Q^{*}_{t+1}) W^{\top} \right)^{-1} w_{k} \right\|^2 \sigma^2 \\
    &= \left\| \left( W \diag(Q^{*}_{t+1}) W^{\top} \right)^{-1} w_{k} \right\|^2 \left( \left( w_k^{\top} (\psi - \theta_t) \right)^2 + \sigma^2 \right) \\
    &\leq \left\| \left( W \diag(Q^{*}_{t+1}) W^{\top} \right)^{-1} w_{k} \right\|^2 \|w_k\|^2 \left( \left\| \psi - \theta_t \right\|^2 + \frac{\sigma^2}{\|w_k\|^2} \right) \\
    &\leq \left\| \left( W \diag(Q^{*}_{t+1}) W^{\top} \right)^{-1} w_{k} \right\|^2 \|w_k\|^2 \left( 2\| \psi \|^2 + 2\| \theta_t \|^2 + \frac{\sigma^2}{\|w_k\|^2} \right) \\
    &\leq 2 \left\| \left( W \diag(Q^{*}_{t+1}) W^{\top} \right)^{-1} w_{k} \right\|^2 \|w_k\|^2 \left( \| \psi \|^2  + \frac{\sigma^2}{2\|w_k\|^2} + \| \theta_t \|^2 \right) \\
    &\leq 2 \left\| \left( W \diag(Q^{*}_{t+1}) W^{\top} \right)^{-1} w_{k} \right\|^2 \|w_k\|^2 \max \left\{ \|\psi\|^2 + \frac{\sigma^2}{2 \|w_k\|^2}, 1 \right\} (1 + \|\theta_t\|^2)
  \end{align*}
  Hence, $\Exp{\|M_{t+1}\|^2 | \mathcal{F}_t} \leq C (1 + \|\theta_t\|^2)$, where $C$ is given above. We now check the square integrability of $M'_{t+1}.$ First, we rewrite Eq.~\eqref{eq:learner-q-recursive-relation} in vector notation:
  \begin{equation*}
    Q_{t+1} = Q_t + \frac{1}{t+1} \left( e_{A_t} - Q_t \right)
  \end{equation*}
  We can use the above expression to compute $\Exp{\|M'_{t+1}\|^2 | \mathcal{F}_t}.$
  \begin{align*}
    \Exp{\|M'_{t+1}\|^2 | \mathcal{F}_t} &= \Exp{ \left\| e_{A_t} - Q_t - \Exp{ e_{A_t} - Q_t | \mathcal{F}_t} \right\|^2 | \mathcal{F}_t} \\
                                        &=  \Exp{ \| e_{A_t} - Q_t \|^2  | \mathcal{F}_t} - \left\| \Exp{ e_{A_t} - Q_t | \mathcal{F}_t} \right\|^2 \\
    &\leq \Exp{ \| e_{A_t} - Q_t \|^2  | \mathcal{F}_t} \\
    &\leq \sum_{k=1}^K \pi_k(\theta_t) \| e_k - Q_t \|^2 \\
    &\leq \max_{k \in [K]} \| e_k - Q_t \|^2 \\
    &= \max_{k \in [K]} \| e_k \|^2 + \| Q_t \|^2 - 2 \dotp{e_k,Q_t} \\
    &\leq \max_{k \in [K]} 2\| e_k \|^2 + 2\| Q_t \|^2 \\
    &\leq 2 (1 + \| Q_t \|^2)
  \end{align*}
  because $\| e_k \|^2 = 1$ for all $k \in [K]$ by definition. This proves that $M'_{t+1}$ is square integrable as well. \\

  \textbf{A4.} (Boundedness of iterates). From the definition of $\theta_t,$ we get, for all $t > 0,$
  \begin{align*}
    \| \theta_{t+1} \| &= \left\| \left( W \diag{Q_{t+1}} W^{\top} \right)^{-1} \left( \frac{1}{t} \sum_{s=1}^{t}w_{A_s} R_s \right) \right\| \\
    &\leq \left\| \left( W \diag{Q_{t+1}} W^{\top} \right)^{-1} \right\| \left\| \left( \frac{1}{t} \sum_{s=1}^{t}w_{A_s} (w_{A_s}^{\top} \psi + \epsilon_t) \right) \right\| \\
    &\leq \max_{s \in [T]} \left\| \left( W \diag{Q_{t+1}} W^{\top} \right)^{-1} \right\| \left\| \left( w_{A_s} (w_{A_s}^{\top} \psi + \epsilon_t) \right) \right\| \\
    &\leq \max_{s \in [T]} \left\| \left( W \diag{Q_{t+1}} W^{\top} \right)^{-1} \right\| \left\| w_{A_s} \right\| \left( \|w_{A_s}\| \|\psi\| + \epsilon_t \right)
  \end{align*}
  The RHS is finite almost surely because $ \left\| \left( W \diag{Q_{t+1}} W^{\top} \right)^{-1} \right\|$ is bounded for all $t>0,$ $\|w_k\|$ is bounded for all $k \in [K],$ $\| \psi \|$ is bounded and $\epsilon_t$ is real valued (so $P(\epsilon_t = \infty) = 0$).

  Now, we analyze the other iterate $Q_t.$ Using vector notation along with the definition of $Q_t$ from Eq.~\eqref{eq:learner-q-defn}:
  \begin{align*}
    \| Q_{t+1} \| &= \left\| \frac{1}{t} \left( \lambda \mathbf{1}_K + \sum_{s=1}^t e_{A_s} \right) \right\| \\
    &\leq \frac{\lambda}{t} \left\| \mathbf{1}_K \right\| +  \frac{1}{t} \left\| \sum_{s=1}^t e_{A_s} \right\| \\
    &\leq \frac{\lambda}{t} \sqrt{K} +  \frac{1}{t} \sum_{s=1}^t \left\| e_{A_s} \right\|   
    \leq \frac{\lambda}{t} \sqrt{K} + 1
  \end{align*}
  The above upper bound is finite for all $t>0.$ \\
  
  \textbf{A5.} (Almost sure convergence of $\gamma_t \to 0$) Define $\tilde{\gamma}_t$ such that
  \begin{align*}
    \tilde{\gamma}_t &:=  W \diag(\pi(\theta_t \odot C_t)) W^{\top} (\psi - \theta_t) \\
    \implies \| \tilde{\gamma}_t \|_2 &\leq \| W \diag(\pi(\theta_t \odot C_t)) W^{\top} \| (\| \psi \| + \| \theta_t \|)
  \end{align*}
  Observe that $\|\tilde{\gamma}_t\|_2 < \infty$ a.s. because $\| W \diag(\pi(\theta_t \odot C_t)) W^{\top}\|$ and $\|\psi\|$ are bounded, and it was shown that $\|\theta_t\|$ is bounded almost surely. This implies $\| \tilde{\gamma}_t \|_{\infty} < \infty$ a.s. From this, we conclude $\gamma_t = \frac{\tilde{\gamma}_t}{t} \to 0$ a.s.
\end{proof}

\section{Analyzing the equilibria of the asymptotic ODE}
  %   TODO: Elaborate this section. Replace the current proof with the new one that does not use the invariance principle.

The asymptotic ODE of the system, as obtained from \cref{thm:level-1-ode}, can be written using the expressions of $h_{\theta}$ and $h_q$ from \cref{eq:learner-mean-ode,eq:learner-q-mean-ode} respectively.
\begin{align*}
  \dot{\theta}(\tau) &= (W \diag(q(\tau)) W^{\top})^{-1} W \diag(\pi(\theta(\tau))) W^{\top} (\psi - \theta(\tau)) \\
  \dot{q}(\tau) &= \pi(\theta(\tau)) - q(\tau)
\end{align*}
To find the equilibrium points of this ODE, we set $\dot{\theta}(\tau)=0$ and $\dot{q}(\tau)=0.$ Then, the equilibrium points $\bar{\theta}$ and $\bar{q}$ satisfy
\begin{align*}
    0 &= (W \diag(\bar{q}) W^{\top})^{-1} W \diag(\pi(\bar{\theta})) W^{\top} (\psi - \bar{\theta}) &\implies \bar{\theta} &= \psi\\
    0 &= \pi(\bar{\theta}) - \bar{q} &\implies \bar{q} &= \pi(\bar{\theta}) = \pi(\psi)
\end{align*}
In the preceding, the first implication is due to $W \diag(\pi(x)) W^{\top}$ being a non-singular matrix for any $x \in \R^K.$ This indicates that the ODE has a unique equilibrium point $(\psi,\pi(\psi)).$ The next objective is to check whether the equilibrium point is (locally or globally) asymptotically stable. In order to show asymptotic stability, the standard procedure is to use positive definite (or positive semidefinite) functions called Lyapunov functions~\citep[Chapter~4]{khalil2002nonlinear}. We use the following result, popularly known as the \emph{invariance principle} in dynamical systems theory literature, to discuss the stability of the equilibrium point.
\begin{lemma}[La Salle's Invariance Principle]
  \label{lemma:la-salle-invariance-principle}
  Consider an ODE $\dot{x} = f(x),$ where $f: D \to \R^n$ is a locally Lipschitz map from a domain $D \subset \R^n$ to $\R^n.$ Let $\Omega \subset D$ be a compact set that is positively invariant with respect to the ODE, and let $V: D \to \R$ be a continuously differentiable function such that $\dot{V}(x) \leq 0$ in $\Omega.$ Let $E$ be the set of all points in $\Omega$ where $\dot{V}(x) = 0.$ Let M be the largest invariant set in E. Then every solution starting in $\Omega$ approaches M as $t \to \infty.$
\end{lemma}
The invariance principle can be used when one can come up with an energy function $V$ which satisfies $V(\theta,q) \geq 0$ for all $(\theta,q),$ and $\dot{V}(\theta,q) \leq 0$ for all $(\theta,q).$ Using the invariance principle, we can show that $\theta(\tau)$ and $q(\tau)$ converge to an invariant set contained in the set of points that satisfies $\dot{V}(\theta,q)=0.$ The following result shows that such an invariant set is the singleton set containing the equilibrium point, thereby establishing the asymptotic stability of the equilibrium.
\begin{theorem}
  $(\psi,\pi(\psi))$ is an equilibrium point for the ODE \eqref{eq:level-1-ode}. Moreover, when $K=p,$ the equilibrium point is globally asymptotically stable.
\end{theorem}
% \begin{proof} %TODO: Write the stability proof for the general case using Taylor's theorem and change of variables, if you can.
%   Define new objects $\underline{\theta} = \theta - \psi$ and $\underline{q} = q - \pi(\psi).$ The ODEs corresponding to these both objects are given by
%   \begin{align*}
%     \dot{\underline{\theta}} &= - (W \diag(\underline{q} + \pi(\underline{\theta})) W^{\top})^{-1} W \diag(\pi(\underline{\theta}+\psi)) W^{\top} \underline{\theta} \\
%     \dot{\underline{q}} &= \pi(\theta) - \pi(\psi) -\underline{q}
%   \end{align*}
% \end{proof}
\begin{proof}
  The ODE is given by
  \begin{align*}
    \dot{\theta}(\tau) &= (W \diag(Q) W^{\top})^{-1} W \diag(\pi(\theta)) W^{\top} (\psi - \theta) \\
    \dot{q}(\tau) &= \pi(\theta) - q(\tau).
  \end{align*}
  It can be seen that $(\theta,q) = (\psi,\pi(\psi))$ gives $\dot{\theta} = 0$ and $\dot{q} = 0,$ and hence is an equilibrium point. To prove asymptotic stability, we use the following positive semidefinite function.
  \begin{align*}
    V(\theta,q) &= (\theta - \psi)^{\top} W \diag(\pi(\psi)) W^{\top} (\theta - \psi) \\
    \implies \dot{V}(\theta,q) &= (\theta - \psi)^{\top} W \diag(\pi(\psi)) W^{\top} (W \diag(q) W^{\top})^{-1} W \diag(\pi(\theta)) W^{\top} (\psi - \theta)\\
    \implies \dot{V}(\theta,q) &= - (\theta - \psi)^{\top} W \diag(\pi(\psi)) (W^{\dagger} W)^{\top} (\diag(q))^{-1} (W^{\dagger} W) \diag(\pi(\theta)) W^{\top} (\theta - \psi)
  \end{align*}
  When $K=p,$ we have $W^{\dagger}W=I,$ which makes $\dot{V} \leq 0$ for all $(\theta,q).$ In particular, $\dot{V}=0$ for points $(\psi,q)$ for all possible $q,$ and $\dot{V}<0$ for all other points.\\
  Using the invariance principle (Lemma~\ref{lemma:la-salle-invariance-principle}), we can narrow down the limit set to the largest invariant set contained in the set of points satisfying $\dot{V}=0,$ which is the set $\{(\psi,q): q > 0\}.$ Clearly, the equilibrium point $(\psi,\pi(\psi))$ must belong to the invariant set. Moreover, note that no other point can belong to the invariant set, because one can come up with a trajectory that begins outside the set and passes through the point under consideration. Hence, the point $(\psi,\pi(\psi))$ is the only point in the invariant set.
\end{proof}

The global asymptotic stability of the equilibrium point, along with the result that relates the equilibrium points of the ODE to the limit set of the algorithm state (Lemma~\ref{lemma:ode-method-isolated-equilibria}), has the following key implication: the limit set $L(\theta_t,t \leq 0)=\{(\psi,\pi(\psi))\}$ is a singleton set containing the equilibrium of the ODE. Hence, the algorithm state $\theta_t$ converges to the fixed user preference $\psi$ almost surely as $t \to \infty.$ This property enables the algorithm to perform very well when the user preferences are time-invariant.

\section{Summary}
In this chapter, we introduced a contextual bandit-based recommendation algorithm that is learning preferences of users that do not change with time. Then, we discussed a method of analysis that uses the ODE method of stochastic approximation to understand the long term behavior of the algorithm. The asymptotic analysis of the algorithm showed that the algorithm state asymptotically converges to the user preferences, which implies that the recommendation algorithm asymptotically learns the optimal policy.  However, something different can happen when the user preferences change with time. In such a case, both the algorithm and the user preferences evolve with time. In order to understand the long term consequences of the interaction between the algorithm and the user, one must analyze their co-evolution rather than understanding them in isolation. This topic is the subject of discussion in the next chapter.

%===========================
% Chapter 3
\chapter{User-RS Interaction: A Coupled Dynamical System}
\label{ch:one-user-model}
In this chapter we describe and analyze a model of interaction between a RS and a user whose preferences are influenced by the recommendations. Specifically, we extend the model from the previous chapter such that the user preference vector $\psi$ is not static as in \cref{ch:recommendation-model}, but is a time-varying user preference $\psi_t$ indexed by time $t$ that is sensitive to recommendations made by the RS. We analyze the long term behavior of both the algorithm states and user preferences. The resulting model establishes a relation between long term properties of the algorithm state and the user preferences, and the parameters that characterize the evolution of the same. For example, we relate the heterogenity of user preferences to the exploration-exploitation tradeoff made by the recommendation algorithm.

We describe the model with recommendation-influenced user preferences in \cref{sec:level-2-model-description}, and obtain the asymptotic ODE of such a model in \cref{sec:level-2-model-analysis}. We characterize the limit set of the user preferences and algorithm state by analyzing the equilibrium points of the asymptotic ODE in \cref{sec:level-2-equilibrium-analysis}. In this section, we also discuss the relation between the user preferences and the exploration-exploitation tradeoff made by the recommendation algorithm.

\section{Model Description}
\label{sec:level-2-model-description}
%TODO: Add a picture that illustrates the theorem. Get it from ICML.
Consider a discrete-time sequence of interactions between the recommendation algorithm \RS described in \cref{sec:level-1-model-description} and a user with recommendation-susceptible preferences. Recall, from \cref{sec:level-1-model-description}, that the algorithm used by \RS consists of a recommendation policy and a learning algorithm, where the recommendation policy is defined by
\begin{equation}
  \label{eq:level-2-recommendation-policy}
  \pi_k(\theta_t) = \frac{\exp(a w_k^{\top} \theta_t)}{\sum_{j=1}^K \exp(a w_j^{\top} \theta_t)}
\end{equation}
and the learning algorithm is defined by the recursive equations
\begin{align}
    \theta_{t+1} &= \theta_t + \frac{1}{t+1} \left(\sum_{k=1}^K Q_{t+1}^k w_k w_k^{\top} \right)^{-1} \left( w_{A_t} R_t - w_{A_t} w_{A_t}^{\top} \theta_t \right) \label{eq:level-2-learner-recursion}, \\
  Q_{t+1}^k &= Q_t^k + \frac{1}{t+1} \left( [A_{t} = k] - Q_t^k \right)   \label{eq:level-2-learner-q-recursion}
\end{align}
for all $t \in \{1,2,\ldots\}$ with initial values are $\theta_1=0$ and $Q_1=\lambda \mathbf{1}_K.$

In the following, we discuss the model for the dynamics of user preferences. Instead of considering a fixed preference vector $\psi$ like in \cref{ch:recommendation-model}, we consider recommendation-influenced user preferences.
Let $\psi_t$ denote the user preference at time $t.$
%Consider a model in which user preferences change with time---these preferences at time $t$ are denoted by $\psi_t.$
We consider a model for the dynamics of user preferences in which the user preferences shift towards the recommendation made by \RS at time $t,$ \footnote{For other kinds of user preference dynamics models, see \cite{curmei2022towards}. One can also refer to literature in opinion dynamics that discusses models of influence among people.} i.e.,
\begin{equation}
  \label{eq:level-2-user-recursion}
  \psi_{t+1} = \psi_t + \beta_t (w_{A_t} - \psi_t).
\end{equation}
Here, the initial user preferences are described by the vector $\psi_1.$ The scalar $\beta_t > 0$ is called the \textit{sensitivity parameter} of the user; it is a positive value that indicates how much the user preference vector changes due to the recommendation at time $t.$ We also assume that the sequence $(\beta_t)$ decays to 0, and the ratio of $\beta_t$ and the update rate of the recommendation algorithm converges to a constant, i.e.,
\begin{equation}
  \label{eq:step-size-asymptotic-assumption}
  \lim_{t \to \infty} \beta_t = 0
  \qquad \text{and} \qquad
  \lim_{t \to \infty} \frac{\beta_t}{\frac{1}{t+1}} = \lim_{t \to \infty} (t+1) \beta_t = \rho.
\end{equation}
The first assumption in \cref{eq:step-size-asymptotic-assumption} is justified by observations of how people develop preferences. Initial experiences tend to have the most impact on people's preferences and established preferences tend to stabilize over time~\cite{hoeffler1999constructing}. The second assumption in \cref{eq:step-size-asymptotic-assumption} relates the rate of change of the algorithm state to that of the user preferences. If $\rho>1,$ then the user preferences change faster than the algorithm state in the long term and vice versa. We will analyze the effect of $\rho$ on the system in \cref{ssec:level-2-effect-of-rho}. The interaction between \RS and the user is described concisely in \cref{alg:level-2-model}.

%This assumption relates the step sizes of user adaptation with the step sizes of the learning algorithm. When this assumption is satisfied, the change in user preferences $\psi_t$ is $\rho$ times larger than the change in the learning algorithm parameter $\theta_t$ for very large $t.$ \\

\begin{algorithm}
  \caption{Recommendation policy and learning algorithm used by \RS}
  \begin{algorithmic}[1]
    \Require Number of items $K,$ attribute size $d,$ item attributes $w_k \in \R^{d}$ for all $k \in [K],$ regularization parameter $\lambda,$ recommendation parameter $a,$ initial user preference $\psi_1.$
    % \Ensure Output description (e.g., sorted array $A$)
    \State $t \gets 1$
    \State $\theta_1 \gets 0_d$
    % \State $S_0 \gets \lambda \sum_{k=1}^K w_k w_k^{\top}$
    % \For{t=1,2,3,\ldots}
    \Loop
      \State Recommend $A_t \in [K]$ with $P(A_t=k) = \frac{\exp(a w_k^{\top} \theta_t)}{\sum_{j=1}^K \exp(a w_j^{\top} \theta_t)}$
      \State Collect reward $R_t \gets w_{A_t}^{\top} \psi_t + \epsilon_t$
      \State $\theta_{t+1} \gets  \arg \min_{\theta} \lambda \|\theta\|_{W W^{\top}}^2 + \sum_{s=1}^t \left( R_s - w_{A_s}^{\top} \theta \right)^2$
      \State $\psi_{t+1} \gets \psi_t + \beta_t (w_{A_t}-\psi_t)$
      \State $t \gets t+1$
    \EndLoop
    % \State \Return result
  \end{algorithmic}
  \label{alg:level-2-model}
\end{algorithm}

\section{Obtaining the Asymptotic ODE}
\label{sec:level-2-model-analysis}
Our goal in this model is to understand the long term behavior of the sequence of algorithm state $(\theta_t,Q_t)$ and sequence of user preferences $(\psi_t).$ In particular, we are interested in obtaining the limit sets of both of these sequences. To that end, we follow the analysis method used in \cref{ch:recommendation-model}---we use the ODE method of stochastic approximation to obtain the asymptotic ODE for this model. Asymptotically stable equilibria of the asymptotic ODE, if they exist, give insights into the limit sets of interest. In the following, we find the stochastic approximation (SA) recursions for the algorithm states $(\theta_t,Q_t)$ and user preferences $(\psi_t).$ Then, we use the SA recursions to obtain the asymptotic ODE for the system. For a more elaborate discussion on this method, we refer the reader to \cref{sec:level-1-asymptotic-ode}.
%\subsection{Obtaining the stochastic approximation type equations}

We first rewrite \cref{eq:level-2-learner-recursion,eq:level-2-learner-q-recursion,eq:level-2-user-recursion} to take the form of stochastic-approximation-type recursion given in \cref{eq:ode-method-sa-recursion}.  Define $\mathcal{F}_t := \sigma(\theta_s,Q_s,\psi_s, s \leq t)$. Using the approach given in \cref{ssec:level-1-stochastic-approximation-recursions}, we obtain the following recursions for $(\theta_t,Q_t)$
\begin{align}
  \theta_{t+1} &= \theta_t + \frac{1}{t+1} \left( h_{\theta}(\theta_t,Q_t,\psi_t) + M_{t+1} + \gamma_{t+1} \right) \label{eq:level-2-learner-sa-recursion} \\
  Q_{t+1} &= Q_t + \frac{1}{t+1} (h_q(\theta_t,q_t,\psi_t) + M'_{t+1}) \label{eq:level-2-learner-q-sa-recursion}.
\end{align}
Here, $M_{t+1} := (t+1) (\theta_{t+1} - \theta_t - \Exp{\theta_{t+1} - \theta_t | \mathcal{F}_t}),$ $M'_{t+1} := (t+1) (Q_{t+1}-Q_t - \Exp{Q_{t+1}-Q_t| \mathcal{F}_t}).$ and the sequence $(\gamma_t)$ asymptotically converges to 0 almost surely. Further, the functions $h_{\theta}$ and $h_q$ are defined as
\begin{align}
  h_{\theta}(\theta,Q,\psi) &:= (W \diag(Q) W^{\top})^{-1} W \diag(\pi(\theta)) W^{\top} (\psi - \theta) \label{eq:level-2-learner-mean-ode} \\
  h_q(\theta, Q, \psi) &:= \pi(\theta) - Q \label{eq:level-2-learner-q-mean-ode}.
\end{align}
For the sake of avoiding repetition, we refer the reader to \cref{ssec:level-1-stochastic-approximation-recursions} for the process of obtaining the above recursions. Most of the derivation details follow identically to that of \cref{ssec:level-1-stochastic-approximation-recursions} except for the usage of time-varying preference $\psi_t$ instead of a static preference vector $\psi.$

We are now left with the task of obtaining a stochastic approximation type recursion for the user preferences $(\psi_t).$ To do so, we begin by analyzing $\frac{E[\psi_{t+1}-\psi_t | \theta_t,Q_t,\psi_t]}{\beta_t},$ which represents the expected deviation of the user preference vector, given the state of the model at time $t.$ This term will be used in describing the asymptotic ODE (as discussed in \cref{ssec:level-1-stochastic-approximation-recursions}).
% The mean ODE analysis: The previously derived functions for the mean ODE can be used for $\theta_t$ and $Q_t,$ with the addition of $\psi.$ We redefine these functions as follows for this section.
% \begin{align}
%   h_{\theta}(\theta,Q,\psi) &:= (W \diag(Q) W^{\top})^{-1} W \diag(\pi(\theta)) W^{\top} (\psi - \theta) \\
%   h_q(\theta, Q, \psi) &:= \pi(\theta) - Q
% \end{align}
% We find the mean trajectory of $\psi.$
\begin{align*}
  \frac{\Exp{\psi_{t+1} - \psi_t | \theta_t, Q_t, \psi_t}}{\beta_t} &= \Exp{w_{A_t} - \psi_t | \theta_t, Q_t, \psi_t} \\
  &= \sum_{k=1}^K \pi_k(\theta_t) w_k - \psi_t = W \pi(\theta_t) - \psi_t
\end{align*}
Interpreting the above conditional expectation as a function of $(\theta_t,Q_t,\psi_t),$ we define $h_{\psi}$ s.t.
\begin{equation}
  \label{eq:level-2-user-mean-ode}
  h_{\psi}(\theta,q,\psi) := W \pi(\theta) - \psi.
\end{equation}
Using \cref{eq:level-2-user-mean-ode}, we can now derive the stochastic approximation recursion for $\psi_t.$
   \begin{align*}
    \frac{\psi_{t+1}-\psi_t}{\frac{1}{t+1}} &= \left( \frac{\psi_{t+1}-\psi_t}{\frac{1}{t+1}} - \Exp{\frac{\psi_{t+1}-\psi_t}{\frac{1}{t+1}} \Bigg| \mathcal{F}_t} \right) +  \Exp{\frac{\psi_{t+1}-\psi_t}{\frac{1}{t+1}} \Bigg| \mathcal{F}_t} \\
                                                &= M''_{t+1} + (t+1)\Exp{\psi_{t+1}-\psi_t | \mathcal{F}_t} \\
                                                &= M''_{t+1} + (t+1)\beta_t h_{\psi}(\theta_t,Q_t,\psi_t) \\
                                                &= M''_{t+1} + \rho h_{\psi}(\theta_t,Q_t,\psi_t) + \gamma''_{t+1}
   \end{align*}
   where $M''_{t+1} := (t+1) (\psi_{t+1} - \psi_t - \Exp{\psi_{t+1} - \psi_t | \mathcal{F}_t})$ and $\gamma''_t := \left( (t+1)\beta_t - \rho \right) h_{\psi}(\theta_t, Q_t, \psi_t).$ This lets us rewrite Eq.~\eqref{eq:level-2-user-recursion} as
  \begin{equation}
    \psi_{t+1} = \psi_t + \frac{1}{t+1} \left(\rho h_{\psi}(\psi_t,Q_t,\psi_t) + M''_{t+1} + \gamma''_{t+1} \right)
    \label{eq:level-2-user-sa-recursion}.
  \end{equation}
This equation is the stochastic approximation equation for the user preference $\psi_t$ (one can verify that by comparing it with the generic stochastic approximation recursion given by \cref{eq:ode-method-sa-recursion}).
%
%\subsection{Obtaining the Asymptotic ODE}
  Concatenating the stochastic approximation recursions for all the iterates $(\theta_t,Q_t,\psi_t),$ given by \cref{eq:level-2-learner-sa-recursion,eq:level-2-learner-q-sa-recursion,eq:level-2-user-sa-recursion}, results in a single recursive relation:
  \begin{equation}
    \begin{bmatrix}
      \theta_{t+1} \\ Q_{t+1} \\ \psi_{t+1}
    \end{bmatrix} =
    \begin{bmatrix}
      \theta_t \\ Q_t \\ \psi_t
    \end{bmatrix} + \frac{1}{t+1} \left(
      \begin{bmatrix}
        h_{\theta}(\theta_t,Q_t,\psi_t) \\ h_q(\theta_t,Q_t,\psi_t) \\ \rho h_{\psi} (\theta_t, Q_t, \psi_t)
      \end{bmatrix} +
      \begin{bmatrix}
        M_{t+1} \\ M'_{t+1} \\ M''_{t+1}
      \end{bmatrix} +
      \begin{bmatrix}
        \gamma_{t+1} \\ 0 \\ \gamma''_{t+1}
      \end{bmatrix}
    \right)
    \label{eq:level-2-sa-recursion}
  \end{equation}
  The asymptotic ODE for this system can be obtained by using the ODE method (Lemma~\ref{lemma:ode-method}) on the above recursive equation. The following result uses that method to derive the asymptotic ODE.
\begin{theorem}
  \label{thm:level-2-ode}
  Let the sequence $(\theta_t, Q_t, \psi_t)$ follow the recursions given by \cref{eq:level-2-learner-sa-recursion,eq:level-2-learner-q-sa-recursion,eq:level-2-user-sa-recursion} such that the sequence of sensitivity parameters $(\beta_t)$ follows \cref{eq:step-size-asymptotic-assumption}. Then, the sequence $(\theta_t, Q_t, \psi_t)$ converges to a connected internally chain recurrent set of the ODE
  \begin{equation}
    \begin{split}
      \dot{\theta}(\tau) &= h_{\theta}(\theta,q,\psi) \\
      \dot{q}(\tau) &= h_{q}(\theta,q,\psi) \\
      \dot{\psi}(\tau) &= \rho h_{\psi}(\theta,q,\psi)
    \end{split}
    \label{eq:level-2-ode}
  \end{equation}
\end{theorem}
\begin{proof}
  % Concatenating the stochastic approximation recursions for all the iterates $(\theta_t,Q_t,\psi_t),$ given by \cref{eq:level-2-learner-sa-recursion,eq:level-2-learner-q-sa-recursion,eq:level-2-user-sa-recursion}, results in a single recursive relation.
  % \begin{equation*}
  %   \begin{bmatrix}
  %     \theta_{t+1} \\ Q_{t+1} \\ \psi_{t+1}
  %   \end{bmatrix} =
  %   \begin{bmatrix}
  %     \theta_t \\ Q_t \\ \psi_t
  %   \end{bmatrix} + \frac{1}{t+1} \left(
  %     \begin{bmatrix}
  %       h_{\theta}(\theta_t,Q_t,\psi_t) \\ h_q(\theta_t,Q_t,\psi_t) \\ h_{\psi} (\theta_t, Q_t, \psi_t)
  %     \end{bmatrix} +
  %     \begin{bmatrix}
  %       M_{t+1} \\ M'_{t+1} \\ M''_{t+1}
  %     \end{bmatrix} +
  %     \begin{bmatrix}
  %       \gamma_{t+1} \\ 0 \\ \gamma''_{t+1}
  %     \end{bmatrix}
  %   \right)
  % \end{equation*}
  The proof idea is to use the ODE method (Lemma~\ref{lemma:ode-method}) on the recursive equation given by \cref{eq:level-2-sa-recursion} to obtain the asymptotic ODE. The rest of the proof shows that the recursive relation satisfies the assumptions needed to use the ODE method. This proof is similar to the proof of Theorem~\ref{thm:level-1-ode}. Many steps in this proof related to $\theta_t$ and $Q_t$  are similar to what was discussed in the proof of Theorem~\ref{thm:level-1-ode}, and hence such steps are omitted to avoid repetition. We will primarily focus on $\psi_t,$ which was not analyzed before, in this proof. \\
  
  \textbf{A1.} The stepsize sequence $(\frac{1}{t+1})$ satisfy $\sum_{t=1}^{\infty} \frac{1}{t+1} = \infty$ and $\sum_{t=1}^{\infty} \frac{1}{(t+1)^2} = \frac{\pi^2}{6} < \infty.$\\

  \textbf{A2.} (Locally Lipschitz maps) To prove that $h_{\psi}$ is locally Lipschitz, it suffices to show that $h_{\psi}$ is a continuously differentiable function for all $(\theta,Q,\psi).$ Recall
  \begin{align*}
    h_{\psi}(\theta,q,\psi) &= W \pi(\theta) - \psi \\
    \nabla_{\theta} h_{\psi} &= W \nabla_{\theta} (\pi(\theta)) = a W (\diag(\pi(\theta)) - \pi(\theta) \pi(\theta)^{\top}) \\
    \nabla_s h_{\psi} &= 0\\
    \nabla_{\psi} h_{\psi} &= -I
  \end{align*}
  Since $\pi(\theta)$ is a continuous function for all $a \in \R,$ $\nabla_{\theta}h_{\psi}$ is also continuous. Hence, $h_{\psi}$ is locally Lipschitz.\\
  Recall that in the proof of Theorem~\ref{thm:level-1-ode}, we showed that the derivatives of $h_{\theta}$ and $h_s$ with respect to $\theta$ and $q$ are continuous functions. Thus, it suffices to show that the derivatives of $h_{\theta}$ and $h_s$ with respect to $\psi$ are continuous. Recall that
  \begin{align*}
    h_{\theta}(\theta,q,\psi) &= (W \diag(q) W^{\top})^{-1} W \diag(\pi(\theta)) W^{\top} (\psi - \theta) \\
    \implies \nabla_{\psi} h_{\theta} &= (W \diag(q) W^{\top})^{-1} W \diag(\pi(\theta)) W^{\top}
  \end{align*}
  and
  \begin{align*}
    h_q(\theta, q, \psi) &= \pi(\theta) - q \\
    \implies \nabla_{\psi} h_q &= 0
  \end{align*}
  Since $\nabla_{\psi} h_{\theta}$ and $\nabla_{\psi} h_q$ are continuous, we can conclude that $h_{\theta}$ and $h_q$ are locally Lipschitz.

  \textbf{A3.} (Square integrability of Martingale difference sequence)
  We show that $M_{t+1}$ and $M'_{t+1}$ are square integrable in the proof of Theorem~\ref{thm:level-1-ode} assuming that $\psi_t$ is a constant $\psi$---the same proof is applicable under the additional assumption that $\|\psi_t\|$ is bounded for all $t$ (which is shown to hold later in this proof). In the following, we turn our attention to $M''_{t+1}.$ Firstly, it is straightforward to check that $M''_{t+1}$ is a martingale difference sequence.
  \begin{align*}
    \Exp{M''_{t+1} | \mathcal{F}_t} &= \Exp{(t+1) (\psi_{t+1} - \psi_t - \Exp{\psi_{t+1} - \psi_t | \mathcal{F}_t}) | \mathcal{F}_t} \\
    &= (t+1) \left( \Exp{ \psi_{t+1} - \psi_t | \mathcal{F}_t} - \Exp{\psi_{t+1} - \psi_t | \mathcal{F}_t} \right) = 0.
  \end{align*}
  We now show the square integrability of $M''_{t+1}.$
  \begin{align*}
    \Exp{\|M''_{t+1}\|^2 | \mathcal{F}_t} &= \Exp{ \left\| (t+1) \beta_t (w_{A_t} - \psi_t) - \Exp{(t+1) \beta_t (w_{A_t} - \psi_t) | \mathcal{F}_t} \right\|^{2} | \mathcal{F}_t} \\
     &= \Exp{ \left\| (t+1) \beta_t (w_{A_t} - \psi_t) \right\|^2 \Big| \mathcal{F}_t } - \left\| \Exp{(t+1) \beta_t (w_{A_t} - \psi_t) | \mathcal{F}_t} \right\|^2 \\
     &\leq ((t+1) \beta_t)^2 \Exp{ \left\| (w_{A_t} - \psi_t) \right\|^2 \Big| \mathcal{F}_t } \\
     &= ((t+1) \beta_t)^2 \sum_{k=1}^K \pi_k(\theta_t) \left\| (w_k - \psi_t) \right\|^2  \\
     &\leq ((t+1) \beta_t)^2 \max_{k \in [K]} \left\| (w_k - \psi_t) \right\|^2
  \end{align*}
  Let $k_{\ast} = \arg \max_{k \in [K]} \left\| (w_k - \psi_t) \right\|^2.$ Then
  \begin{align*}
      \Exp{M''_{t+1} | \mathcal{F}_t} &\leq ((t+1) \beta_t)^2 \left\| w_{k_{\ast}} - \psi_t \right\|^2 \\
        &\leq ((t+1) \beta_t)^2 ( \| w_{k_{\ast}} \|^2 + \| \psi_t \|^2 ) \\
        &\leq ((t+1) \beta_t)^2 \max\{ \| w_{k_{\ast}} \|^2, 1 \} ( 1 + \| \psi_t \|^2 )
  \end{align*}
  Since $(t+1)\beta_t$ is finite for all $t \in \R$ and $\lim_{t \to \infty} (t+1) \beta_t = \rho,$ the object $\sup_t (t+1)\beta_t$ is bounded from above. Hence, $M''_{t+1}$ is square integrable.

  \textbf{A4.} (Boundedness of iterates). We will first show that $\|\psi_t\|$ is bounded. Recall that
  \begin{align*}
    \|\psi_{t+1}\| &= \| (1-\beta_t) \psi_t + \beta_t w_{A_t} \| \\
    &\leq (1-\beta_t) \| \psi_t\| + \beta_t \| w_{A_t} \| \\
    &\leq \max\{\|\psi_t\|, \|w_{A_t}\|\} \leq \max\{\|\psi_t\|, \max_{k \in [K]}\|w_k\|\} 
  \end{align*}
  Recursively applying this to all past $t,$ we get
  \begin{equation*}
    \|\psi_{t+1}\| \leq \max\{ \| \psi_0 \|, \max_{k \in [K]} \|w_k\| \}
  \end{equation*}
  Since this bound is applicable to all $t,$ $\|\psi_t\|$ is bounded almost surely. Moreover, using this, one can show that $\|\theta_t\|$ is bounded for all $t$ using the same steps as in the proof of Theorem~\ref{thm:level-1-ode}.

  \textbf{A5.} (Almost sure convergence of $\gamma''_t \to 0.$) We analyze the norm of $\gamma''_t.$
  \begin{align*}
    \|\gamma''_t\| &= \| \left( (t+1)\beta_t - \rho \right) h_{\psi}(\theta_t, Q_t, \psi_t) \| \\
     &= | (t+1)\beta_t - \rho | \| W \pi(\theta_t) - \psi_t \| \\
     &\leq | (t+1)\beta_t - \rho | ( \| W \| \| \pi(\theta_t) \| + \| \psi_t \| ) \\
     &\leq | (t+1)\beta_t - \rho | ( \| W \| + \| \psi_t \| )
  \end{align*}
  where the last step is because $\|\pi(\theta)\| \leq 1$ for any $\theta.$ Since $\|W\|$ is bounded and $\|\psi_t\|$ is bounded a.s. for all $t,$ we have $\|\gamma''_t\| \to 0$ because $(t+1)\beta_t - \rho \to 0$ as $t \to \infty.$ This implies that $\gamma_t \to 0$ asymptotically.
\end{proof}
The preceding result relates the long term behavior of the algorithm state $(\theta_t,Q_t)$ and the user preferences $\psi_t$ to that of the trajectories of the ODE $(\theta(\tau),q(\tau),\psi(\tau)).$ Motivated by this result, we analyze the asymptotic ODE to better understand the algorithm states and user preferences.

\section{Long term behavior of algorithm state and user preferences}
\label{sec:level-2-equilibrium-analysis}
% Find the equilibrium points and show local stability
In this section, we study the asymptotic ODE and its trajectories.
Throughout this section, we refer to $\theta(\tau)$ and/or $q(\tau)$ as the algorithm state and we refer to $\psi(\tau)$ as the user preferences at time $\tau.$
From \cref{thm:level-2-ode}, and the definitions of $h_{\theta},h_q,$ and $h_{\psi},$ the asymptotic ODE is expressed as
\begin{align*}
      \dot{\theta}(\tau) &= (W \diag(q) W^{\top})^{-1} W \diag(\pi(\theta)) W^{\top} (\psi - \theta) \\
      \dot{q}(\tau) &= \pi(\theta) - q \\
      \dot{\psi}(\tau) &= \rho ( W \pi(\theta) - \psi ).
\end{align*}
In the preceding, we omit the usage of $\tau$ and refer to $\theta(\tau)$ and $q(\tau)$ as $\theta$ and $q$ respectively.
% We can generate a trajectory using the asymptotic ODE that begins at some initial point $(\theta(0),q(0),\psi(0)).$
Note that an ODE trajectory captures the behavior of the sequence $(\theta_t,Q_t,\psi_t)$ in an expected sense. Furthermore, \cref{thm:level-2-ode} guarantees that $(\theta_t,Q_t,\psi_t)$ converges to the same set that the trajectories of the asymptotic ODE converge to.

\begin{figure}[tb]
  \centering
  \includegraphics{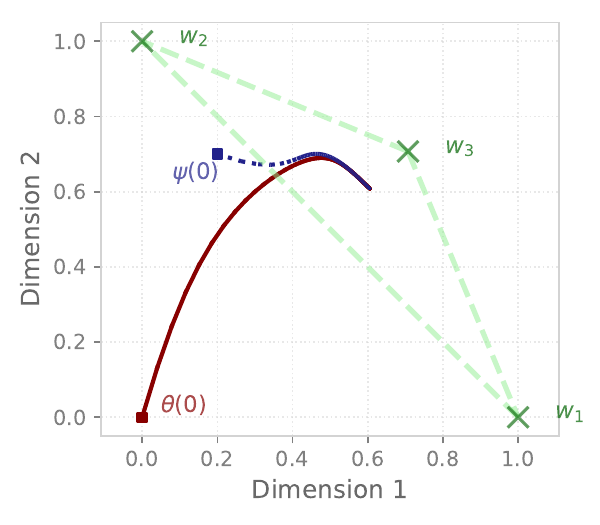}
  \caption{Numerical simulation of a trajectory of the asymptotic ODE}
  \label{fig:level-2-example}
\end{figure}

We validate this using a numerical simulation to plot the ODE trajectories. Consider an example in which we set $d=2;$ this enables us to plot the trajectories of the ODE on the 2-dimensional Euclidean plane. The item attributes are $w_1=[1~0]^{\top},$ $w_2=[0~1]^{\top},$ and $w_3=[\frac{1}{\sqrt{2}}~\frac{1}{\sqrt{2}}]^{\top},$ and the model parameters are set as $a=3$ and $\rho=0.5.$ The initial values are given by $\theta(0)=\theta_1=0,$ $q(0)=q_1=\frac{1}{K} \mathbf{1}_K,$ and $\psi(0)=[0.2~0.7^{\top}].$ The trajectories of the ODE given the preceding values, called an \emph{initial value problem}, are plotted in \cref{fig:level-2-example}. The initial points are denoted by a square marker, and the trajectories of $\theta$ and $\psi$ are plotted in dark red and dark blue respectively. Furthermore, the item attributes are plotted in green.

We now discuss several observations on the example trajectory shown in \cref{fig:level-2-example}. First, the trajectory of $\theta$ appears to asymptotically meet with the trajectory of $\psi,$ i.e., the algorithm asymptotically drives its least squares loss to its minimum possible value despite the presence of recommendation-influenced user preferences.  Second, the trajectories of the user preferences asymptotically end up in the convex hull of all the item attributes (shown by dashed green lines). This indicates that the set of all available items dictate long term user preferences.

The cause of this observed behavior can be understood by analyzing the asymptotic ODE. To explain the first observation, note that the norm $\|\dot{\theta}(\tau)\|$ depends on $\|\psi-\theta\|,$ which is the distance between the algorithm state and user preferences. Hence, $\|\dot{\theta}(\tau)\|$ should decrease with time, provided that $\theta(\tau)$ is moving in the direction that decreases $\|\psi-\theta\|.$ To understand the second observation, see that $\dot{\psi}(\tau)$ always points in the direction that takes the current user preference $\psi(\tau)$ towards $W \pi(\theta(\tau)),$ which is a point in the convex hull of the item attributes.

A third key observation is that the both the trajectories $\theta(\tau)$ and $\psi(\tau)$ appear to converge to a point. Convergence of trajectories in ODEs are associated with asymptotically stable equilibrium points. Recall that the equilibria of the ODE have the property that all trajectories beginning at such a point will remain at that point at all times. Further, asymptotically stable equilibria are points that have attracting neighbourhoods: trajectories that are initiated sufficiently close to an equilibrium point converge to that equilibrium. The existence of such asymptotically stable equilibria indicates the possibility of the system \emph{stabilizing in the long term}, i.e., the user preferences settle down over time and the learning algorithm stabilizes. The convergence of trajectories of the asymptotic ODE motivates the study of asymptotically stable equilibria of the ODE.

Let $(\bar{\theta},\bar{q},\bar{\psi})$ be an equilibrium point of the asymptotic ODE. Such an equilibrium point, if it exists, must satisfy the equations $\dot{\theta}(\tau)=\dot{q}(\tau)=\dot{\psi}(\tau)=0.$ Substituting the equilibrium points into these equations gives us
\begin{equation*}
  \bar{\psi} = \bar{\theta}, \qquad
  \bar{q} = \pi(\bar{\theta}), \qquad
  \bar{\psi} = W \pi(\bar{\theta}).
\end{equation*}
Before we proceed to the analysis, we must verify if such an equilibrium point even exists. From the preceding, observe that $\bar{\psi}$ satisfies $\bar{\psi} = W \pi(\bar{\psi}).$ This equation always has a solution in the convex hull of item attributes. This can be seen by first observing that $\pi(x)$, for any $x \in \R^K,$ is a vector in the $(K-1)$-dimensional probability simplex, and consequently, $W \pi(x)$ belongs to the convex hull of item attributes $\conv(w_1,w_2,\ldots,w_K).$ Given this fact, Brouwer's fixed point theorem guarantees the existence of at least one solution to the equation $\bar{\psi}= W \pi(\bar{\psi})$ in $\conv(w_1,w_2,\ldots,w_K) \subset \R^d.$ This guarantees that there exists at least one equilibrium point to the asymptotic ODE given by \cref{eq:level-2-ode}.

Now that we have established the existence of equilibria to the asymptotic ODE, we proceed to discuss the asymptotic stability of those equilibria.
\begin{theorem}
  Let the set $\{Y\}$ contain the solutions to the equation
  \begin{equation*}
    y = \sum_{k=1}^K \pi_k(y) w_k = \sum_{k=1}^K \frac{\exp(a w_k^{\top} y)}{\sum_{j=1}^K \exp(a w_j^{\top} y)} w_k
  \end{equation*}
  The set $\{(y_k,\pi(y_k),y_k): y_k \in Y\}$ is the set of equilibrium points to the ODE~\eqref{eq:level-2-ode}. Moreover, all the equilibrium points of the ODE are locally asymptotically stable.
  \label{thm:level-2-equilibrium-stability}
\end{theorem}
\begin{proof}
  It is easily seen that all points of the form $\theta=y$ result in $\dot{\theta}=0$ and all points of the form $q=\pi(\theta)$ result in $\dot{q} = 0.$ Hence, all the points in the set $\{(y_k,\pi(y_k),y_k): y_k \in Y\}$ satisfy $\dot{\theta}=0$ and $\dot{q}=0.$ Moreover, for all the points in the proposed set, observe that
  \begin{equation*}
    \dot{\psi}(\tau) = W \pi(y) - y = \sum_{k=1}^K \pi_k(y) w_k - y = 0.
  \end{equation*}
  This proves that the set of points $\{(y_k,\pi(y_k),y_k): y_k \in Y\}$ are indeed the set of equilibrium points of the ODE~\eqref{eq:level-2-ode}.

  To prove the asymptotic stability of an equilibrium point $(y,\pi(y),y),$ where $y \in Y,$ we first define new objects $\underline{\theta} = \theta - y,$ $\underline{q} = q - \pi(y),$ and $\underline{\psi} = \psi - y.$ To show local asymptotic stability of the equilibrium point, it suffices to show that the origin is the equilibrium point of the dynamical system characterized by $(\underline{\theta}(\tau), \underline{q}(\tau), \underline{\psi}(\tau)).$ The ODE of such a system is given by
  \begin{align*}
    \dot{\underline{\theta}}(\tau) &= (W \diag(\underline{q} + \pi(y)) W^{\top})^{-1} W \diag(\pi(\underline{\theta} + y)) W^{\top} (\underline{\psi} - \underline{\theta}) \\
      \dot{\underline{q}}(\tau) &= \pi(\underline{\theta} + y) -\pi(y) - \underline{q} \\
      \dot{\underline{\psi}}(\tau) &= \rho (W \pi(\underline{\theta} + y) - y - \underline{\psi})
  \end{align*}
  We can use Taylor's theorem to obtain an expression for $\pi(\underline{\theta} + y)$ as shown. %TODO: State and cite Taylor's theorem
  \begin{align*}
    \pi_k(\underline{\theta} + y) = \pi_k(y) + \sum_{j=1}^d R_k^j(\underline{\theta} + y) \underline{\theta} &;\quad R_k^j(\underline{\theta}+y) = \int_0^1 \frac{\partial}{\partial \underline{\theta}_j} \pi_k(y + t \underline{\theta}) dt \\
    \implies \pi(\underline{\theta} + y) = \pi(y) + R(\underline{\theta} + y) \underline{\theta} &;\quad R(\underline{\theta} + y) := 
          \begin{bmatrix}
            \sum_{j=1}^d R_1^j(\underline{\theta}+y) \\
            \sum_{j=1}^d R_2^j(\underline{\theta}+y) \\
            \vdots \\
            \sum_{j=1}^d R_K^j(\underline{\theta}+y) \\
          \end{bmatrix}_{K \times d}
  \end{align*}
  Using this Taylor expansion, the ODE can be expressed as
  \begin{align*}
    \dot{\underline{\theta}}(\tau) &= (W \diag(\underline{q} + \pi(y)) W^{\top})^{-1} W \diag(\pi(\underline{\theta} + y)) W^{\top} (\underline{\psi} - \underline{\theta}), \\
      \dot{\underline{q}}(\tau) &= R(\underline{\theta} + y)\underline{\theta} - \underline{q}, \\
      \dot{\underline{\psi}}(\tau) &= \rho (W R(\underline{\theta} + y) \underline{\theta} - \underline{\psi}).
  \end{align*}
  For ease of exposition, we define $A(\underline{\theta},\underline{q}):= (W \diag(\underline{q} + \pi(y)) W^{\top})^{-1} W \diag(\pi(\underline{\theta} + y)) W^{\top},$ and $x(\tau) = (\underline{\theta}(\tau), \underline{q}(\tau), \underline{\psi}(\tau)).$ Then the ODE can be expressed as
  \begin{equation*}
    \dot{x}(\tau) =
    \begin{bmatrix}
      -A(\underline{\theta},\underline{q}) & 0 & A(\underline{\theta},\underline{q}) \\
      R(\underline{\theta} + y) & -I & 0 \\
      \rho W R(\underline{\theta} + y) & 0 & -\rho I
    \end{bmatrix} x(\tau)
  \end{equation*}
  To show that the origin is an asymptotically stable equilibrium, it is sufficient to show that all of the eigenvalues of the Jacobian of the RHS evaluated at the origin are negative. %TODO: Cite the relevant lemma
  From the above expression, it is easy to see that
  \begin{equation*}
    \nabla_x \dot{x}(\tau) = \begin{bmatrix}
      -A(\underline{\theta},\underline{q}) & 0 & A(\underline{\theta},\underline{q}) \\
      R(\underline{\theta} + y) & -I & 0 \\
      \rho W R(\underline{\theta} + y) & 0 & -\rho I
    \end{bmatrix}_{x=0} =  \begin{bmatrix}
      -I & 0 & I \\
      0 & -I & 0 \\
      0 & 0 & -\rho I
    \end{bmatrix}
  \end{equation*}
  From Gershgorin's disc theorem, % TODO: Cite Gershgorin's disc theorem
  the eigenvalues of the above matrix must be negative or 0. However, since the determinant of the matrix is non-zero, and because the determinant is the product of the eigenvalues, none of the eigenvalues can be 0. Hence, all the eigenvalues must be negative. This proves the local asymptotic stability of the equilibrium point.
\end{proof}
Note that the above proof can be extended to determine the domain of attraction for a given equilibrium point by carefully analyzing the remainder term $R(\underline{\theta},y)$ that came from the Taylor series expansion. We choose not to prioritize such analysis in this thesis and leave it for future work.

\cref{lemma:ode-method-isolated-equilibria} guarantees that \emph{all the equilibria} of the asymptotic ODE are locally asymptotically stable, i.e. trajectories that begin sufficiently close to an equilibrium point converges to that equilibrium.
To understand where the trajectories of the asymptotic ODE lie in the long term, it suffices to know where the equilibrium lie.
%Further, \cref{lemma:ode-method-isolated-equilibria} tells us that these equilibria belong to the limit set of $(\theta_t,Q_t,\psi_t)$.
Understanding the properties of the equilibria helps us better understand the long term behavior of the algorithm states and user preferences. Hence, we focus our analysis on understanding other properties of these equilibria.

Firstly, observe that any equilibria $(\bar{\theta},\bar{q},\bar{\psi})$ of the asymptotic ODE satisfies $\bar{\theta}=\bar{\psi}.$ This indicates that the algorithm state converges to the user preference vector in the long term, despite the user preferences being non-stationary. Moreover, the user preference vector converges to some solution of the equation $y=W\pi(y).$ The solution set of this equation, however, is difficult to characterize. Observe that
\begin{align*}
  y &= W\pi(y) = \sum_{k=1}^K \frac{\exp(a w_k^{\top} y)}{\sum_{j \in [K]} \exp(a w_j^{\top} y)} w_k \\
  \implies 0 &= \sum_{k=1}^K \frac{\exp(a w_k^{\top} y)}{\sum_{j \in [K]} \exp(a w_j^{\top} y)} w_k - y\\
  \implies 0 &= \nabla_y \left( \frac{1}{a} \log \left( \sum_{k=1}^K \exp(a w_k^{\top} y) \right) - \frac{\|y\|^2}{2} \right) \\
  \implies 0 &= \nabla_y \left( \frac{1}{a} \log \left( \sum_{k=1}^K \exp\left( a w_k^{\top} y - \frac{a \|y\|^2}{2} \right) \right) \right) \\
  \implies 0 &= \nabla_y \left( \frac{1}{a} \log \left( \sum_{k=1}^K \exp \left( \frac{a \|w_k\|^2}{2} \right) \exp\left(\frac{-a \|y - w_k\|^2}{2} \right) \right) \right).
\end{align*}
Because $\log$ is a strictly increasing function, and because $a$ is finite, the solutions to the above equation are also solutions to
\begin{equation*}
  \nabla_y \left( \sum_{k=1}^K \xi_k \exp \left( \frac{-a \|y-w_k\|^2}{2} \right) \right) = 0
\end{equation*}
where $\xi_k=\exp \left( \frac{a \|w_k\|^2}{2} \right)$ for all $k \in [K].$ One can interpret that the above equation obtains the critical points of a weighted combination of Gaussian density functions. Hence, the number of solutions to the equation $y=W\pi(y)$ is greater than or equal to the number of modes of the density function of a Gaussians mixture in $d$-dimensions. \citet{carreira2003number} show that the exact number of modes of such a mixture is difficult to obtain in a general case. However, the solution set of interest is well understood when $a,$ the parameter in the recommendation policy that determines the exploration-exploitation tradeoff, is large or small. What is considered large or small is made more precise in the following section.
\subsection{Effect of the exploration-exploitation tradeoff}
\label{ssec:level-2-effect-of-a}
Recall the part of the asymptotic ODE that describes the user preference vector $\psi(\tau)$ is
\begin{equation*}
  \dot{\psi}(\tau) = \rho ( W \pi(\theta) - \psi ) = \rho \left( \sum_{k=1}^K \frac{\exp(a w_K^{\top} \theta)}{\sum_{j=1}^K \exp(a w_J^{\top}\theta)} w_k - \psi \right).
\end{equation*}
When $a=0,$ observe that the centroid of all the item attributes $\frac{1}{K}\sum_{k=1}^K w_k$ is the only equilibrium point for $\psi.$ This is the case in which all items are recommended uniformly at random to the user, i.e., the case in which the recommender explores but does not exploit. Since there exists only one equilibrium point, user preferences converge to the same point irrespective of the initial preferences. When $a \approx \infty,$ an equilibrium point of the ODE takes the form of $w_j,$ for some $j \in [K],$ that satisfies $w_j^{\top}\theta \geq w_k^{\top}\theta$ for all $k \in [K].$ In this case, the recommender only exploits and does not explore, and $w_j$ is the item that the recommender decides to suit the user best. This is a case in which the recommender repeatedly recommends the same type of item with attribute $w_j,$ and the user grows to prefer that item over any other item over time. A key difference from the $a=0$ case is that multiple equilibria can exist for the case where $a$ is large. Further, the initial preferences $\psi(0)$ play a role in determining the equilibria. This is because the recommendation algorithm relies on initial user activity (determined by initial user preferences) to determine which item(s) the user prefers more.

The edge cases discussed above provide insight into the role of the exploration-exploitation parameter $a$ in determining the equilibria of the asymptotic ODE. The following theorem elaborates on the relation between $a$ and the set of equilibria of user preferences.
\begin{theorem}
  Let $\mathcal{Y}_a$ be the set of all equilibrium points of $\psi$ for the ODE with the recommendation parameter $a.$ Let $g: \R^K \to \R^K$ be the generalized argmax function over $K$-dimensional vectors i.e.
  \begin{equation*}
    g(x) = \frac{1}{|J(x)|}\sum_{j \in J(x)} x_j ;\quad J(x) = \{j \in [K]: x_j = \max_{k \in [K]} x_k\}
  \end{equation*}
  and let $\tilde{W}$ be the set of solutions to the equation $y = W g(W^{\top} y).$ Then
  % \begin{equation*}
  %   %\tilde{W} = \{w \in \conv(W) \ |\ \exists y \in \R^q: w^{\top}y = \max_{x \in \conv(W)} x^{\top}y \}. \text{ Then}
  %   \tilde{W} = \{y \in \R^q | y = W \arg \max(W^{\top}y) \}
  % \end{equation*}
  \begin{enumerate}
    \item If $a < \frac{2}{\|W\|^2},$ then $|\mathcal{Y}_a|=1.$ Else, $|\mathcal{Y}_a| \geq 1.$
    \item Consider a increasing sequence of reals i.e. $a_n \to \infty$ as $n \to \infty,$ and let $(y_{a_n})$ be any sequence of equilibrium points such that $y_{a_n} \in \mathcal{Y}_{a_n}.$ Then the limit points of all convergent subsequences of $(y_{a_n})$ belong to the set $\tilde{W}.$
  \end{enumerate}
\end{theorem}

\begin{proof}
  \textbf{Part 1:} %TODO: Finish part 1 of proof
  Recall that all points $y_a \in \mathcal{Y}_a$ satisfy
  \begin{equation*}
    y_a = \sum_{k=1}^K \frac{\exp(a w_k^{\top} y_a)}{\sum_{j=1}^K \exp(a w_j^{\top} y_a)} w_k.
  \end{equation*}
  Define $h: \R \times \R^d \to \R$ where
  \begin{equation*}
    h(a,y) = \frac{1}{a} \log \left( \sum_{k=1}^K \exp(a w_k^{\top} y) \right) - \frac{\|y\|^2}{2}.
  \end{equation*}
  Observe that all points $y_a \in \mathcal{Y}_a$ are solutions to the equation $\nabla_y h(a,y) = 0,$ i.e. they are the critical points of the function $h.$ A sufficient condition for the existence of a unique critical point is the strict convexity/concavity of $h(a,y)$ with respect to $y.$ \\
  In the following, we show that $h(a,y)$ is strictly concave in $y$ for a range of $a.$ A sufficient condition for $h$ to be strictly concave is for $\nabla_y^2 h$ to be negative definite everywhere. From the definition for $h,$ we observe that
  \begin{equation*}
    \nabla^2_y h(a,y) = a W \left( \diag(\pi(y)) - \pi(y) \pi(y)^{\top} \right) W^{\top} - I
  \end{equation*}
  For negative definiteness, we check if $v^{\top} \nabla^2_y h v$ is negative for all $v \in \R^d \setminus \{0\}.$
  \begin{align*}
    v^{\top} \nabla^2_y h(a,y) v
    &= v^{\top} \left( a W \left( \diag(\pi(y)) - \pi(y) \pi(y)^{\top} \right) W^{\top} - I \right) v \\
    &= a v^{\top} \left( W \left( \diag(\pi(y)) - \pi(y) \pi(y)^{\top} \right) W^{\top} \right) v - \|v\|^2 \\
    &\leq a v^{\top} \left( \left\| W \left( \diag(\pi(y)) - \pi(y) \pi(y)^{\top} \right) W^{\top} \right\| v \right) - \|v\|^2 \\
    &\leq a \| W \| \| \diag(\pi(y)) - \pi(y) \pi(y)^{\top} \| \| W^{\top}\| \|v\|^2 - \|v\|^2 \\
    &\leq ( a \| W \|^2 \| \diag(\pi(y)) - \pi(y) \pi(y)^{\top} \| - 1 ) \|v\|^2.
  \end{align*}
  Note that $\|\cdot\|$ denotes the operator norm of a matrix over the Euclidean norm. Since $\|v\|^2>0$ for all $v \in \R^d,$ it suffices to show that $a \|W\|^2 \|\diag(\pi(y)) - \pi(y)\pi(y)^{\top}\| - 1 < 0$ to prove that $\nabla_y^2 h$ is negative definite. We next simplify the matrix norm $\|\diag(\pi(y)) - \pi(y)\pi(y)^{\top}\|,$ which is given by
  \begin{align*}
    \|\diag(\pi(y)) - \pi(y)\pi(y)^{\top}\| = \left\|
    \begin{bmatrix}
      \pi_1(y) - \pi_1^2(y) & -\pi_1(y)\pi_2(y) & \cdots & -\pi_1(y)\pi_K(y) \\
      -\pi_2(y) \pi_1(y) & \pi_2(y) - \pi_2^2(y) & \cdots & -\pi_2(y)\pi_K(y) \\
      \vdots & \vdots & \ddots & \vdots \\
      -\pi_K(y) \pi_1(y) & -\pi_K(y) \pi_2(y) & \cdots & \pi_K(y) - \pi_K^2(y) \\
    \end{bmatrix}
    \right\|
  \end{align*}
  Note that the matrix norm is the largest eigenvalue of the matrix. Due to Gershgorin's disc theorem, all eigenvalues of the matrix lie on the union of circles in the complex plane centered at the diagonal element having a radius equal to the sum of the absolute values of the off-diagonal elements. From the theorem, we obtain an upper bound on the largest eigenvalue to be
  \begin{align*}
    \|\diag(\pi(y)) - \pi(y)\pi(y)^{\top}\| &\leq \max_{k \in [K]} \left( \pi_k(y) - \pi_k^2(y) + \sum_{j \in [K];j \neq k} |-\pi_k(y)\pi_j(y)| \right) \\
    &= \max_{k \in [K]} \left( \pi_k(y) (1 - \pi_k(y)) + \sum_{j \in [K];j \neq k} \pi_k(y)\pi_j(y) \right) \\
    &= \max_{k \in [K]} \left( \pi_k(y) (1 - \pi_k(y)) + \pi_k(y) \sum_{j \in [K];j \neq k} \pi_j(y) \right) \\
    &= \max_{k \in [K]} \left( \pi_k(y) (1 - \pi_k(y)) + \pi_k(y) (1 - \pi_k(y)) \right) \\
    &= \max_{k \in [K]}  2 \pi_k(y) (1 - \pi_k(y)).
  \end{align*}
  Substituting this into the sufficient condition gives the following condition
  \begin{align*}
    0 &> a \|W\|^2 \max_{k \in [K]}  2 \pi_k(y) (1 - \pi_k(y)) - 1 \\
    \implies a &< \frac{1}{2 \|W\|^2 \max_{k \in [K]} \pi_k(y) (1 - \pi_k(y))}.
  \end{align*}
  This sufficient condition must hold for all $y.$ To obtain a loose bound, one can use the fact that $\max_{k \in [K]} \pi_k(y) (1 - \pi_k(y)) \leq \frac{1}{4}$ because $\pi_k(y) \in [0,1].$ This gives us
  \begin{equation*}
    a < \frac{1}{2 \|W\|^2 \frac{1}{4}} \implies a < \frac{2}{\|W\|^2}.
  \end{equation*}
  This concludes the proof. Note that one can obtain a tighter bound by finding the maximum value that $\pi_k(y)$ can take if $y \in \conv(w_1,w_2,\ldots,w_K).$ This method checks for concavity of the function inside the convex hull of the item attributes, which is sufficient because all the equilibria are guaranteed to lie within the convex hull.
  % where the last step is due to Holder's inequality $u^{\top} v \leq \|u\|_1 \|v\|_{\infty}.$ Observe that $\sum_k \pi_k(y) = 1$ and $\min_y \sum_k \pi_k^2(y) = \frac{1}{K}.$ Using these, and the Cauchy Schwarz inequality, we get
  % \begin{align*}
  %   v^{\top} \nabla^2_y h(a,y) v
  %   &\leq a \left( 1 - \frac{1}{K} \right) \max_{k \in [K]} (w_k^{\top} v)^2 - \|v\|^2 \\
  %   &\leq a \left( 1 - \frac{1}{K} \right) \max_{k \in [K]} \|w_k\|^2 \|v\|^2 - \|v\|^2 \\
  %   &\leq \|v\|^2 \left( a \left( 1 - \frac{1}{K} \right) \max_{k \in [K]} \|w_k\|^2 - 1 \right)
  % \end{align*}
  % For negative definiteness, the upper bound derived above must be strictly less than 0. Hence
  % \begin{align*}
  %   0 &> a \left(1-\frac{1}{K}\right) \max_{k \in [K]} \|w_k\|^2 - 1 \\
  %   \implies a &< \frac{K}{(K-1) \max_{k \in [K]} \|w_k\|^2}
  % \end{align*}
  % Hence, for all $a > 0$ that satisfy the above condition, the function $h(a,y)$ is strictly concave in $y$ and consequently has a unique critical point. This, in turn, implies the existence of a unique solution in the set $\mathcal{Y}_a$ for all $a$ under consideration.
  
  \textbf{Part 2:} Define a function $f:\R \times \R^d \to \R^d$ such that
  \begin{equation*}
    f(a,y) = W \pi_a(y) = \sum_{k=1}^K \frac{\exp(a w_k^{\top} y)}{\sum_{j=1}^K \exp(a w_j^{\top} y)} w_k.
  \end{equation*}
  Observe that all points in the set $\mathcal{Y}_a$ are solutions to the equation $y=f(a,y).$ Then, given an increasing sequence of reals $(a_n),$ construct a sequence of sets $(\mathcal{Y}_{a_n}),$ and a sequence of vectors $(y_{a_n})$ such that $y_{a_n} \in \mathcal{Y}_{a_n}.$ Construct another sequence of vectors $(z_n)$ such that  $z_n := y_{a_n} - W \arg \max (W^{\top} y_{a_n}).$

  We will now show that $\lim_{n \to \infty}\|z_n\| = 0,$ thereby showing that $z_n$ converges to the origin. First observe that
  \begin{align*}
    \| z_n \| &= \| f(a_n, y_{a_n}) - W \arg \max (W^{\top} y_{a_n}) \| \\
              & \leq \sup_{y \in (y_{a_n})} \| f(a_n,y) - W \arg \max (W^{\top} y) \| \\
    \implies \lim_{n \to \infty} \| z_n \| &\leq \lim_{n \to \infty} \sup_{y \in (y_{a_n})} \| f(a_n,y) - W \arg \max (W^{\top} y) \| = 0 \\
    \implies \lim_{n \to \infty} \| z_n \| &= 0.
  \end{align*}
  The penultimate step is due to the pointwise convergence property of the softmax function to the argmax function, i.e., %TODO: Cite the pointwise convergence property
  \begin{equation*}
      \lim_{a \to \infty} f(a,y) = W \arg \max (W^{\top}y) \quad \forall y \in \R^d.
  \end{equation*}
  % \begin{align*} %TODO: Fix the math. Use limits of norms, pointwise convergence and sandwich theorem.
  %   \lim_{a \to \infty} f(a,y) &= W \arg \max (W^{\top}y) \quad \forall y \in \R^q \\
  %   \implies \exists N, \epsilon :\quad |f(a_n,y) - W \arg \max(W^{\top} y) | &< \epsilon \quad \forall n > N, y \in \R^q \\
  %   \implies \exists N, \epsilon :\quad | y_{a_n} - W \arg \max(W^{\top} y_{a_n}) | &< \epsilon \quad \forall n > N \\
  %   \implies \lim_{n \to \infty} y_{a_n} - W \arg \max (W^{\top} y_{a_n}) &= 0
  %   %\lim_{a \to \infty} f(a,y) &\in \tilde{W} \quad \forall y \in \R^q
  % \end{align*}
  Hence, for every increasing sequence of positive reals $(a_n),$ there exists a sequence of vectors $(z_n),$ where $z_n := y_{a_n} - W \arg \max (W^{\top} y_{a_n}),$ that converges to the origin.
  Let $\tilde{W}$ be the set of solutions to the equation $y = W \arg \max(W^{\top}y).$
  Because $\lim_{n \to \infty} z_n = 0,$ as shown above, all points $y_{a_n} \in \mathcal{Y}_{a_n}$ are arbitrarily close to $\tilde{W}$ as $n \to \infty.$
\end{proof}

\begin{figure}[tb]
  \centering
  \includegraphics[width=0.45\linewidth]{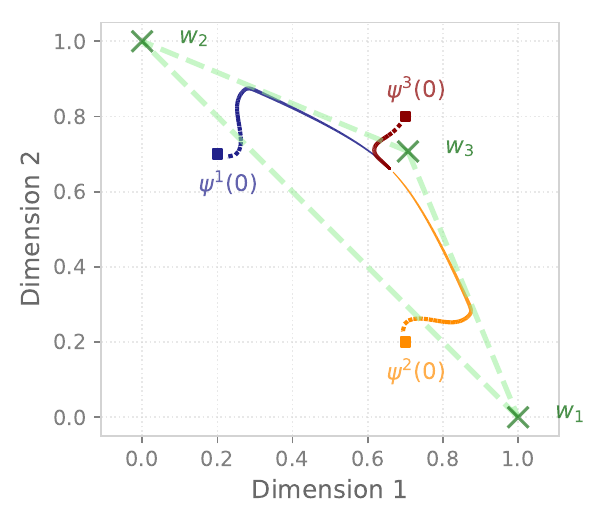}
  \includegraphics[width=0.45\linewidth]{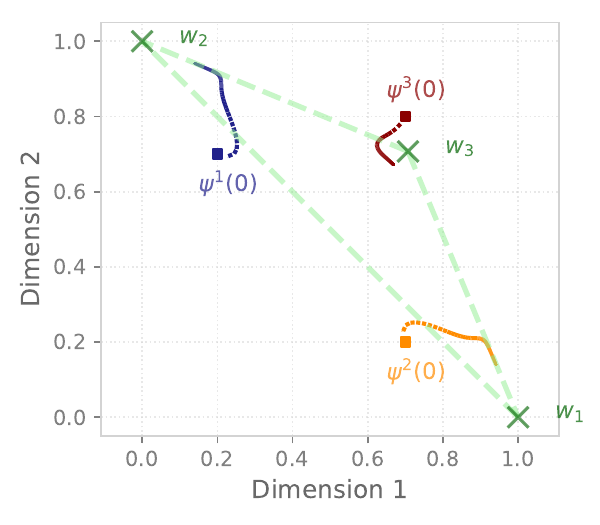}
  \caption{Effect of the value of $a$ on the set of equilibria.}
  \label{fig:level-2-effect-of-a}
\end{figure}

% TODO: Discuss the properties of the equilibrium points (effect of recommendation parameter a) using figures and examples.
This result points to the existence of a critical value for $a$ that defines two different regimes of model properties. If $a$ falls below this critical value then the set of equilibria is a singleton set, and if $a$ is above the critical value, then the set of equilibria is greater than 1. In the small $a$ regime, the effect of exploration dominates that of exploitation, and the initial preferences have no relation with the long term preferences. When $a$ is large, the effect of exploitation dominates that of exploration, and user preferences shift towards the item that \RS recommends aggresively. Initial preferences play a role in determining the item that \RS thinks is best, and consequently affect the long term preferences. The value of $a$ qualitatively changes the properties of the underlying system, and such a phenomenon is known as a \emph{bifurcation} in dynamical systems theory.

Consider the example shown in \cref{fig:level-2-effect-of-a} that plots multiple trajectories of the asymptotic ODE given different initial values. These figures show three different trajectories for the user preference vector $\psi^1(\tau),\psi^2(\tau),\psi^3(\tau),$ each starting with different initial values given by $\psi^1(0)=[0.2~0.7]^{\top},$ $\psi^2(0)=[0.7~0.2]^{\top},$ and $\psi^3(0)=[0.7~0.8]^{\top}$ respectively. When $a=0.7$ (left), the user preferences converge to the only equilibrium point of the asymptotic ODE. When $a=0.8$ (right), the preferences converge to different equilibria. Here, the item attributes are $w_1=[1~0]^{\top},$ $w_2=[0~1]^{\top},$ and $w_3=[\frac{1}{\sqrt{2}}~\frac{1}{\sqrt{2}}]^{\top}.$ The green dotted line plots the convex hull of all the item attributes. For $\theta$ and $q,$ the initial values for all the above simulations are $\theta(0)=0$ and $q(0)=\frac{1}{K}\mathbf{1}_K$ respectively. The figure on the left plots trajectories of $\psi(\tau)$ for the model with $a=7,$ and the figure on the right plots the same after changing the value of $a$ from 7 to 8 while keeping everything else unchanged. It is observed that the multiple trajectories of $\psi(\tau) $ converge to the same point irrespective of the initial value when $a=7.$ However, when $a=8,$ the trajectories converge to different points depending on the initial value.

%TODO: Talk about the effect of $\rho$
\subsection{Effect of the rate of change of user preferences}
\label{ssec:level-2-effect-of-rho}
Another important parameter that influences the behavior of the trajectories is $\rho.$ Recall that $\rho$ is the ratio of the rates of change of the user preferences and the algorithm state in the long term, i.e.,
\begin{equation*}
  \rho = \lim_{t \to \infty} \frac{\beta_t}{\frac{1}{t+1}}.
\end{equation*}
$\rho$ appears in the part of the asymptotic ODE that describes how $\psi(\tau)$ changes. While $\rho$ affects the trajectories of the asymptotic ODE, $\rho$ does not affect the set of equilibria.
\begin{figure}[tb]
  \centering
  \includegraphics[width=0.45\linewidth]{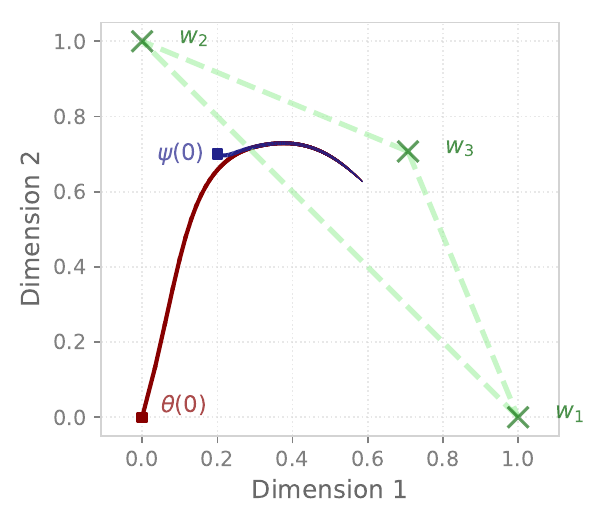}
  \includegraphics[width=0.45\linewidth]{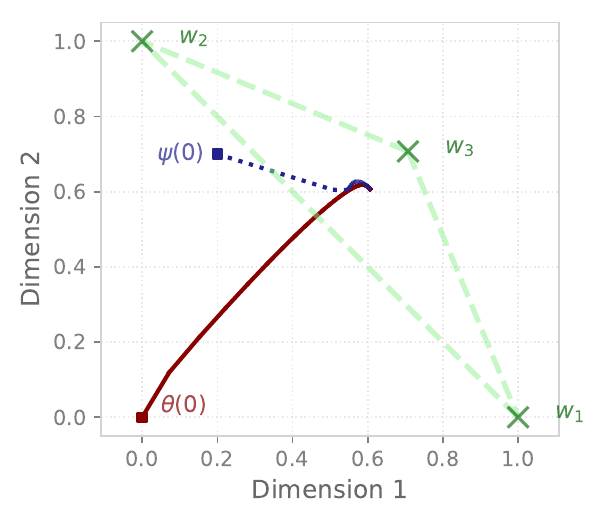}
  \caption{Effect of $\rho$ on the trajectories of the asymptotic ODE.}
  \label{fig:level-2-effect-of-rho}
\end{figure}
Consider the trajectories of the asymptotic ODE plotted in \cref{fig:level-2-effect-of-rho}. The simulation parameters are identical to those used to plot \cref{fig:level-2-example}, except for the change in $\rho.$ When $\rho=0.01$ (left of \cref{fig:level-2-effect-of-rho}), the rate of change of the algorithm state is much larger than that of the user preferences. In such a case, the trajectory of $\theta$ is observed to overlap with the trajectory of $\psi,$ indicating that the algorithm state is closely ``tracking'' the user preferences. When $\rho=10,$ (right of \cref{fig:level-2-effect-of-rho}), the rate of change of algorithm state is smaller than that of the user preferences. The trajectories of both $\theta$ and $\psi$ appear much different than what was observed in the previous case. However, in both cases, the trajectories appear to converge to the same point. This observation indicates that the relative rates of change of user preferences and the algorithm state has little effect on the set of limit points of the trajectories.

\section{Summary}
In this chapter, we introduced a model of interaction between \RS and a user whose preferences are sensitive to the recommendations of \RS. To understand the long term properties of the algorithm and the user preferences, we obtained and analyzed an ODE that has the same asymptotic properties as that of the objects of interest. We show that the trajectories of the asymptotic ODE converge to the (locally) asymptotically stable equilibria of the ODE. Using this result, we show that the algorithm state always asymptotically converges the user preferences well despite the dynamic nature of the latter. We show the effect that model parameters have, namely $a$ and $\rho,$ over the set of equilibria of the asymptotic ODE. When \RS explores more, the user preferences evolve independent of their initial state, and when \RS exploits more, users grow to develop strong preferences towards aggressively suggested recommendations. We also show that the relative rate of change between the algorithm and the user has no effect on the set of all possible long term preferences.

In the next chapter, we discuss the interaction of \RS with multiple users. Many results from the current chapter carry over with the additional complexity introduced by probabilities of user arrivals and side information about users that \RS uses for personalization.

%===========================
% Chapter 4
\chapter{Model of Interaction with Multiple Users}
\label{ch:multi-user-model}

In this chapter, we model the interaction of a recommender system (RS) with multiple users. To do so, we extend the model described in \cref{ch:one-user-model} to include multiple users. We use the modeling and analysis method developed in the preceding chapters to characterize the long term behavior of the algorithm state and the users' preferences. We show that the algorithm state can accurately converge to all users' preferences in the long term and discuss sufficient conditions for the same. Moreover, it is observed that users' preferences influence each other through the RS. In particular, users that use the RS more often tend to influence the preferences of other users.

The model of interaction is described in \cref{sec:level-3-model-description}, following which the asymptotic ODE of the model is obtained in \cref{sec:level-3-asymptotic-ODE}. The limit sets of the algorithm state and the user preferences are understood by analyzing the equilibria of the asymptotic ODE in \cref{sec:level-3-model-analysis}.
\section{Model Description}
\label{sec:level-3-model-description}
We consider a sequential, discrete-time model of interaction between a recommender system \RS and $N$ of its users, in which the preferences of each of these users is influenced by recommendations. The preferences of user $n \in [N]$ are characterized by the $d$-dimensional vector $\psi_{n,t}$ at time $t.$ We build upon the model discussed in the previous chapters.

At any time $t,$ a user $U_t \in [N]$ arrives at \RS to receive recommendations. \RS recommends item $A_t \in [K]$ to the user, and the user responds to the recommendation with a reward $R_t.$ Each item $k$ is associated with a $d$-dimensional item attribute $w_k,$ and let $W=
\begin{bmatrix}
  w_1 & w_2 & \cdots & w_K
\end{bmatrix}
$ denote the $d \times K$ attribute matrix with $\rank(W)=d.$ The reward depends on the preferences of the user $U_t$ and recommendation $A_t$ as follows.
\begin{equation}
  \label{eq:level-3-reward}
  R_t = w_{A_t}^{\top} \psi_{U_t,t} + \epsilon_t
\end{equation}
Here, $w_{A_t}$ is the item attribute of the recommendation $A_t,$ $\psi_{U_t,t}$ is the preference vector of the user $U_t,$ and $(\epsilon_t)$ is zero-mean, fixed-variance sequence of noise vectors that are conditionally independent of other elements in the sequence. Furthermore, each elements $\epsilon_t$ is distributed identically to $\epsilon$ (see \ref{eq:reward-learner-model} for the exact definition). \RS and the users are affected by this interaction: \RS uses the reward to improve subsequent recommendations, and the preferences of user $U_t$ is influenced by the recommendation $A_t.$ In the following, we discuss the models of \RS and the users in greater detail.

\paragraph{Recommendation policy:}
\RS is able to observe user $U_t$ at time $t$ and uses that information to make personalized recommendations. In particular, \RS associates each user $n$ with a $p$-dimensional user attribute $v_n.$ Let $V :=
\begin{bmatrix}
  v_1 & v_2 & \cdots & v_N
\end{bmatrix}
$ denote the $p \times N$ matrix containing all the user attributes, such that $\rank(V)=p.$ These assumptions on the user attributes are useful for \RS. To see this, consider the case in which $rank(V)<p.$ In this case, \RS is better off using a full-rank user attribute matrix $V'$ obtained by row transformations on $V.$ The matrix $V'$ would have fewer parameters than $V,$ which makes $V'$ more desirable. Another useful assumption made implicitly is that $N \geq p.$ This is because if $N < p,$ then \RS is better off using an $N \times N$ identity matrix $I_N$ as the user attribute matrix. This is because $I_N$ has fewer parameters than $V$ if $N<p.$

Using these attributes, \RS defines a \textit{context} for each user-item combination. We define the context $c_{nk}$ of user $n$ and item $k$ as
\begin{equation}
  \label{eq:context}
  c_{nk} = v_n \otimes w_k,\qquad C = V \otimes W.
\end{equation}
Here, $C$ is a $dp \times NK$ matrix that contains all the contexts. Observe that $C$ is a full-rank matrix ($\rank(C)=dp$) because both $V$ and $W$ are full-rank matrices.
\RS makes recommendations using these contexts. At time $t,$ \RS uses a policy $\pi_n(\theta_t),$ parameterized by the algorithm state $\theta_t \in \R^{pd},$ to make recommendations to user $n$ at time $t,$ where
\begin{equation*}
  \pi_n(\theta) :=
  \begin{bmatrix}
    \pi_{n1}(\theta) \\ \pi_{n2}(\theta) \\ \vdots \\ \pi_{nK}(\theta)
  \end{bmatrix},
  \qquad
  \pi_{nk}(\theta) := \frac{\exp(a \ c_{nk}^{\top} \theta)}{\sum_{j=1}^K \exp (a \ c_{nj}^{\top} \theta)}.
\end{equation*}
Here, $\pi_{nk}(\theta_t)$ is the probability of recommending item $k$ to user $n$ at time $t.$

\paragraph{Learning algorithm}
The goal of the learner is to maximize future rewards. To do so, the algorithm state is updated at each time step to minimize the following regularized least squares loss
\begin{equation}
  \theta_{t+1} =  \arg \min_{\theta} \zeta \|\theta\|_{CC^{\top}}^2 + \sum_{s=1}^t \left( R_s - c_{U_s A_s}^{\top} \theta \right)^2.
  \label{eq:level-3-algorithm-loss}
\end{equation}
The preceding loss is strictly convex, which implies that a unique minimizer exists for the loss. Hence, the following closed form expression is obtained for the algorithm state $\theta_t$ for all $t>0,$
\begin{equation}
  \theta_{t+1} = \left( C \diag{Q_{t+1} C^{\top}} \right)^{-1} \left( \frac{1}{t+1} \sum_{s=1}^{t}c_{U_s A_s} R_s \right) ;\quad \theta_1 = 0.
  \label{eq:level-3-learner-defn}
\end{equation}
Here, the $NK$-dimensional vector $Q_t$ is the sum of a regularization parameters and the recommendation history until time $t-1.$ For all $n \in [N]$ and $k \in [K],$ the $(K(n-1)+k)$-th element of the vector $Q_t,$ denoted by $Q_t^{nk},$ is given by
\begin{equation}
  Q_{t}^{nk} := \frac{1}{t} \left( \lambda + \sum_{s=1}^{t-1} [U_s = n] [A_s = k] \right)
  \label{eq:level-3-learner-q-defn}
\end{equation}
$Q_t$ can be considered as an auxiliary algorithm state, because it is used to obtain the algorithm state of interest $\theta_t.$

\paragraph{User arrivals and user preferences}
Each user $n \in [N]$ arrives at \RS with probability $P(U_t=n) = \lambda_n.$ We consider a model in which exactly one user arrives at \RS at any given time, hence we have $\sum_{n \in [N]} \lambda_n = 1.$ The initial preferences of users (preferences at time $t=1$) are given by the $d$-dimensional vectors $\{\psi_{1,1},\psi_{2,1},\ldots,\psi_{N,t}\}.$ Further, the preferences of user $U_t$ are influenced by the recommendation $A_t.$ The preferences are affected such that the user will find the recommendation $A_t$ more appealing in the future. In particular, the dynamics of the preferences of any user $n \in [N]$ is characterized by the recursive equation
\begin{equation}
  \label{eq:level-3-single-user-recursion}
  \psi_{n,t+1} = \psi_{n,t} + \beta_t [U_t = n] (w_{A_t} - \psi_{n,t}).
\end{equation}
Note that the preferences of users that are not receiving recommendations from \RS remain unaffected.

\section{Obtaining the Asymptotic ODE}
\label{sec:level-3-asymptotic-ODE}
In this section, we obtain an asymptotic ODE for the current system in order to learn about the long term properties of the system. We refer the reader to \cref{sec:level-1-asymptotic-ode} for an elaborate discussion on the asymptotic ODE and the steps used to obtain such an ODE. The process broadly consists of three steps: obtaining recursive relations satisfied by the states of interest (algorithm state $(\theta_t,Q_t)$ and user preferences $\psi_t$), rewriting the recursions as stochastic approximation (SA) recursions, and using the recursions to determine the asymptotic ODE. In the following, we follow the steps to obtain an asymptotic ODE for the current system.

From \cref{eq:level-3-learner-defn,eq:level-3-learner-q-defn}, it can be shown that the algorithm states satisfy the recursive relations
\begin{align}
    \theta_{t+1} &= \theta_t + \frac{1}{t+1} \left( C \diag(Q_{t+1}) C^{\top} \right)^{-1} \left( c_{U_t A_t} R_t - c_{U_t A_t} c_{U_t A_t}^{\top} \theta_t \right) \label{eq:level-3-learner-recursion},\\
  Q_{t+1}^{nk} &= Q_t^{nk} + \frac{1}{t+1} \left( [U_t = n] [A_{t} = k] - Q_t^{nk} \right) \label{eq:level-3-learner-q-recursion}.
  %& \vdots \\
  %\psi_{K,t+1} &= \psi_{K,t} + [U_t = K] \beta_t (w_{A_t} - \psi_t).
\end{align}
We refer the reader to \cref{ssec:level-1-obtaining-recursions} for details on how these recursions are obtained.
Now, we derive a recursion for the user preferences. Using the recursion from \cref{eq:level-3-single-user-recursion} that is defined for the preferences of any user $n,$ we can express the preference dynamics of all $N$ users using a single recursive relation. Define a $Nd$-dimensional vector $\psi_t \in \R^{Nd}$ that represents all user preferences at time $t.$ The dynamics of all the user preferences are characterized by the recursive equation
\begin{equation}
  \psi_t :=
  \begin{bmatrix}
    \psi_{1,t} \\ \psi_{2,t} \\ \vdots \\ \psi_{K,t}
  \end{bmatrix}
  ;\quad
  \psi_{t+1} = \psi_t + \beta_t
  \begin{bmatrix}
    [U_t = 1] (w_{A_t} - \psi_{1,t}) \\
    [U_t = 2] (w_{A_t} - \psi_{2,t}) \\
    \vdots \\
    [U_t = N] (w_{A_t} - \psi_{N,t}) \\
  \end{bmatrix}.
  % \label{eq:level-3-psi-defn}
  \label{eq:level-3-user-recursion}
\end{equation}
% Let $e_n \in \R^N$ denote a $N$-dimensional unit vector that has 1 as its $n$-th element and is zero everywhere else. Further, let $\mathbf{1}_q \in \R^q$ be a $q$-dimensional vector containing all 1's, and let $I_q \in \R^{q \times q}$ denote the $q$-dimensional identity matrix. Then the recursion in \cref{eq:level-3-psi-defn} can be rewritten as
% \begin{equation}
%   \psi_{t+1} = \psi_t + \beta_t (\diag(e_{U_t}) \otimes I_q) (\mathbf{1}_q \otimes w_{A_t} - \psi_t)
% \end{equation}

% \begin{equation}
%   \psi_t :=
%   \begin{bmatrix}
%     \psi_{1,t} \\ \psi_{2,t} \\ \vdots \\ \psi_{K,t}
%   \end{bmatrix}
%   ;\quad\quad
%   \psi_{t+1} = \psi_t + \beta_t e_{U_t} \otimes (w_{A_t} - \psi_t).
% \end{equation}
% Here, $e_{U_t} \in \R^N$ is a $N$-dimensional unit vector with 1 at index $U_t$ and 0 everywhere else.
The time-evolution of the system is characterized by the recursions in \cref{eq:level-3-learner-recursion,eq:level-3-learner-q-recursion,}, along with \cref{eq:level-3-user-recursion} that defines the recursion followed by the preferences of all users $\psi_t.$

%\subsection{Obtaining stochastic approximation equations}
%TODO: Explain how these expressions are obtained
To obtain the desired stochastic approximation recursions, we use the analysis method that was used in the previous chapters. First, define $\Lambda \in \R^{N \times N}$ as the diagonal matrix containing arrival probabilities of all the users, and let $\pi(\theta)$ be a $NK$-dimensional vector created by concatenating all the probability vectors $\pi_n(\theta)$ used for recommendation. These terms are defined as follows:
\begin{equation}
  \Lambda := \diag(\lambda_1,\lambda_2,\ldots,\lambda_N),
  \qquad 
  \pi(\theta) :=
  \begin{bmatrix}
    \pi_{1}(\theta) \\ \pi_{2}(\theta) \\ \vdots \\ \pi_{N}(\theta)
  \end{bmatrix}.
  \label{eq:probs-vec-notation}
\end{equation}
Following a method similar to \cref{ch:recommendation-model}, we obtain functions $h_{\theta}$ and $h_q$ that are used to define the asymptotic ODEs for the algorithm state.
\begin{align}
    h_{\theta}(\theta,Q,\psi) &:= (C \diag(Q) C^{\top})^{-1} C \Lambda_K \diag(\pi(\theta)) (W_N^{\top} \psi - C^{\top}\theta) \label{eq:level-3-learner-mean-ode} \\
  h_q(\theta, Q, \psi) &:= \Lambda_K \pi(\theta) - Q \label{eq:level-3-learner-q-mean-ode}
\end{align}
where $W_N := I_N \otimes W$ and $\Lambda_K := \Lambda \otimes I_K.$
Now, we obtain a stochastic approximation recursion for the user preferences. From Eq.~\eqref{eq:level-3-user-recursion}, we obtain an expression for the expected deviation in user preferences
\begin{align*}
  \frac{1}{\beta_t} \Exp{\psi_{t+1}-\psi_t|\theta_t,Q_t,\psi_t} &= \Exp{  \begin{bmatrix}
    [U_t = 1] (w_{A_t} - \psi_{1,t}) \\
    \vdots \\
    [U_t = N] (w_{A_t} - \psi_{N,t})
  \end{bmatrix} \Bigg| \theta_t,Q_t,\psi_t}.
\end{align*}
Conditioning on $U_t,$ the user arrival at time $t,$ helps simplify this expression.
\begin{align*}
  \frac{1}{\beta_t} \Exp{\psi_{t+1}-\psi_t|\theta_t,Q_t,\psi_t} &= \begin{bmatrix}
    P(U_t = 1) \Exp{w_{A_t} - \psi_{1,t} | \theta_t,Q_t,\psi_t, U_t=1} \\
    \vdots \\
    P(U_t = N) \Exp{w_{A_t} - \psi_{N,t} | \theta_t,Q_t,\psi_t, U_t=N}
  \end{bmatrix}.
\end{align*}
Using the fact that $\lambda_n=P(U_t=n),$ we simplify the expression further.
\begin{align*}
  \frac{1}{\beta_t} \Exp{\psi_{t+1}-\psi_t|\theta_t,Q_t,\psi_t} &= \begin{bmatrix}
    \lambda_1 \left( \sum_{k=1}^K P(A_t=k|\theta_t,U_t=1) w_k - \psi_{1,t} \right) \\
    \vdots \\
    \lambda_N \left( \sum_{k=1}^K P(A_t=k|\theta_t,U_t=N) w_k - \psi_{N,t} \right)
  \end{bmatrix} \\
  &= \begin{bmatrix}
    \lambda_1 \left( W \pi_1(\theta_t) - \psi_{1,t} \right) \\
    \vdots \\
    \lambda_N \left( W \pi_N(\theta_t) - \psi_{N,t} \right)
  \end{bmatrix}.
\end{align*}
We express the preceding expectation compactly in vector notation, using terms defined in \cref{eq:probs-vec-notation}.
\begin{equation*}
    \frac{1}{\beta_t} \Exp{\psi_{t+1}-\psi_t|\theta_t,Q_t,\psi_t} = (\Lambda \otimes I_d) \left( (I_N \otimes W) \pi(\theta_t) - \psi_t \right).
\end{equation*}
Here, $I_d$ and $I_N$ are identity matrices of orders $d$ and $N$ respectively. We define $\Lambda_d = \Lambda \otimes I_d.$ The preceding expression for the expected deviation of user preferences motivates the definition of $h_{\psi},$ which is the function that characterizes the asymptotic ODE, i.e.,
\begin{equation}
  \label{eq:level-3-user-mean-ode}
  h_{\psi}(\theta,q,\psi) := \Lambda_d( W_N \pi(\theta) - \psi ).
\end{equation}
We are now ready to state the theorem connecting the asymptotic ODE to the model of interaction.
\begin{theorem}
  When the assumption given by Eq.~\eqref{eq:step-size-asymptotic-assumption} is satisfied, the sequence $(\theta_t, Q_t, \psi_t)$ converges to a connected internally chain recurrent set of the ODE
  \begin{equation}
    \begin{split}
      \dot{\theta}(\tau) &= h_{\theta}(\theta,q,\psi) \\
      \dot{q}(\tau) &= h_{q}(\theta,q,\psi) \\
      \dot{\psi}(\tau) &= \rho h_{\psi}(\theta,q,\psi)
    \end{split}
    \label{eq:level-3-ode}
  \end{equation}
  \label{thm:level-3-ode}
\end{theorem}
\begin{proof}
  For the sake of avoiding repetition, we refer the reader to the proof of Theorem~\ref{thm:level-2-ode}. The proof of the current result is very similar to that of Theorem~\ref{thm:level-2-ode}.
\end{proof}
This result related the asymptotic behavior of the sequence of states $(\theta_t,Q_t,\psi_t)$ to that of the trajectories of the ODE $(\theta(\tau),q(\tau),\psi(\tau)).$ We now proceed to use the asymptotic ODE, a deterministic model for which analysis is tractable, to understand the stochastic system of interest.
\section{Long term analysis of algorithm state and user preferences}
\label{sec:level-3-model-analysis}
In this section, we focus our analysis on the asymptotic ODE. This is justified by \cref{thm:level-3-ode}: the limiting behavior of $(\theta_t,Q_t,\psi_t),$ the algorithm state and the user preferences, is identical to the limiting behavior of the trajectories of the ODE given by \cref{eq:level-3-ode}. Understanding the trajectories of the ODE helps us understand the long-term behavior of the algorithm state and user preferences.

We observe that the trajectories of the asymptotic ODE are convergent, and the points of convergence happen to be certain equilibria of the ODE. Recall that such equilibria are called asymptotically stable equilibria, and these equilibria can be locally stable or globally stable. The property of a locally asymptotically equilibrium is that the trajectories that begin sufficiently close to the equilibrium converge to the equilibrium point as time passes. Presence of such equilibria indicates that the user preferences and the algorithm's policy stabilize over time, and understanding the equilibria helps us understand their long-term properties. We refer the reader to \cref{sec:level-2-equilibrium-analysis} for a longer discussion on the significance of studying equilibrium points.
In the following, we show sufficient conditions for equilibria to be locally asymptotically stable, and discuss their implications.

\subsection{Sufficient conditions for effectively learning user preferences}
Recall that the asymptotic ODE is given by
\begin{align*}
  \dot{\theta}(\tau) &= (C \diag(Q) C^{\top})^{-1} C \Lambda_K \diag(\pi(\theta)) (W_N^{\top} \psi - C^{\top}\theta) \\
  \dot{q}(\tau) &= \Lambda_K \pi(\theta) - Q \\
  \dot{\psi}(\tau) &=  \Lambda_d( W_N \pi(\theta) - \psi ).
\end{align*}
Note that the context matrix C can be expressed as $C = V \otimes W = (V \otimes I_d) (I_N \otimes W) = V_d W_N,$ where $V_d := V \otimes I_d$ is a matrix of order $dp \times dN.$ Using this, the ODE for $\theta(\tau)$ can be written as
\begin{equation*}
  \dot{\theta}(\tau) = (C \diag(Q) C^{\top})^{-1} V_d W_N \Lambda_K \diag(\pi(\theta)) W_N^{\top} (\psi - V_d^{\top}\theta)
\end{equation*}
Observe that $W_N \Lambda_K \diag(\pi(\theta)) W_N^{\top}$ is a positive definite matrix because $W_N$ is full rank and the diagonal matrix $\Lambda_K \diag(\pi(\theta))$ is also full rank. Hence, $\dot{\theta}(\tau)=0$ happens under one of two conditions: when $\psi-V_d^{\top} \theta = 0$ or when $W_N \Lambda_K \diag(\pi(\theta)) W_N^{\top} (\psi - V_d^{\top}\theta)$ belongs to the null space of $V_d.$ The latter is a weaker condition that subsumes the former, which is a stronger condition. In the following, we focus on equilibria that satisfy the stronger condition
\begin{equation}
  \label{eq:level-3-stable-equilibrium-condition}
  \psi = V_d^{\top} \theta \implies \psi_n = \Theta v_n, \quad n \in [N]
\end{equation}
where $\Theta := \text{vec}^{-1}(\theta)$ is a $d \times p$ matrix created by rearranging the elements of the vector $\theta.$
There are two important reasons for our interest in this condition.
First, such equilibria correspond to minimum loss for the algorithm, i.e., the least possible least squares loss given by \cref{eq:level-3-algorithm-loss} is achieved in the expected sense
To see this, observe that the the regularization term in the expected least square loss in \cref{eq:level-3-algorithm-loss} scaled by $\frac{1}{t}$ becomes negligible for large $t$ and we can write the loss as
% To see this, we first see that the expected least squares loss in \cref{eq:level-3-algorithm-loss} scaled by $\frac{1}{t}$ when \cref{eq:level-3-stable-equilibrium-condition} holds. For large $t,$ the regularization term is negligible.
\begin{align*}
  \frac{1}{t} \sum_{s=1}^t \left( R_s - c_{U_s A_s}^{\top} \theta_t \right)^2
  &= \frac{1}{t} \sum_{s=1}^t \left( (w_{A_s}^{\top} \psi_{U_s,s} + \epsilon_t) - c_{U_s A_s}^{\top} \theta_t \right)^2
\end{align*}
Further, for large $t,$ we assume that the user preferences and the algorithm state are approximately equal to an asymptotically stable equilibrium point. For the equilibrium point $(\theta,\psi),$ we have $\theta_t \approx \theta$ and $\psi_t \approx \psi.$ In this case, the initial loss contributed by algorithm states $\theta_s$ and user preferences and $\psi_s$ for $s < t$ before $\theta_t$ and $\psi_t$ stabilized can also be neglected due to large $t.$ This gives us the following.
\begin{align*}
  \frac{1}{t} \sum_{s=1}^t \left( R_s - c_{U_s A_s}^{\top} \theta_t \right)^2
  &\approx \Exp{ \left( w_{A_t}^{\top} \psi_{U_t} + \epsilon_t - c_{U_t A_t}^{\top} \theta \right)^2 } \\
  &= \Exp{ \left( w_{A_t}^{\top} \psi_{U_t} - c_{U_t A_t}^{\top} \theta \right)^2 } + \Exp{\epsilon_t^2} \\
  &= \sum_{n=1}^N \sum_{k=1}^K \lambda_n \pi_k(\theta) \left( w_k^{\top} - c_{n k}^{\top} \theta \right)^2 + \Exp{\epsilon^2} \\
  &= \|W_N^{\top} \psi - C^{\top} \theta\|^2_{\Lambda_K \diag(\pi(\theta))} + \Exp{\epsilon^2} \\
  &= \|W_N^{\top} ( \psi - V_d^{\top} \theta )\|^2_{\Lambda_K \diag(\pi(\theta))} + \Exp{\epsilon^2} \\
  &= \Exp{\epsilon^2}
\end{align*}
The second step uses the fact that the conditional expectation of $\epsilon_t$ is 0, and the second-to-last step uses the condition in \cref{eq:level-3-stable-equilibrium-condition} cancel out the first term.
The above analysis shows that, when the expected loss is decomposed into bias and variance terms, the bias term is observed to be zero at the equilibrium. The loss at equilibrium constitutes only of noise in the reward.

The second desirable property of equilibria that satisfy \cref{eq:level-3-stable-equilibrium-condition} is that such equilibria are asymptotically stable. Pairing this with the property of minimum loss, the algorithm states of trajectories near this equilibrium point change such that the algorithm loss is minimized over time. This can be interpreted as the algorithm effectively learning the user preferences in the long term. The following result shows that an equilibrium point satisfying Eq.~\eqref{eq:level-3-stable-equilibrium-condition} is locally asymptotically stable.
\begin{theorem}
  The set of points
  \begin{equation*}
    \{( \theta,\pi(\theta), V^{\top}_d \theta):
    \theta \in \R^{dp}\}  
  \end{equation*}
  are equilibria to the asymptotic ODE given by \cref{eq:level-3-ode} when $y := V^{\top}_d \theta$ comprises of $d$-dimensional vectors $\{y_n\}_{n \in [N]}$ such that
  \begin{equation*}
    y =
  \begin{bmatrix}
    y_1 \\ y_2 \\ \vdots \\ y_N
  \end{bmatrix},  \quad \text{and} \quad y_n = \sum_{k=1}^K \pi_k(y_n) w_k = \sum_{k=1}^K \frac{\exp(a w_k^{\top} y_n)}{\sum_{j=1}^K \exp(a w_j^{\top} y_n)} w_k, \quad n \in [N]
\end{equation*}
\label{thm:level-3-equilibrium-stability}
\end{theorem}
% \begin{theorem}
%    Let the set $\{Y\}$ contain the solutions to the equation
%   \begin{equation*}
%     y = \sum_{k=1}^K \pi_k(y) w_k = \sum_{k=1}^K \frac{\exp(a w_k^{\top} y)}{\sum_{j=1}^K \exp(a w_j^{\top} y)} w_k
%   \end{equation*}
%   Then, the set
%   \begin{equation*}
%   \{( \theta,\pi(\theta), V^{\top}_q \theta): y =
%   \begin{bmatrix}
%     y_1 \\ y_2 \\ \vdots \\ y_N
%   \end{bmatrix},
%   y_n \in Y ~\forall n \in [N]\}  
%   \end{equation*}
%   belongs to the set of equilibrium points to the ODE~\eqref{eq:level-3-ode}. Moreover, all equilibrium points in this set are locally asymptotically stable.
% \end{theorem}
\begin{proof} %TODO
  The proof of this theorem is very similar to the proof of \cref{thm:level-2-equilibrium-stability}, with the key difference being that \cref{thm:level-2-equilibrium-stability} deals with one user, while the current result deals with $N$ users. We refer the reader to the proof of \cref{thm:level-2-equilibrium-stability} for the sake of avoiding repetition.
\end{proof}
This result relates the equilibria of the asymptotic ODE to the equilibria discussed in \cref{sec:level-2-equilibrium-analysis}. In particular, the set of possible equilibria for the user preferences is the same as that which was discussed in \cref{ssec:level-2-effect-of-a}. All the results discussed in \cref{ssec:level-2-effect-of-a} carry over to the equilibria corresponding to the preferences of each user $\psi_n(\tau).$

While the condition in \cref{eq:level-3-stable-equilibrium-condition} guarantees the asymptotic stability of a given equilibrium point, it does not tell us anything about the existence of asymptotically stable equilibria. We are interested in finding conditions on the model parameters that are sufficient for asymptotically stable equilibria to exist.
First, note that the condition in \cref{eq:level-3-stable-equilibrium-condition} need not always hold. This is because $\psi,$ which is in the LHS of \cref{eq:level-3-stable-equilibrium-condition}, belong to $\R^{Nd},$ while the RHS belongs to $\R^{pd}$ because the rank of $V_d^{\top}$ is $pd.$ This motivates us to look for cases in which the condition in \cref{eq:level-3-stable-equilibrium-condition} holds. An obvious case in which the condition can hold is when $N=p;$ in this case, both the LHS and RHS can be vectors anywhere in $\R^{Nd}.$

However, there exists another case in which \cref{eq:level-3-stable-equilibrium-condition} can hold. This case is independent of the number of users $N$ that only requires that the user attributes follow a certain condition.
The following result discusses the sufficient conditions needed for \cref{eq:level-3-stable-equilibrium-condition} to hold.

\begin{theorem}
  The following are sufficient conditions for \cref{eq:level-3-stable-equilibrium-condition} to hold:
  \begin{enumerate}
    \item $N=p,$ i.e., the number of users is the same as the size of the attribute vectors.
    \item $\{v_1,v_2,\ldots,v_N\}$ lie on a $p-1$-dimensional
      hyperplane, i.e., for any
      arbitrary ordering of vectors $\{v_{i_1}, v_{i_2}, \ldots, v_{i_N}
      \}$, there exist scalars $\{\alpha_{l 1}, \alpha_{l 2}, \ldots,
      \alpha_{l p} \}$ such that
      \begin{equation*}
        \sum_{j = 1}^p \alpha_{l j} = 1 \quad \text{and} \quad v_{i_l} = \sum_{j = 1}^p \alpha_{l j} v_{i_j}
        %\label{eq:hyperplane-condition}
      \end{equation*}
      for all $l \in \{p + 1, p + 2, \ldots, N\}$.
      In this case, the equilibrium point $(\bar{b},\bar{s},\bar{\psi})$ that satisfies condition~\eqref{eq:level-3-stable-equilibrium-condition} can also satisfy $\psi_1=\psi_2=\cdots=\psi_N.$
  \end{enumerate}
\end{theorem}
\begin{proof}
  To prove the first part of the theorem, observe that the matrix $V_d = V \otimes I_d$, which is a $p \times N$ matrix in general, is an invertible square matrix when $N=p.$ This implies that $\dot{\theta}(\tau) = 0 \implies \psi = V_d^{\top} \theta.$ This is because
  \begin{equation*}
      \dot{\theta}(\tau) = 0 \implies (C \diag(Q) C^{\top})^{-1} V_d W_N \Lambda_K \diag(\pi(\theta)) W_N^{\top} (\psi - V_d^{\top}\theta) = 0
    \end{equation*}
    and the matrices $(C \diag(Q) C^{\top})^{-1},$ $V_d,$ and $W_N \Lambda_K \diag(\pi(\theta)) W_N^{\top}$ are all non-singular square matrices. This tells us that all equilibria of the asymptotic ODE satisfy $\psi = V_d^{\top} \theta,$ and because there exists at lease one equilibrium point due to Brouwer's fixed point theorem, the condition $N=p$ is sufficient for the equilibria to satisfy $\psi = V_d^{\top} \theta.$
    
  We now prove the second part of the result. If all user attributes lie in a $(p-1)$-dimensional hyperplane, then any user attribute can be expressed as an affine combination of any other $p$ attributes.
Define ${\Theta} := \text{vec}^{- 1} ({\theta})$ as a $d \times p$ matrix that is obtained by rearranging the elements of ${\theta}.$ Then $\psi = V_d^{\top} \theta$ is equivalent to
\begin{equation*}
    \psi_n = {\Theta} v_n \quad \forall n \in [N]
\end{equation*}
The linear relation between $\psi_n$ and $v_n$ implies that all the user
states $\{ \psi_1, \psi_2, \ldots, \psi_N \}$ lie on a $p - 1$
dimensional hyperplane, which in turn means that there exist constants $\{
\alpha_{11}, \alpha_{12}, \ldots \alpha_{1 p}, \alpha_{21}, \ldots,
\alpha_{(N - p) p} \}$ such that
\begin{equation*}
 \sum_{j = 1}^p \alpha_{l j} = 1 \quad \text{and} \quad \bar{y}_{p + l} =
 \sum_{j = 1}^p \alpha_{l j}  \bar{y}_j
\end{equation*}
for all $l = 1, 2, \ldots, N - p$. From the equilibrium expression for
$\bar{y}$, we get
\begin{equation*}
W \bar{p}_{p + l} = \sum_{j = 1}^p \alpha_{l j} W \bar{p}_j \implies
W \left( \bar{p}_{p + l} - \sum_{j = 1}^p \alpha_{l j}  \bar{p}_j \right)
= 0
\end{equation*}
for all $l = 1, 2, \ldots, N - p$. The above expression holds true when
$\bar{p}_1 = \bar{p}_2 = \cdots = \bar{p}_N$.
\end{proof}
We discuss the interesting implications of this result in the following.
\paragraph{Learning recommendation-sensitive user preferences is possible.} Despite the algorithm's assumption that the user preferences are not changing with time, the algorithm state is successfully able to converge to the user preferences in the long term. We established two sufficient conditions to enable the learning of preferences. The first, is for the algorithm to adjust the user attributes such that the attribute size $p$ must equal the total number of users $N.$ Furthermore, the user attribute matrix must be full rank, i.e., $\rank(V)=N=p.$ The second sufficient condition is for all the user attributes to lie in a $(p-1)$-dimensional hyperplane. This is a surprising result because it provides a way for \RS to learn user preferences that is not affected by the number of users that are using \RS. The practical way for \RS to enforce this is to append ``1'' to the user attributes that are already available.

\paragraph{Diversity in user preferences is reduced.} This is because the set of equilibria is a much smaller set as compared to the set of possible initial states (which can be assumed to be a compact set in $\R^d$). In \cref{sec:level-2-equilibrium-analysis}, we saw that the number of possible equilibria for such asymptotic ODEs is small: one when $a$ is small, and $K$ when $a$ is very large. Furthermore, when the user attributes lie on a $(p-1)$-dimensional hyperplane, the preference vector that satisfies $\psi_1 = \psi_2 = \cdots = \psi_N$ belongs to the set of equilibria.

In the following, we discuss how various model parameters affect the set of equilibria. We pay more attention to equilibria that satisfy the relation in \cref{eq:level-3-stable-equilibrium-condition}.
% Talk about interesting equilibrium points and their stability
\subsection{Filter bubbles and polarization in user preferences}
\label{ssec:level-3-effect-of-a}
Recall the analysis from \cref{sec:level-2-equilibrium-analysis} in which the effect of the exploration-exploitation parameter $a$ on the long term user preferences is discussed. When $a$ is small, the limit set is a singleton set, and when $a$ is large, the user preferences converge to the item that is recommended aggressively to the user. The same results are carried over to each user in the current setting; this has been established in \cref{thm:level-3-equilibrium-stability}. These results take an interesting form in the setting with $N$ users. When $a$ is very small, the preferences of all users appear to become identical as a result of receiving recommendations uniformly at random. When $a$ is large, preferences of users tend to gather into groups, where each group prefers a specific item more when compared to all other items. The groups are formed as a result of personalized recommendations; \RS decides and recommends the most appropriate recommendation for each user, and the users' preferences are influenced by those recommendations. This is the well-known filter bubble phenomenon, and such phenomenon is observed to emerge as a consequence in our model of interaction.
\begin{figure}[ht]
  \centering
  \includegraphics[width=0.48\linewidth]{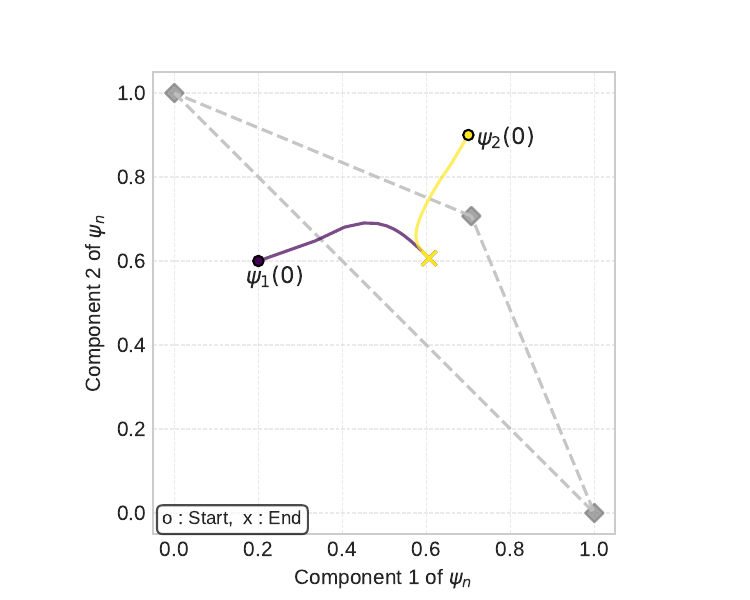}
  \includegraphics[width=0.48\linewidth]{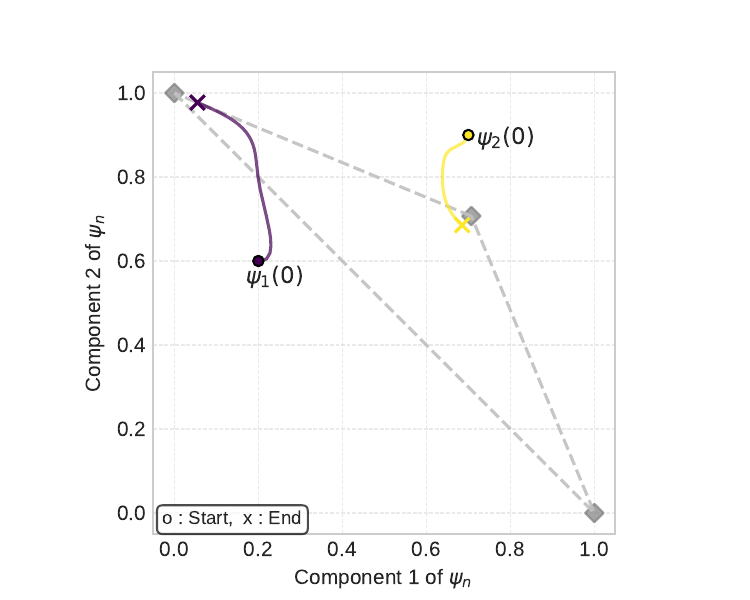}
  \caption{Effect of $a$ on preferences of $N$ users.}
  \label{fig:level-3-effect-of-a}
\end{figure}
We plot the trajectories of the preferences of two users in \cref{fig:level-3-effect-of-a}. When $a$ is small (left), the preferences of all users tend to converge to the same point. When $a$ is large, the preferences tend to converge towards attributes of items that \RS recommends them.

% \subsection{Effect of user attributes}
% \label{ssec:level-3-effect-of-V}
% They have no effect on the equilibrium set. But arrival probabilities can influence the long term preferences.

\subsection{Misidentification of user preferences}
% Discuss the effect of attributes, and the role that arrival probabilities play in misidentification.
The user attributes $\{v_1,v_2,\ldots,v_N\}$ that are used by \RS to identify users tend to impact the long term preferences, albeit indirectly. This is because the equilibrium set of user preferences is invariant for different $V.$ However, user attributes have an effect on which equilibrium point(s) the user preferences converge to. Particularly, the users whose attributes are similar (but not identical) tend to get similar recommendations if the algorithm is not able to differentiate between the attributes. As a result, their long term preferences end up being similar. We emphasize this using a numerical example.

\begin{figure}[h]
  \centering
  \includegraphics[width=0.45\linewidth]{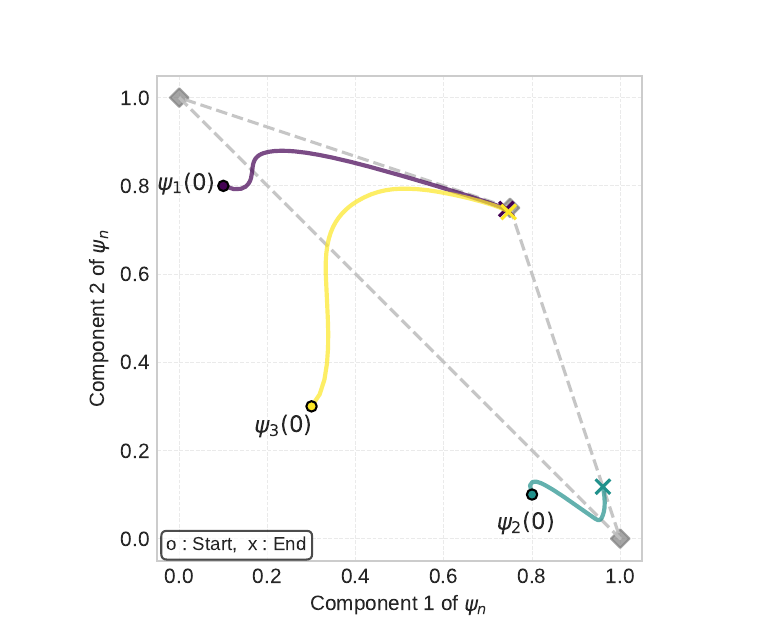}
  \includegraphics[width=0.45\linewidth]{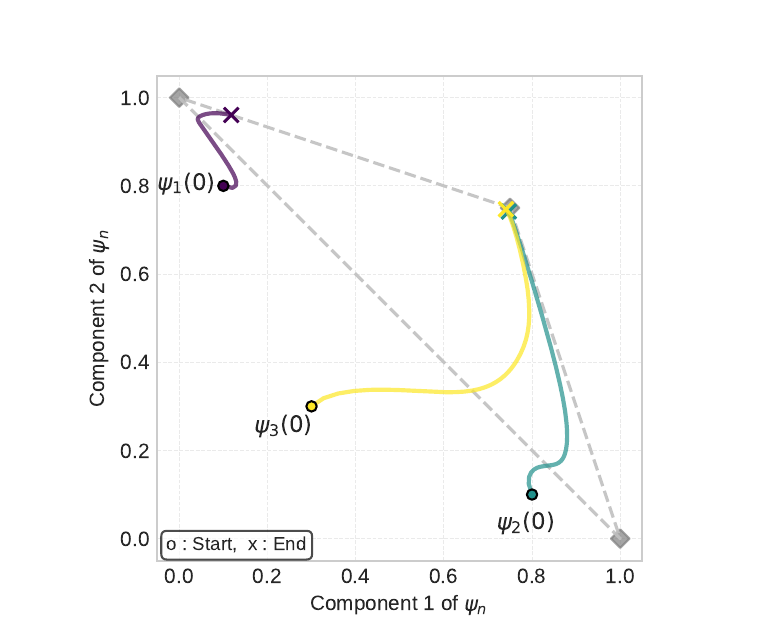}
  \caption{Effect of user attributes $\{v_n\}$ and arrival probabilities $\{\lambda_n\}$ on user preferences.}
  \label{fig:level-3-effect-of-V}
\end{figure}

Consider the trajectories from a numerical simulation shown in \cref{fig:level-3-effect-of-V}. The item attributes are given by $w_1=[1~0],$ $w_2=[0~1],$ $w_3=[0.75,0.75],$ and the model parameters are $a=12$ and $\rho=1.$ The simulation consists of three users with arrival probabilities $\lambda_1=0.3,$ $\lambda_2=0.3,$ and $\lambda_3=0.4.$ When the first and third users have similar attributes (left of \cref{fig:level-3-effect-of-V}), i.e., when $v_1=[1~0],$ $v_2=[0~1],$ and $v_3=[0.9~0.1],$ we observe that the long term preferences of users $1$ and $3$ become identical. Similarly, when the second and third users have similar attributes (right of \cref{fig:level-3-effect-of-V}), i.e., when $v_1=[1~0],$ $v_2=[0~1],$ and $v_3=[0.1~0.9],$ we observe that the long term preferences of users $2$ and $3$ become identical. This shows that users with similar attributes can end up having similar long term preferences.
\begin{figure}[ht]
  \centering
  \includegraphics[width=0.45\linewidth]{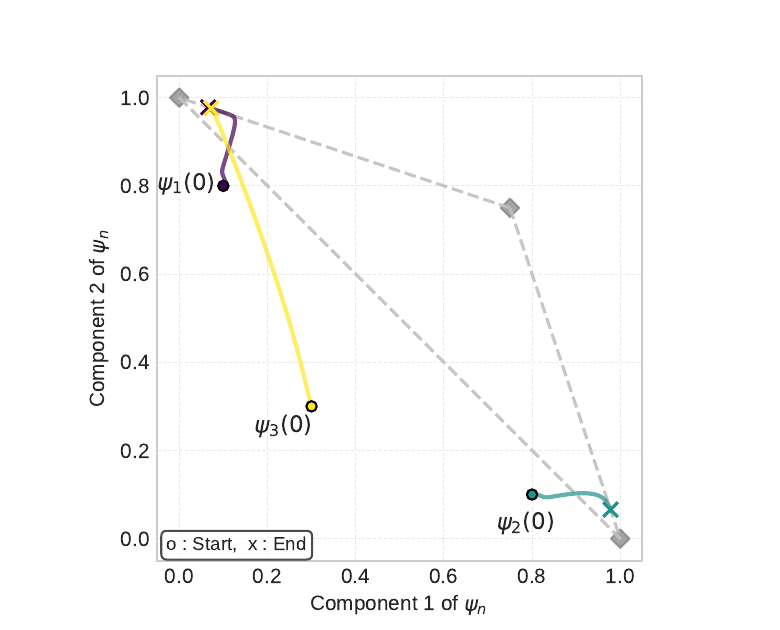}
  \includegraphics[width=0.45\linewidth]{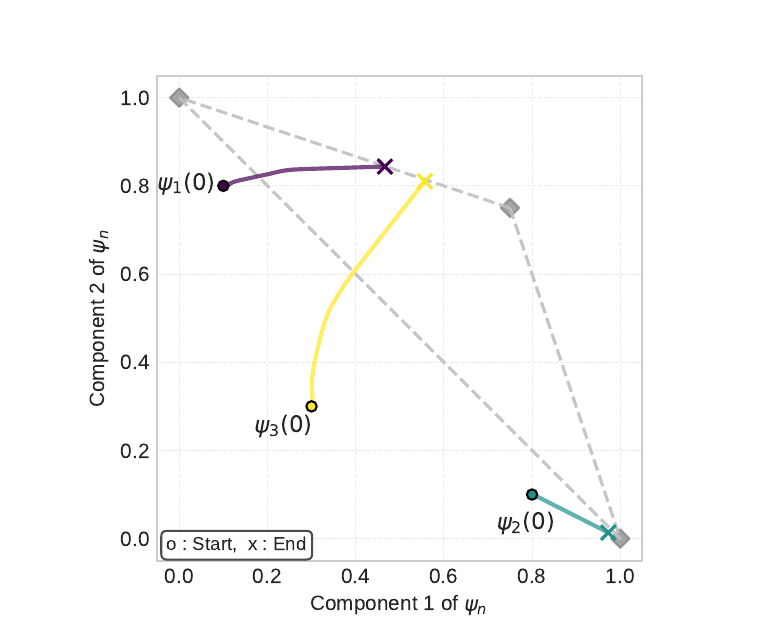}
  \caption{Effect of arrival probabilities $\{\lambda_n\}$ on user preferences.}
  \label{fig:level-3-effect-of-lambda}
\end{figure}

However, \RS can learn to differentiate between users with similar attributes if there is more data for the algorithm to learn from. This happens when similar users arrive to use \RS more often. In this way, the user arrival probabilities $\lambda_n$ also affect the long term preferences of the users. Consider the simulation showed in \cref{fig:level-3-effect-of-lambda}, in which users 1 and 3 are similar, i.e., the attributes are given by $v_1=[1~0],$ $v_2=[0~1],$ and $v_3=[0.99~0.01].$ When the arrival rates are given by $\lambda_1=0.49,$ $\lambda_2=0.49,$ and $\lambda_3=0.02$ (left of \cref{fig:level-3-effect-of-lambda}) users 1 and 3 are treated to be similar and are given similar recommendations, because there is not enough data on user 3 for \RS to differentiate user 3 from user 1. However, when the arrival rates are given by $\lambda_1=0.49,$ $\lambda_2=0.02,$ and $\lambda_3=0.49$ (right of \cref{fig:level-3-effect-of-lambda}), \RS receives more data on users 1 and 3, and is able to differentiate between them both. \RS is able to learn the preferences of users 1 and 3 and gives them different recommendations, which induces different long term preferences in both the users. It is observed that the long term preferences have changed as a consequence of the change in arrival probabilities.
\begin{figure}[ht]
  \centering
   \includegraphics[width=0.5\linewidth]{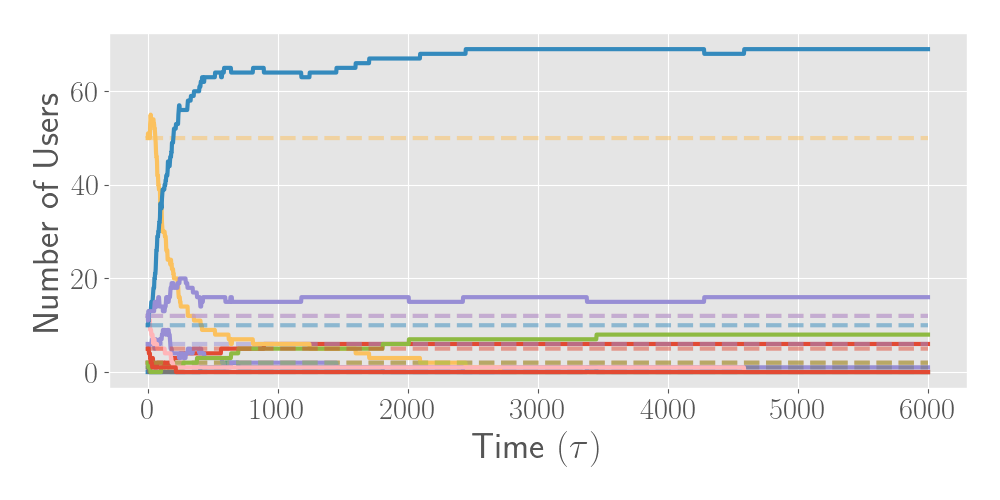}
  \includegraphics[width=0.5\linewidth]{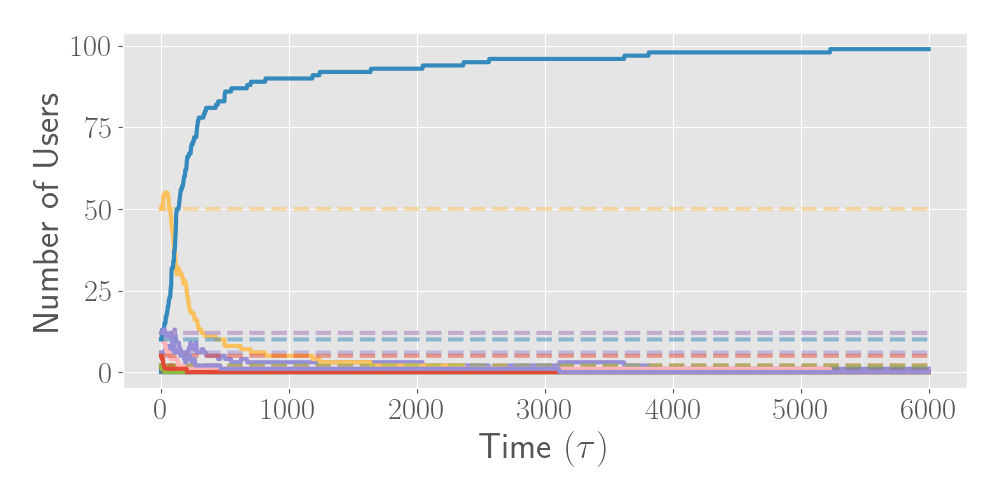}
 \caption{Effect of a large number of users on the preferences of users.}
  \label{fig:level-3-effect-of-N}
\end{figure}
\subsection{Effect of large number of users}
\label{ssec:level-3-large-N}
When there are large number of users, \RS can update the user attributes such that all the attributes lie in a $(p-1)$-dimensional hyperplane. While doing so ensures that \RS can learn user preferences, it is also observed that the preference tend to become homogeneous over time. Consider the plot in \cref{fig:level-3-effect-of-N} that shows the number of users who prefer a given item the most at any time $\tau.$ This simulation consists of 100 users whose preferences are randomly initialized within the unit hypersphere in $d$-dimensions. When the attributes are randomly generated (top subfigure in \cref{fig:level-3-effect-of-N}) the largest number of users that prefer a given item is around 70, but when the user attributes lie in a $(p-1)$-dimensional hyperplane, all the users end up preferring the same item the most over time. This is likely due to the inability of the recommendation algorithm to personalize items in the presence of large number of users. As more recommendations are made, the user preferences tend to become identical, making it easier for the algorithm to learn the same preferences.
\section{Summary}
In this chapter, we present a model of interaction with multiple users. To understand the long term behavior of the algorithm and the users, we obtain and analyze an asymptotic ODE that has the same long term properties as that of the model of interest. We show sufficient conditions for the convergence of the algorithm state and user preferences to equilibria with properties that are suitable for \RS to learn user preferences. We also show how the set of equilibria are affected by model parameters such as the exploration-exploitation parameter and the number of users.

%===========================
% Chapter 5
\chapter{Concluding Remarks}
\label{ch:conclusion}

In this thesis, we present and analyze a model of interaction between
a contextual bandit recommendation system and its users. The focus of
this work is to understand the long term consequences of the
interaction of a learning algorithm and its environment. Towards this,
we utilize the ODE method of stochastic approximation to study the stochastic model of interaction using a deterministic dynamical system that is more amenable for analysis. We identify that recommendation algorithms are capable of learning the user preferences in most cases. This is enabled by the evolution of user preferences such that they take simpler forms. Such simplification is observed when the recommendation algorithm explores more, in which case all user preferences converge to a single point. When the recommendation algorithm exploits more, the preferences tend to converge to a subset of the recommended items. This is indicative of the filter bubble phenomenon, in which the recommendation system learns to recommend specific types of items to users such that users eventually grow to prefer that item over others.

There are many avenues for future work. The effect of initial
conditions of the system on the long-term preferences can be studied.
When using the ODE method, this can be done by studying the basins of
attraction for each of the asymptotically stable equilibria. Also,
applying the method of analysis to other online learning settings is
an obvious next step. One can also relax the assumptions made by the
linear bandit model. An immediate extension is to allow the RS and the
users to have different views of the product attributes. Two-timescale
stochastic approximation theory can be applied to analyze the case
when the evolution of the learner and the environment happen at
different timescales. Using constant stepsizes for both the evolutions
of both the algorithm and the user preferences is also a natural next
question.

%=====================================================================
% APPENDIX
%  Appendices, if any, must precede the cited literatures.
%  Appendices shall be numbered in Roman Capitals (e.g. Appendix IV)

\begin{appendices}
  % Symbols

\chapter{List of Symbols}
\label{ch:sym}
%\addcontentsline{toc}{chapter}{\nameref{ch:sym}}
\makeatletter
\newcommand{\tocfill}{\cleaders\hbox{}\hfill}
\makeatother
\newcommand{\abbrlabel}[1]{\makebox[3cm][l]{\textbf{#1}\ \tocfill}}
\newenvironment{symbols}{\begin{list}{}{\renewcommand{\makelabel}{\abbrlabel}
                                            \setlength{\itemsep}{0pt}}}{\end{list}}
\begin{symbols}
\item[$\mathbb{R}^d$] 	Euclidean space of dimension $d$
\item[$K$]  Total number of available items
\item[$N$]  Total number of available users
\item[$p$]  Dimension of user attributes
\item[$q$]  Dimension of item attributes and user preferences.
\item[$t$] Time index of the stochastic interaction between RS and user(s)
\item[$\tau$] Time index used for ordinary differential equations
\item[$\theta_t$] 	Learner state at time $t$
\item[$Q_t$] Vector containing item recommendation history until time $t$
\item[$\psi_t$]	User preference (also called user state) at time $t$
\item[$\theta(\tau)$] 	The mean learner state at time $\tau$
\item[$q(\tau)$] The mean item recommendation history until time $\tau$
\item[$\psi(\tau)$]	The mean user preference at time $\tau$
\item[$w_k$] $q$-dimensional attribute of item $k$
\item[$W$] $q \times K$-dimensional matrix containing all item attributes
\item[$v_n$] $p$-dimensional attribute of user $n$
\item[$V$] $p \times N$-dimensional matrix containing all user attributes
\item[$c_{nk}$] $pq$-dimensional context of user $n$ and item $k$
\item[$C$] $pq \times NK$-dimensional matrix containing all contexts
\item[$\lambda_n$] The probability of arrival of user $n$ at any time
\item[$\alpha_t$] Learning rate of the learner
\item[$\beta_t$] Adaptability rate of the users
\item[$\rho$] The asymptotic ratio between the adaptability and learning rates
\item[$(a_t)$] A sequence of elements indexed by $t$
\item[$\{a_t\}$] A set of elements indexed by $t$
\item[{$[m]$}]  The set of elements $\{1,2,3,\ldots,m\}$ given a positive integer $m$
\item[$e_m$] A unit vector with 1 at index $m$ and 0 everywhere else 
\item[$\mathbf{1}_m$] A $m$-dimensional vector with all components equal to 1.
\item[$I_m$] Identity matrix of dimension $m \times m$
\item[$\|v\|$]  Euclidean norm of vector $v$
\item[$\|v\|_M$] Mahanalobis norm of vector $v$ with respect to a positive definite matrix $M$
\item[$\dotp{u,v}$] The canonical inner product of vectors $u$ and $v$
\item[$M^{\top}$] Transpose of matrix $M.$
\item[$\|M\|$] The operator norm of matrix $M$ with respect to the euclidean norm
\item[$M \otimes P$] The Kronecker product between vectors/matrices $M$ and $P$
\item[$\diag(v)$] A diagonal matrix having vector $v$ as its diagonal
\item[$\nabla_v f$] Gradient of a differentiable function $f$ with respect to $v$
\item[{$[\cdot]$}] Iverson bracket (similar to the indicator function)
\end{symbols}

\end{appendices}

%=====================================================================
% BIBLIOGRAPHY
\newpage
\setlength{\parskip}{5mm}
\titlespacing{\chapter}{0cm}{0mm}{0mm}
\titleformat{\chapter}[display]
  {\normalfont\huge\bfseries}
  {\chaptertitlename\ \thechapter}{20pt}{\Huge}

\bibliographystyle{unsrtnat}
% Add the bib file
\bibliography{refs}

\begin{thebibliography}{49}
\providecommand{\natexlab}[1]{#1}
\providecommand{\url}[1]{\texttt{#1}}
\expandafter\ifx\csname urlstyle\endcsname\relax
  \providecommand{\doi}[1]{doi: #1}\else
  \providecommand{\doi}{doi: \begingroup \urlstyle{rm}\Url}\fi

\bibitem[Resnick and Varian(1997)]{resnick1997recommender}
Paul Resnick and Hal~R Varian.
\newblock Recommender systems.
\newblock \emph{Communications of the ACM}, 40\penalty0 (3):\penalty0 56--58,
  1997.

\bibitem[Covington et~al.(2016)Covington, Adams, and Sargin]{covington2016deep}
Paul Covington, Jay Adams, and Emre Sargin.
\newblock Deep neural networks for youtube recommendations.
\newblock In \emph{Proceedings of the 10th ACM conference on recommender
  systems}, pages 191--198, 2016.

\bibitem[Smith and Linden(2017)]{smith2017twoamazon}
Brent Smith and Greg Linden.
\newblock Two decades of recommender systems at amazon.com.
\newblock \emph{Ieee internet computing}, 21\penalty0 (3):\penalty0 12--18,
  2017.

\bibitem[Munn(2020)]{munn2020angry}
Luke Munn.
\newblock Angry by design: toxic communication and technical architectures.
\newblock \emph{Humanities and Social Sciences Communications}, 7\penalty0
  (1):\penalty0 1--11, 2020.

\bibitem[Nguyen et~al.(2014)Nguyen, Hui, Harper, Terveen, and
  Konstan]{nguyen2014exploring}
Tien~T Nguyen, Pik-Mai Hui, F~Maxwell Harper, Loren Terveen, and Joseph~A
  Konstan.
\newblock Exploring the filter bubble: the effect of using recommender systems
  on content diversity.
\newblock In \emph{Proceedings of the 23rd international conference on World
  wide web}, pages 677--686, 2014.

\bibitem[Franklin et~al.(2022)Franklin, Ashton, Gorman, and
  Armstrong]{franklin2022recognising}
Matija Franklin, Hal Ashton, Rebecca Gorman, and Stuart Armstrong.
\newblock Recognising the importance of preference change: A call for a
  coordinated multidisciplinary research effort in the age of {AI}.
\newblock \emph{arXiv preprint arXiv:2203.10525}, 2022.

\bibitem[Dean et~al.(2024)Dean, Dong, Jagadeesan, and Leqi]{dean2024accounting}
Sarah Dean, Evan Dong, Meena Jagadeesan, and Liu Leqi.
\newblock Accounting for {AI} and users shaping one another: The role of
  mathematical models.
\newblock \emph{arXiv preprint arXiv:2404.12366}, 2024.

\bibitem[Koren et~al.(2009)Koren, Bell, and Volinsky]{koren2009matrix}
Yehuda Koren, Robert Bell, and Chris Volinsky.
\newblock Matrix factorization techniques for recommender systems.
\newblock \emph{Computer}, 42\penalty0 (8):\penalty0 30--37, 2009.

\bibitem[Jiang et~al.(2019)Jiang, Chiappa, Lattimore, Gy{\"o}rgy, and
  Kohli]{jiang2019degenerate}
Ray Jiang, Silvia Chiappa, Tor Lattimore, Andr{\'a}s Gy{\"o}rgy, and Pushmeet
  Kohli.
\newblock Degenerate feedback loops in recommender systems.
\newblock In \emph{Proceedings of the 2019 AAAI/ACM Conference on AI, Ethics,
  and Society}, pages 383--390, 2019.

\bibitem[Rossi et~al.(2021)Rossi, Polderman, and Frasca]{rossi2021closed}
Wilbert~Samuel Rossi, Jan~Willem Polderman, and Paolo Frasca.
\newblock The closed loop between opinion formation and personalized
  recommendations.
\newblock \emph{IEEE Transactions on Control of Network Systems}, 9\penalty0
  (3):\penalty0 1092--1103, 2021.

\bibitem[Kalimeris et~al.(2021)Kalimeris, Bhagat, Kalyanaraman, and
  Weinsberg]{kalimeris2021preference}
Dimitris Kalimeris, Smriti Bhagat, Shankar Kalyanaraman, and Udi Weinsberg.
\newblock Preference amplification in recommender systems.
\newblock In \emph{Proceedings of the 27th ACM SIGKDD Conference on Knowledge
  Discovery \& Data Mining}, pages 805--815, 2021.

\bibitem[Dean and Morgenstern(2022)]{dean2022preference}
Sarah Dean and Jamie Morgenstern.
\newblock Preference dynamics under personalized recommendations.
\newblock In \emph{Proceedings of the 23rd ACM Conference on Economics and
  Computation}, pages 795--816, 2022.

\bibitem[Brown and Agarwal(2022)]{brown2022diversified}
William Brown and Arpit Agarwal.
\newblock Diversified recommendations for agents with adaptive preferences.
\newblock \emph{Advances in Neural Information Processing Systems},
  35:\penalty0 26066--26077, 2022.

\bibitem[Kleinberg et~al.(2024)Kleinberg, Mullainathan, and
  Raghavan]{kleinberg2024challenge}
Jon Kleinberg, Sendhil Mullainathan, and Manish Raghavan.
\newblock The challenge of understanding what users want: Inconsistent
  preferences and engagement optimization.
\newblock \emph{Management science}, 70\penalty0 (9):\penalty0 6336--6355,
  2024.

\bibitem[Ekstrand et~al.(2011)Ekstrand, Riedl, Konstan,
  et~al.]{ekstrand2011collaborative}
Michael~D Ekstrand, John~T Riedl, Joseph~A Konstan, et~al.
\newblock Collaborative filtering recommender systems.
\newblock \emph{Foundations and Trends{\textregistered} in Human--Computer
  Interaction}, 4\penalty0 (2):\penalty0 81--173, 2011.

\bibitem[Zhang et~al.(2019)Zhang, Yao, Sun, and Tay]{zhang2019deep}
Shuai Zhang, Lina Yao, Aixin Sun, and Yi~Tay.
\newblock Deep learning based recommender system: A survey and new
  perspectives.
\newblock \emph{ACM computing surveys (CSUR)}, 52\penalty0 (1):\penalty0 1--38,
  2019.

\bibitem[Li et~al.(2010)Li, Chu, Langford, and Schapire]{li2010contextual}
Lihong Li, Wei Chu, John Langford, and Robert~E Schapire.
\newblock A contextual-bandit approach to personalized news article
  recommendation.
\newblock In \emph{Proceedings of the 19th international conference on World
  wide web}, pages 661--670, 2010.

\bibitem[McInerney et~al.(2018)McInerney, Lacker, Hansen, Higley, Bouchard,
  Gruson, and Mehrotra]{mcinerney2018explore}
James McInerney, Benjamin Lacker, Samantha Hansen, Karl Higley, Hugues
  Bouchard, Alois Gruson, and Rishabh Mehrotra.
\newblock Explore, exploit, and explain: personalizing explainable
  recommendations with bandits.
\newblock In \emph{Proceedings of the 12th ACM conference on recommender
  systems}, pages 31--39, 2018.

\bibitem[Meshram et~al.(2015)Meshram, Manjunath, and
  Gopalan]{meshram2015restless}
Rahul Meshram, D~Manjunath, and Aditya Gopalan.
\newblock A restless bandit with no observable states for recommendation
  systems and communication link scheduling.
\newblock In \emph{2015 54th IEEE Conference on Decision and Control (CDC)},
  pages 7820--7825. IEEE, 2015.

\bibitem[Kleinberg and Immorlica(2018)]{kleinberg2018recharging}
Robert Kleinberg and Nicole Immorlica.
\newblock Recharging bandits.
\newblock In \emph{2018 IEEE 59th Annual Symposium on Foundations of Computer
  Science (FOCS)}, pages 309--319. IEEE, 2018.

\bibitem[Shah et~al.(2018)Shah, Blanchet, and Johari]{shah2018bandit}
Virag Shah, Jose Blanchet, and Ramesh Johari.
\newblock Bandit learning with positive externalities.
\newblock \emph{Advances in Neural Information Processing Systems}, 31, 2018.

\bibitem[Zhou et~al.(2021)Zhou, Liu, Dong, and Deng]{zhou2021incentivized}
Tianchen Zhou, Jia Liu, Chaosheng Dong, and Jingyuan Deng.
\newblock Incentivized bandit learning with self-reinforcing user preferences.
\newblock In \emph{International Conference on Machine Learning}, pages
  12824--12834. PMLR, 2021.

\bibitem[Koren(2009)]{koren2009collaborativetemporal}
Yehuda Koren.
\newblock Collaborative filtering with temporal dynamics.
\newblock In \emph{Proceedings of the 15th ACM SIGKDD international conference
  on Knowledge discovery and data mining}, pages 447--456, 2009.

\bibitem[Li et~al.(2017)Li, Ren, Chen, Ren, Lian, and Ma]{li2017neural}
Jing Li, Pengjie Ren, Zhumin Chen, Zhaochun Ren, Tao Lian, and Jun Ma.
\newblock Neural attentive session-based recommendation.
\newblock In \emph{Proceedings of the 2017 ACM on Conference on Information and
  Knowledge Management}, pages 1419--1428, 2017.

\bibitem[Li et~al.(2018)Li, Zhao, Liu, Huang, Mei, and Chen]{li2018learning}
Zhi Li, Hongke Zhao, Qi~Liu, Zhenya Huang, Tao Mei, and Enhong Chen.
\newblock Learning from history and present: Next-item recommendation via
  discriminatively exploiting user behaviors.
\newblock In \emph{Proceedings of the 24th ACM SIGKDD international conference
  on knowledge discovery \& data mining}, pages 1734--1743, 2018.

\bibitem[Afsar et~al.(2022)Afsar, Crump, and Far]{afsar2022reinforcement}
M~Mehdi Afsar, Trafford Crump, and Behrouz Far.
\newblock Reinforcement learning based recommender systems: A survey.
\newblock \emph{ACM Computing Surveys}, 55\penalty0 (7):\penalty0 1--38, 2022.

\bibitem[Dandekar et~al.(2013)Dandekar, Goel, and Lee]{dandekar2013biased}
Pranav Dandekar, Ashish Goel, and David~T Lee.
\newblock Biased assimilation, homophily, and the dynamics of polarization.
\newblock \emph{Proceedings of the National Academy of Sciences}, 110\penalty0
  (15):\penalty0 5791--5796, 2013.

\bibitem[Proskurnikov and Tempo(2017)]{proskurnikov2017tutorial}
Anton~V Proskurnikov and Roberto Tempo.
\newblock A tutorial on modeling and analysis of dynamic social networks.
  {P}art {I}.
\newblock \emph{Annual Reviews in Control}, 43:\penalty0 65--79, 2017.

\bibitem[Curmei et~al.(2022)Curmei, Haupt, Recht, and
  Hadfield-Menell]{curmei2022towards}
Mihaela Curmei, Andreas~A Haupt, Benjamin Recht, and Dylan Hadfield-Menell.
\newblock Towards psychologically-grounded dynamic preference models.
\newblock In \emph{Proceedings of the 16th ACM Conference on Recommender
  Systems}, pages 35--48, 2022.

\bibitem[Carroll et~al.(2022)Carroll, Dragan, Russell, and
  Hadfield-Menell]{carroll2022estimating}
Micah~D Carroll, Anca Dragan, Stuart Russell, and Dylan Hadfield-Menell.
\newblock Estimating and penalizing induced preference shifts in recommender
  systems.
\newblock In \emph{International Conference on Machine Learning}, pages
  2686--2708. PMLR, 2022.

\bibitem[Wang et~al.(2023)Wang, Lin, Wang, Feng, Ma, and Chua]{wang2023causal}
Wenjie Wang, Xinyu Lin, Liuhui Wang, Fuli Feng, Yunshan Ma, and Tat-Seng Chua.
\newblock Causal disentangled recommendation against user preference shifts.
\newblock \emph{ACM Transactions on Information Systems}, 42\penalty0
  (1):\penalty0 1--27, 2023.

\bibitem[Adomavicius et~al.(2013)Adomavicius, Bockstedt, Curley, and
  Zhang]{adomavicius2013recommender}
Gediminas Adomavicius, Jesse~C Bockstedt, Shawn~P Curley, and Jingjing Zhang.
\newblock Do recommender systems manipulate consumer preferences? {A} study of
  anchoring effects.
\newblock \emph{Information Systems Research}, 24\penalty0 (4):\penalty0
  956--975, 2013.

\bibitem[Porcaro et~al.(2024)Porcaro, G{\'o}mez, and
  Castillo]{porcaro2024assessing}
Lorenzo Porcaro, Emilia G{\'o}mez, and Carlos Castillo.
\newblock Assessing the impact of music recommendation diversity on listeners:
  A longitudinal study.
\newblock \emph{ACM Transactions on Recommender Systems}, 2\penalty0
  (1):\penalty0 1--47, 2024.

\bibitem[Mansoury et~al.(2020)Mansoury, Abdollahpouri, Pechenizkiy, Mobasher,
  and Burke]{mansoury2020feedback}
Masoud Mansoury, Himan Abdollahpouri, Mykola Pechenizkiy, Bamshad Mobasher, and
  Robin Burke.
\newblock Feedback loop and bias amplification in recommender systems.
\newblock In \emph{Proceedings of the 29th ACM international conference on
  information \& knowledge management}, pages 2145--2148, 2020.

\bibitem[Chaney et~al.(2018)Chaney, Stewart, and
  Engelhardt]{chaney2018algorithmic}
Allison~JB Chaney, Brandon~M Stewart, and Barbara~E Engelhardt.
\newblock How algorithmic confounding in recommendation systems increases
  homogeneity and decreases utility.
\newblock In \emph{Proceedings of the 12th ACM conference on recommender
  systems}, pages 224--232, 2018.

\bibitem[Chen et~al.(2023)Chen, Dong, Wang, Feng, Wang, and He]{chen2023bias}
Jiawei Chen, Hande Dong, Xiang Wang, Fuli Feng, Meng Wang, and Xiangnan He.
\newblock Bias and debias in recommender system: A survey and future
  directions.
\newblock \emph{ACM Transactions on Information Systems}, 41\penalty0
  (3):\penalty0 1--39, 2023.

\bibitem[Gao et~al.(2024)Gao, Zheng, Wang, Feng, He, and Li]{gao2024causal}
Chen Gao, Yu~Zheng, Wenjie Wang, Fuli Feng, Xiangnan He, and Yong Li.
\newblock Causal inference in recommender systems: A survey and future
  directions.
\newblock \emph{ACM Transactions on Information Systems}, 42\penalty0
  (4):\penalty0 1--32, 2024.

\bibitem[Hardt et~al.(2022)Hardt, Jagadeesan, and
  Mendler-D{\"u}nner]{hardt2022performative}
Moritz Hardt, Meena Jagadeesan, and Celestine Mendler-D{\"u}nner.
\newblock Performative power.
\newblock \emph{Advances in Neural Information Processing Systems},
  35:\penalty0 22969--22981, 2022.

\bibitem[Agarwal et~al.(2024)Agarwal, Usunier, Lazaric, and
  Nickel]{agarwal2024system}
Arpit Agarwal, Nicolas Usunier, Alessandro Lazaric, and Maximilian Nickel.
\newblock System-2 recommenders: Disentangling utility and engagement in
  recommendation systems via temporal point-processes.
\newblock In \emph{Proceedings of the 2024 ACM Conference on Fairness,
  Accountability, and Transparency}, pages 1763--1773, 2024.

\bibitem[Jameson et~al.(2015)Jameson, Willemsen, Felfernig, De~Gemmis, Lops,
  Semeraro, and Chen]{jameson2015human}
Anthony Jameson, Martijn~C Willemsen, Alexander Felfernig, Marco De~Gemmis,
  Pasquale Lops, Giovanni Semeraro, and Li~Chen.
\newblock Human decision making and recommender systems.
\newblock \emph{Recommender systems handbook}, pages 611--648, 2015.

\bibitem[Lattimore and Szepesv{\'a}ri(2020)]{lattimore2020bandit}
Tor Lattimore and Csaba Szepesv{\'a}ri.
\newblock \emph{Bandit algorithms}, chapter~19, pages 237--252.
\newblock Cambridge University Press, 2020.

\bibitem[Shalev-Shwartz et~al.(2012)]{shalev2012online}
Shai Shalev-Shwartz et~al.
\newblock Online learning and online convex optimization.
\newblock \emph{Foundations and Trends{\textregistered} in Machine Learning},
  4\penalty0 (2):\penalty0 107--194, 2012.

\bibitem[Sutton et~al.(1998)Sutton, Barto, et~al.]{sutton1998reinforcement}
Richard~S Sutton, Andrew~G Barto, et~al.
\newblock \emph{Reinforcement learning: An introduction}, volume~1.
\newblock MIT press Cambridge, 1998.

\bibitem[Borkar(2009)]{borkar2009stochastic}
Vivek~S Borkar.
\newblock \emph{Stochastic approximation: a dynamical systems viewpoint},
  volume~48.
\newblock Springer, 2009.

\bibitem[Prashanth et~al.(2025)Prashanth, Bhatnagar,
  et~al.]{prashanth2025gradient}
LA~Prashanth, Shalabh Bhatnagar, et~al.
\newblock Gradient-based algorithms for zeroth-order optimization.
\newblock \emph{Foundations and Trends{\textregistered} in Optimization},
  8\penalty0 (1--3):\penalty0 1--332, 2025.

\bibitem[Miller(1981)]{miller1981inverse}
Kenneth~S Miller.
\newblock On the inverse of the sum of matrices.
\newblock \emph{Mathematics magazine}, 54\penalty0 (2):\penalty0 67--72, 1981.

\bibitem[Khalil(2002)]{khalil2002nonlinear}
Hassan~K Khalil.
\newblock \emph{Nonlinear systems}, volume~3.
\newblock Prentice hall Upper Saddle River, NJ, 2002.

\bibitem[Hoeffler and Ariely(1999)]{hoeffler1999constructing}
Steve Hoeffler and Dan Ariely.
\newblock Constructing stable preferences: A look into dimensions of experience
  and their impact on preference stability.
\newblock \emph{Journal of consumer psychology}, 8\penalty0 (2):\penalty0
  113--139, 1999.

\bibitem[Carreira-Perpin{\'a}n and Williams(2003)]{carreira2003number}
Miguel~A Carreira-Perpin{\'a}n and Christopher~KI Williams.
\newblock On the number of modes of a gaussian mixture.
\newblock In \emph{International Conference on Scale-Space Theories in Computer
  Vision}, pages 625--640. Springer, 2003.

\end{thebibliography}

%=====================================================================
% PUBLICATIONS
%  publications if any may be listed after the literature cited.
%\addcontentsline{toc}{chapter}{List of Publications}
%\include{chap_07_List_of_Publications}

%=====================================================================
% ACKNOWLEDGMENTS
%   This is the last item in the thesis. It should be signed by
%   author, with date.

\end{document}